\documentclass[lettersize,journal]{IEEEtran}
\usepackage{amsmath,amsfonts}
\usepackage{algorithmic}
\usepackage{algorithm}
\usepackage{array}
\usepackage[T1]{fontenc}
\usepackage{textcomp}
\usepackage{stfloats}
\usepackage{url}
\usepackage{textcomp}
\usepackage{verbatim}
\usepackage{graphicx}
\usepackage{cite}
\usepackage{color}
\usepackage{multirow}
\usepackage{booktabs} 
\usepackage{color,xcolor}
\usepackage{bbm}
\usepackage{pifont}
\usepackage{subcaption}
\usepackage{bm}
\usepackage{makecell}
\usepackage{tikz}
\usetikzlibrary{calc}
\usepackage[colorlinks,linkcolor=red,citecolor=green]{hyperref}
\usepackage{ulem}
 
\begin{document}


\title{ProGS: Towards Progressive Coding for 3D Gaussian Splatting}

\author{
        Zhiye~Tang,
        Lingzhuo~Liu,
        Shengjie~Jiao,
        Qiudan~Zhang,~\IEEEmembership{Member,~IEEE,}
        You Yang, ~\IEEEmembership{Senior Member,~IEEE,}
        Junhui~Hou, ~\IEEEmembership{Senior Member,~IEEE,}
        Xu Wang,~\IEEEmembership{Member,~IEEE}
\thanks{
This work was supported in part by the National Natural Science Foundation of China under Grants 62371310, 62501403, and 62422118, in part by the Shenzhen Science and Technology Program (JCYJ20241202124415021), in part by the Guangdong Basic and Applied Basic Research Foundation under Grant 2023A1515011236, and in part by the Hong Kong Research Grants Council under Grant N\_CityU1114/25. \textit{(Corresponding author: Xu Wang.)}

Zhiye Tang, Lingzhuo Liu, Shengjie Jiao, Qiudan Zhang and Xu Wang are with the College of Computer Science and Software Engineering, Shenzhen University, Shenzhen, 518060, China. Email: (tangzhiye2022@foxmail.com, karrie019@qq.com, jiaoshengjie@icloud.com, qiudanzhang@szu.edu.cn, wangxu@szu.edu.cn).

Junhui Hou is with the Department of Computer Science, City University of Hong Kong, Kowloon, Hong Kong SAR, China. Email: (jh.hou@cityu.edu.hk).

You Yang is with the School of Electronic Information and Communications, Huazhong University of Science and Technology, Wuhan, 430074, China. Email: (yangyou@hust.edu.cn).

}
}



\maketitle
\begin{abstract}
Progressive transmission of 3D Gaussian Splatting (3DGS) requires each completed transmission stage to be decodable from received data and directly renderable. This work presents ProGS, a progressive codec that organizes anchor-based 3DGS as parent-closed octree prefixes. ProGS combines parent-causal entropy coding, level-balanced anchor growth, bounded multi-prefix training, and lightweight parent-anchor refinement to improve early-prefix quality without altering the complete-model rendering path. One fixed-$\lambda$ training run yields five deployable rate--quality points from a single bitstream. Experiments on 17 scenes from three datasets evaluate rate--distortion performance, rendering speed, and transmission efficiency against progressive and single-rate baselines. On one representative scene per dataset, ProGS reaches a common quality target with 30.7 $\sim$ 60.7\% fewer bytes than HAC-Rand and 22.6 $\sim$ 53.4\% fewer bytes than HAC++-Rand. Across the three dataset averages, ProGS-LR uses 4.7 $\sim$ 6.1\% fewer bytes than HAC-high while improving SSIM by 0.002 $\sim$ 0.041 and reducing LPIPS by 4.8 $\sim$ 51.2\%. The parent-closed syntax makes every prefix causally decodable and directly renderable without future topology. ProGS-HR also yields higher endpoint SSIM and lower LPIPS than PCGS across all three dataset averages, and all five prefixes render in real time. Code is available at \url{https://github.com/ZhiyeTang/ProGS-Official}.
\end{abstract}


\begin{figure*}
    \centering
    \includegraphics[width=0.95\linewidth]{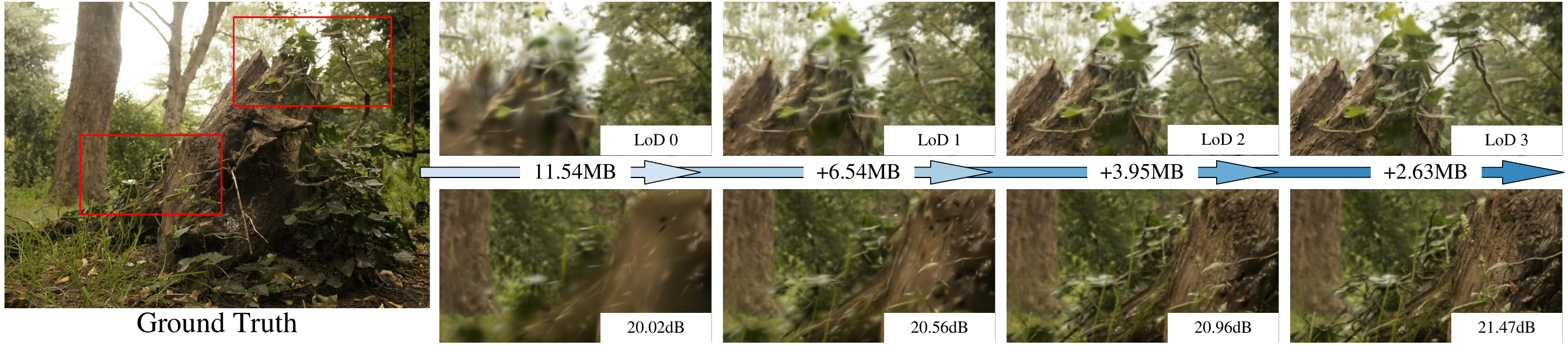}
    \caption{\textbf{ProGS} structures a 3D scene as a forest of layered octrees in a shared coordinate system. Each level-of-detail (LoD) $l$ contains the cumulative octree layers through $l$. With context-based progressive coding, every completed level chunk yields a decodable and renderable model at a larger received payload.}
    \label{fig:progs}
\end{figure*}
   
\section{Introduction}\label{sec:introduction}

3D Gaussian Splatting (3DGS)~\cite{kerbl3Dgaussians} is an explicit representation for novel view synthesis (NVS)~\cite{fei20243d}. Unlike neural radiance fields (NeRF)~\cite{mildenhall2021nerf}, 3DGS represents a scene with learnable anisotropic Gaussians, typically initialized from a structure-from-motion (SfM) point cloud~\cite{schonberger2016structure} and rendered by projecting and $\alpha$-blending these primitives~\cite{lassner2021pulsar}. This differentiable rasterization supports high-quality real-time rendering, but a scene may require millions of Gaussians and therefore substantial memory and storage.

Compressing an unstructured set of Gaussians is difficult because both the primitive count and the per-primitive attributes are large~\cite{fei20243d,chen2024survey3dgaussiansplatting}. Existing methods prune low-utility primitives~\cite{lee2024compact3dgaussianrepresentation,fan2024lightgaussianunbounded3dgaussian,wang2024rdogaussian}, quantize attributes, or share values through codebooks~\cite{navaneet2024compact3dsmallerfastergaussian}. These strategies reduce complete-model storage, but they operate on a flat primitive set with no hierarchy to order partial models.

Building on this, Scaffold-GS~\cite{lu2023scaffoldgsstructured3dgaussians} replaces individually stored Gaussians with voxel-aligned anchors that generate local neural Gaussians at render time, and HAC~\cite{chen2024hachashgridassistedcontext} compresses this representation further through hash-grid context and adaptive quantization. These anchor-based methods provide strong endpoint coding efficiency, but their reported operating points are complete single-rate models. Hierarchical and progressive methods address a different requirement: Octree-GS~\cite{ren2024octreegsconsistentrealtimerendering} organizes anchors into multiple levels of detail (LoDs) for render-time adaptation, while PCGS~\cite{chen2026pcgs} directly supports cumulative, renderable levels and is therefore the closest codec-level baseline for ProGS.

Progressive 3DGS coding adapts scene transmission to available bandwidth by producing increasingly detailed renderable models. In this paper, a \textit{bitstream prefix} consists of the shared header and all level chunks through a completed level. A valid prefix must be structurally complete and entropy-decodable from received data. This couples three design problems: constructing useful intermediate models from a nonhierarchical 3DGS representation, restricting entropy dependencies to decoded state, and optimizing early prefixes without degrading complete-model quality. Although an available anchor subset may be renderable, it may omit the parent attributes required by a hierarchical entropy model. ProGS addresses these problems jointly through a parent-closed octree hierarchy, causal coding order, and prefix-aware optimization.

Following the anchor-based representation of Scaffold-GS, ProGS constructs a parent-closed forest of octrees over scene anchors. The cumulative anchors through a given LoD form an \textit{octree prefix} that contains every parent required to decode its descendants, and all decoded anchors are rendered jointly without future levels. A parent-causal entropy model encodes finer-level attributes using only decoded parent and prefix information. Because parent closure can disproportionately increase the payload of coarse LoDs, level-balanced growth controls this structural overhead. Multi-LoD optimization and lightweight parent-anchor refinement further improve early-prefix reconstruction while leaving the complete-model rendering path unchanged. 

The main contributions are summarized as follows:
\begin{enumerate}
    \item \textbf{Topology-progressive octree prefixes.} ProGS transmits parent-closed octree levels so every completed prefix is structurally self-contained, directly renderable, and causally decodable without future topology.

    \item \textbf{Parent-causal entropy coding.} An entropy coding scheme serializes anchors in coarse-to-fine order and predicts finer-level attributes only from previously decoded parent-level information and octree context, so that every prefix is decodable without accessing future levels.

    \item \textbf{Progressive structure and optimization.} A training pipeline combines level-balanced structure growth, multi-LoD supervision, and lightweight parent-anchor refinement. These components reduce the structural overhead introduced by parent closure and improve the visual quality of low-rate prefixes while preserving the full-LoD reconstruction path. One fixed-$\lambda$ model thus serves five deployment rates without separate complete models.

\end{enumerate}

\section{Related Work}\label{sec:related_work}

\subsection{Compression for 3D Gaussian Splatting}

The explicit primitives of 3D Gaussian Splatting (3DGS)~\cite{kerbl3Dgaussians} enable high-quality real-time rendering but produce large scene representations. Direct compression methods reduce this cost through pruning, quantization, codebooks, and structured attribute transforms~\cite{lee2024compact3dgaussianrepresentation,navaneet2024compact3dsmallerfastergaussian}. Representative systems include LightGaussian~\cite{fan2024lightgaussianunbounded3dgaussian}, RDO-Gaussian~\cite{wang2024rdogaussian}, EAGLES~\cite{girish2024eaglesefficientaccelerated3d}, CompGS~\cite{liu2024compgsefficient3dscene}, R-3DGS~\cite{papantonakis2024reducing}, and Compressed3D~\cite{niedermayr2024compressed}; more structured variants arrange Gaussians on regular grids (SOG~\cite{morgenstern2023compact}), separate explicit and implicit attributes (LocoGS~\cite{shin2025localityaware}), apply post-training voxelization and hierarchical transforms (MesonGS~\cite{xie2025mesongs,7482691}), or target pretrained scenes (FCGS~\cite{chen2025fast}). These methods establish strong storage--quality trade-offs, but their reported operating points are complete models rather than successive prefixes of one encoded stream.

Anchor-based representations provide another route to compact 3DGS: Scaffold-GS~\cite{lu2023scaffoldgsstructured3dgaussians} uses sparse voxel-aligned anchors whose features generate local Gaussians at render time; HAC~\cite{chen2024hachashgridassistedcontext} and HAC++~\cite{chen2025hacplus} add hash-grid context and adaptive quantization; ContextGS~\cite{wang2024contextgscompact3dgaussian} introduces an anchor-level context model; and CAT-3DGS~\cite{zhan2025cat3dgs} strengthens single-rate compression through improved context and rate-distortion optimization. Their rate points are obtained from separately optimized models or complete bitstreams and are therefore reported as single-rate references. HAC and HAC++ are additionally packetized after training to measure the benefit of learned progressive ordering.

\subsection{LoD-Structured and Progressive Coding}

Scalable image and video codecs transmit a base representation followed by enhancement information, e.g., progressive JPEG/JPEG2000~\cite{wallace1992jpeg,skodras2001jpeg}, learned progressive image coding~\cite{toderici2017full,lee2022dpict,jeon2023context}, and scalable video coding~\cite{stutz2011survey,boyce2015overview}. Point-cloud codecs face the additional irregular-geometry problem, which Geometry-based Point Cloud Compression (G-PCC) addresses with octree- and LoD-based scalability~\cite{ISO_GPCC_2nd_Edition}, later combined with learned prediction~\cite{zhang2023scalable,mao2024spac}. These systems motivate coarse-to-fine delivery, but 3DGS must additionally preserve learned appearance, covariance, and anchor attributes in every renderable intermediate representation.

Within 3DGS, Octree-GS~\cite{ren2024octreegsconsistentrealtimerendering} organizes anchors into render-time LoDs but does not encode them as cumulative entropy-coded chunks. PCGS~\cite{chen2026pcgs} provides progressive decoding through progressive masking, quantization, and probability prediction from previous levels. ProGS instead makes the coding hierarchy parent-closed and conditions each level on its decoded parents, aligning structural dependencies with the progressive bitstream.

\definecolor{purplePro}{RGB}{190,134,186}
\definecolor{orangePro}{RGB}{234,112, 13}
\definecolor{redPro}{RGB}{146, 57, 49}
\definecolor{goldPro}{RGB}{255,215,0}
\definecolor{silverPro}{RGB}{191,193,194}
\definecolor{colorProGS}{RGB}{57,81,162}
\newcommand{\methodFirstRank}[1]{\textbf{#1}}
\newcommand{\methodSecondRank}[1]{\underline{#1}}
\newcommand{\methodThirdRank}[1]{\textit{#1}}

\begin{figure*}
    \centering
    \includegraphics[width=\textwidth]{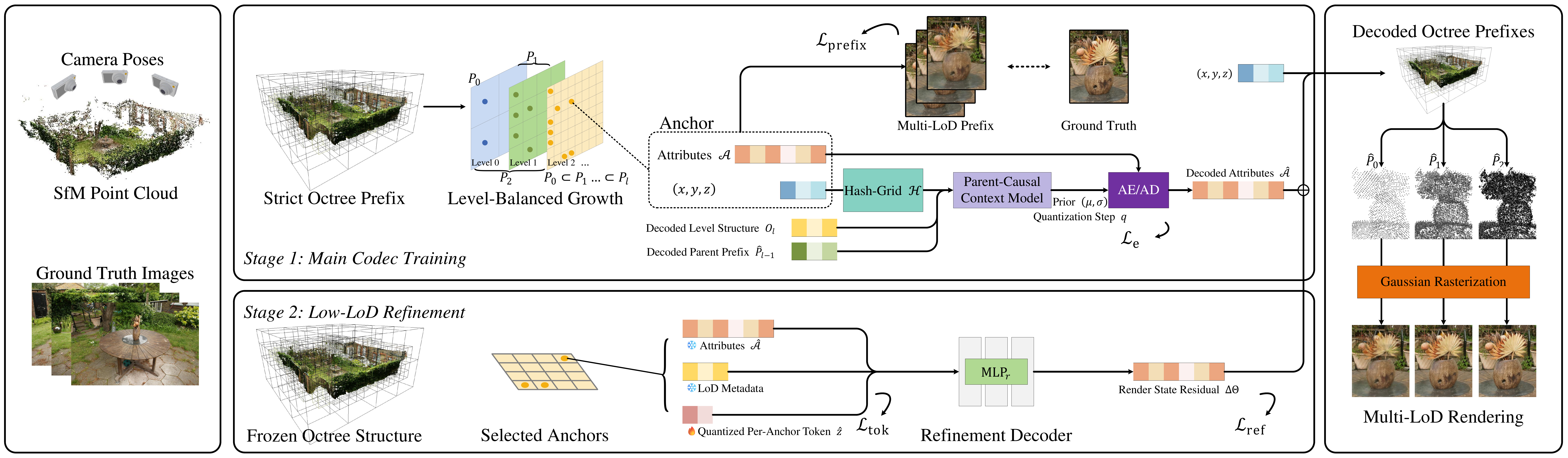}
    \caption{Overview of ProGS. Stage~1 constructs parent-closed octree prefixes via level-balanced growth and multi-LoD supervision, then entropy-codes anchor attributes with priors and quantization steps predicted from the decoded parent prefix and current-level structure (AE/AD: arithmetic encoding/decoding). Stage~2 freezes the base codec and predicts bounded attribute residuals for selected low-LoD anchors from uniformly quantized per-anchor tokens, leaving octree positions and topology unchanged; refinement is bypassed at the complete LoD. A 2D quadtree is shown for clarity; the method uses octrees in 3D.}
    \label{fig:overview}
\end{figure*}

\section{Proposed Method}\label{sec:methodology}

As shown in Fig.~\ref{fig:overview}, ProGS converts an anchor-based 3DGS scene into a sequence of strict octree prefixes. The scene is then optimized in two stages. In Stage~1, anchors initialized from the SfM point cloud form a parent-closed hierarchy with levels from $0$ to $L-1$. Each cumulative LoD prefix receives direct supervision and can serve as a complete rendering model. Level-balanced growth limits the additional parent-chain overhead caused by strict closure. For transmission, the structure of the current level is encoded first. A parent-causal context model then predicts the attribute prior and an adaptive quantization step, whose inputs include the parent prefix, the current-level octree structure, and a shared hash grid. Finally, the current-level attributes are quantized and arithmetic-coded.

After Stage~1 converges, Stage~2 freezes the octree, base attributes, and main codec and refines selected anchors in lv0--lv2. From the decoded attributes, LoD metadata, and compact per-anchor tokens, the refinement decoder predicts bounded residuals that adjust the render state without changing the octree topology. The complete LoD uses the decoded Stage~1 model directly.

\subsection{Strict Octree-Prefix Representation}\label{sec:strict_octree_prefix}

ProGS builds on the anchor-based representation of Scaffold-GS~\cite{lu2023scaffoldgsstructured3dgaussians}. Each anchor $\bm{a}_i$ is a position $\bm{x}_i$ carrying a compact attribute set
\begin{equation}
    \bm{\mathcal{A}}_i=\bigl\{\bm{f}_i\!\in\!\mathbb{R}^{D^f},\;\bm{s}_i\!\in\!\mathbb{R}^{D^s},\;\bm{o}_i\!\in\!\mathbb{R}^{K\times 3}\bigr\},
\end{equation}
where $\bm{f}_i$ is a $D^f$-dimensional feature, $\bm{s}_i$ holds $D^s$ scaling parameters, and the $K$ rows of $\bm{o}_i$ are 3D offsets that place $K$ generated Gaussians relative to the anchor center. A lightweight multi-layer perceptron (MLP), denoted by $\mathrm{MLP}_g$, maps the feature and view-dependent inputs to the remaining Gaussian attributes, including color and opacity. For an anchor subset $\mathcal{S}$ and camera $\pi$, the anchor-based rendering process is denoted by $\mathcal{R}(\mathcal{S};\pi)$.

Progressive transmission requires the representation hierarchy and the bitstream order to describe the same sequence of models. ProGS therefore constructs a forest of octrees in one global coordinate system over the input SfM point cloud $\bm{P}$. Specifically, ProGS estimates a cubic scene bound with origin $\bm{b}$ and side length $v_B$ that encloses $\bm{P}$ after a fixed boundary expansion. This bound is shared by all octree levels, rather than estimating a separate coordinate system for each LoD.

Given a base depth $l_b$ and $L$ levels indexed by $l\in\{0,\ldots,L-1\}$, the voxel size at level $l$ is
\begin{equation}
    v_l = \frac{v_B}{2^{l_b+l}},\label{eq:voxelsize}
\end{equation}
so that each successive level halves the voxel size along every spatial dimension. An anchor $\bm{a}_i^l$ at position $\bm{x}_i^l$ is assigned the integer voxel coordinate
\begin{equation}
    \bm{u}_i^l = \operatorname{round}\!\left(\frac{\bm{x}_i^l-\bm{b}}{v_l}\right)\in\mathbb{Z}^3,
    \qquad
    \bm{x}_i^l=\bm{b}+v_l\bm{u}_i^l.
    \label{eq:octree_grid_coordinate}
\end{equation}
The coordinate of its unique parent and its relative child offset are then determined by
\begin{equation}
    \bm{u}_{p(i)}^{l-1}=\left\lfloor\frac{\bm{u}_i^l}{2}\right\rfloor,
    \qquad
    \bm{r}_i^l=\bm{u}_i^l-2\bm{u}_{p(i)}^{l-1}\in\{0,1\}^3.
    \label{eq:octree_parent_coordinate}
\end{equation}
The three components of $\bm{r}_i^l$ specify one of eight octants and are mapped to the child index $4r_{i,x}^l+2r_{i,y}^l+r_{i,z}^l$. Hence, descendants can be located deterministically from a parent index and a 3-bit octant code.

To initialize the hierarchy, SfM points are first quantized at the finest level. The coarser levels are then constructed in a bottom-up manner by repeatedly mapping each occupied voxel to its parent via integer division of its coordinates. Let $\mathcal{V}_l$ denote the resulting anchor set at level $l$. The LoD-$l$ model is defined as the cumulative prefix
\begin{equation}
    \mathcal{P}_l=\bigcup_{k=0}^{l}\mathcal{V}_k.
    \label{eq:octree_prefix}
\end{equation}
ProGS enforces parent closure throughout initialization and adaptive structure updates:
\begin{equation}
    \bm{a}_i^l\in\mathcal{V}_l,\;l>0
    \quad\Longrightarrow\quad
    p(\bm{a}_i^l)\in\mathcal{V}_{l-1}.
    \label{eq:parent_closure}
\end{equation}
Therefore, every anchor in $\mathcal{P}_l$ has its complete encoded ancestor chain, up to a root in $\mathcal V_0$, in the same prefix. The bitstream follows this topology by coding $\mathcal{V}_0,\mathcal{V}_1,\ldots$ in order. The shared header and all complete chunks through level $l$ are called a \textit{completed level prefix}. Once this prefix has been decoded, all anchors in $\mathcal{P}_l$ are rendered jointly as $\mathcal{R}(\mathcal{P}_l;\pi)$ without accessing structure or attributes from future levels. The parent-closed prefix supplies the context used by the parent-causal decoder in Sec.~\ref{sec:parent_causal_entropy}.

\begin{figure}
    \centering
    \includegraphics[width=\linewidth]{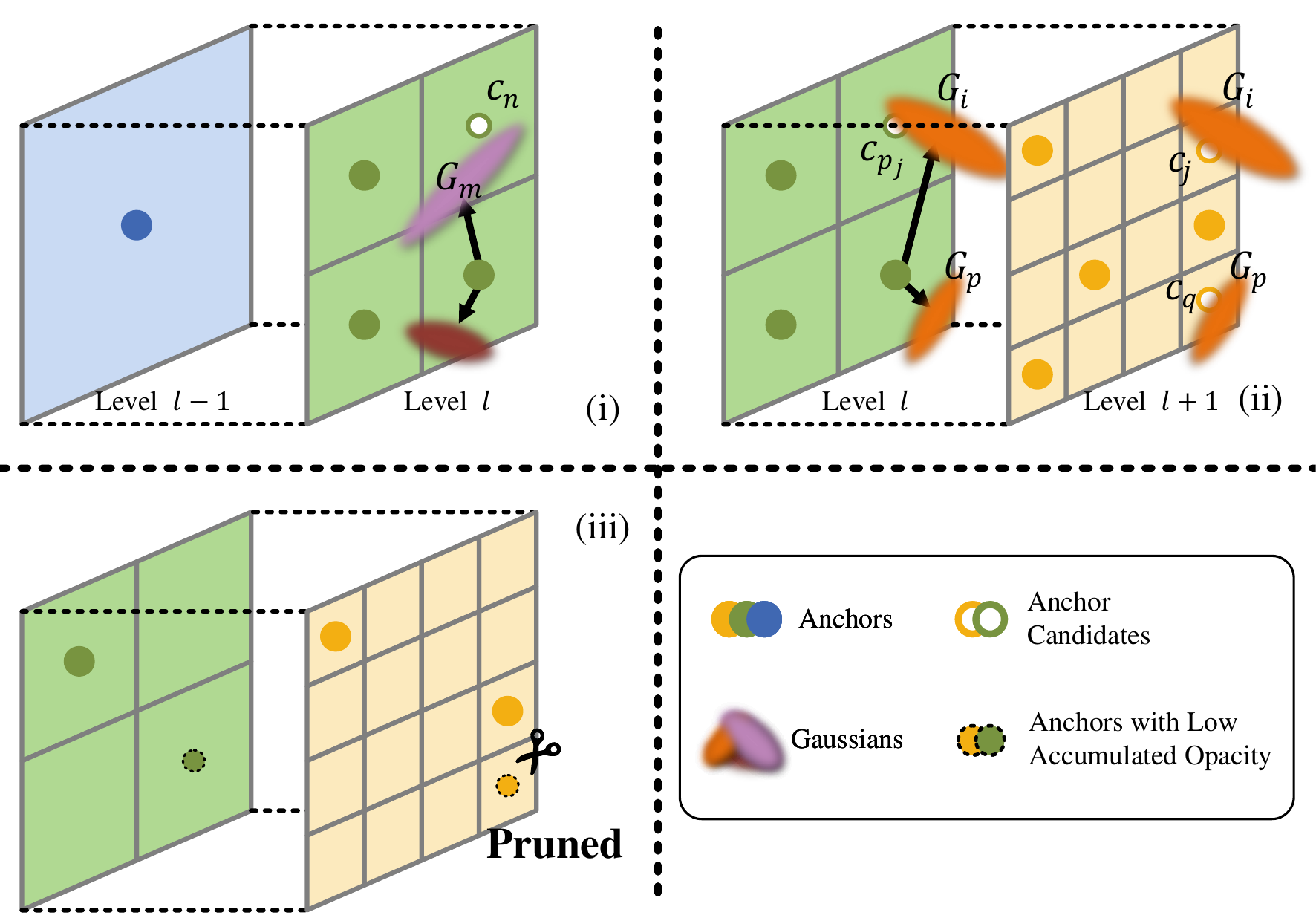}
    \caption{Geometric operations in level-balanced strict-octree growth (2D quadtree planes). \textbf{(i)} Same-level candidates are generated from Gaussian offsets. \textbf{(ii)} Finer-level candidates are quantized to child cells; a retained child without a parent queues the missing parent before materializing. \textbf{(iii)} Low-opacity anchors are pruned only as leaves, preserving parent closure. Prefix-aware scoring, closure penalties, and the top-$B_l$ budget are specified in Algo.~\ref{alg:algo}.}
    \label{fig:adaptive_adjustment}
\end{figure}

\subsection{Level-Balanced Adaptive Anchor Growth}\label{sec:aaa}

The strict prefix constraint changes the cost of adaptive structure growth. A new anchor at level $l$ contributes not only its own attributes but also any missing ancestors required by Eq.~\eqref{eq:parent_closure}. Moreover, a coarse-level anchor is included in more cumulative prefixes than a fine-level anchor. Directly applying gradient-only densification can therefore allocate a disproportionate amount of rate to early prefixes, especially in large-scale scenes. ProGS addresses this issue with prefix-aware candidate scoring, parent-chain closure, and a level-balanced growth budget.

\noindent\textbf{Prefix-aware candidate scoring.} Hierarchical densification determines whether a new anchor should be added and at which level it should be placed. This decision considers three factors: 1) the local reconstruction demand, 2) the utility to progressive LoD prefixes, and 3) the cumulative rate exposure. For the $j$-th Gaussian generated by anchor $\bm{a}_i^l$, let $\bar{\bm{s}}_i^l\in\mathbb{R}^3$ denote the decoded anchor-scale components used to modulate its offset. Its current center
\begin{equation}
    \bm{y}_{i,j}^l=\bm{x}_i^l+\bar{\bm{s}}_i^l\odot\bm{o}_{i,j}^l
    \label{eq:growth_candidate_position}
\end{equation}
is the spatial candidate used by densification. The differentiable renderer records the gradient of its projected center in screen space. The norm of the two image-plane components is accumulated over the latest $T$ iterations, and the observation-normalized average is denoted by $g_{i,j}^l$.
A candidate at a shallow level contributes to both the complete model and multiple low-LoD prefixes. Its prefix utility is therefore defined as
\begin{equation}
    q_l=\operatorname{clip}\!\left(
    \frac{l_\mathrm{low}-l+1}{l_\mathrm{low}+1},0,1
    \right),
    \qquad
    b_l=w_\mathrm{full}+w_\mathrm{low}q_l,
    \label{eq:growth_prefix_benefit}
\end{equation}
where levels $0$ through $l_\mathrm{low}$ form the low-LoD range. Within this range, $q_l$ decreases with depth and becomes zero at deeper levels. The weights $w_\mathrm{full}$ and $w_\mathrm{low}$ control the importance of the complete model and low-LoD prefixes, respectively. Since shallow-level candidates are included in more cumulative prefixes, they also have greater rate exposure. This exposure is approximated by
\begin{equation}
    \rho_l=\frac{b_l}{w_\mathrm{full}+w_\mathrm{low}}.
\end{equation}
These factors are combined into the growth score
\begin{equation}
    s_{i,j}^l=
    \frac{g_{i,j}^l b_l}{1+w_\mathrm{rate}\rho_l},
    \label{eq:growth_score}
\end{equation}
where $w_\mathrm{rate}$ controls the rate penalty. A high score indicates strong reconstruction demand and high progressive utility after accounting for rate exposure.
The level-dependent growth threshold is defined as $\tau_g^l=\tau_g2^{\beta l}$. A candidate is added at level $l$ if
$\tau_g^l<s_{i,j}^l\leq\tau_g^{l+1}$. If
$s_{i,j}^l>\tau_g^{l+1}$, it is assigned to level $l+1$. Candidates with
$s_{i,j}^l\leq\tau_g^l$ are not added. A same-level candidate is obtained by quantizing $\bm{y}_{i,j}^l$ with voxel width $v_l$, whereas a finer-level candidate uses $v_{l+1}$. Thus, the score selects a target level and Eq.~\eqref{eq:growth_candidate_position} selects its target cell; it does not create a Gaussian at an unconstrained continuous location.

\noindent\textbf{Parent-chain closure.} Strict parent closure requires every new anchor to have a complete ancestor chain. The growth decision therefore considers not only the candidate score, but also the cost of adding missing ancestors. Each candidate position is first quantized using Eq.~\eqref{eq:octree_grid_coordinate}, and its parent coordinate is then obtained from Eq.~\eqref{eq:octree_parent_coordinate}. If the parent is absent, the selection threshold is increased to
\begin{equation}
    \tau_\mathrm{cl}^l
    =
    \tau_g^l
    \left(
    1+w_\mathrm{cl}\mathbf{1}_{\mathrm{miss}}
    \right),
    \label{eq:closure_threshold}
\end{equation}
where $\mathbf{1}_{\mathrm{miss}}$ equals one when the parent is missing and zero otherwise. The weight $w_\mathrm{cl}$ controls the penalty for completing the missing ancestor chain. Thus, candidates with existing parents use the original threshold $\tau_g^l$, while candidates with missing parents must provide a larger benefit. For each selected candidate, the missing ancestor chain is traced from fine to coarse and added to the growth queue. Candidates that fall into occupied voxels are discarded. Multiple candidates quantized to the same empty target voxel are merged, retaining the maximum growth score; ancestors shared by several retained branches are likewise queued only once. This procedure keeps at most one anchor in each level-specific voxel and maintains the parent-closure constraint in Eq.~\eqref{eq:parent_closure}.

\noindent\textbf{Level-balanced budget.} Even with closure-aware scoring, too many valid branches may be added to coarse levels in a single update. Such growth can make early prefixes unnecessarily large. ProGS therefore applies a growth budget to each protected coarse level $l\leq l_\mathrm{bal}$:
\begin{equation}
    B_l=
    \max\!\left(
    B_\mathrm{min},
    \left\lceil r_\mathrm{bal}N_lK\right\rceil
    \right),
    \label{eq:level_balance_budget}
\end{equation}
where $N_l$ is the current number of anchors at level $l$, and each anchor generates $K$ Gaussians. The ratio $r_\mathrm{bal}$ controls the allowed growth relative to the current level size. The lower bound $B_\mathrm{min}$ prevents the budget from becoming too small in small scenes. If the number of direct and closure-induced candidates exceeds $B_l$, only the candidates with the highest growth scores in Eq.~\eqref{eq:growth_score} are retained. Their complete parent chains are then added to the hierarchy. This level-balanced policy limits the expansion of coarse prefixes while preserving the strict octree structure. When a retained voxel is materialized, a same-level anchor feature is initialized by elementwise maximum aggregation over the triggering anchor features, while a newly created finer-level anchor starts from a zero feature. In both cases, offsets start at zero and the scaling parameters start from the target voxel width; subsequent training optimizes these attributes normally. Algo.~\ref{alg:algo} summarizes the full growth procedure.

ProGS also prunes unimportant anchors after each growth interval. The accumulated opacity and observation count of every anchor are recorded during training. An anchor is removed only if it satisfies three conditions: it has been observed sufficiently often, its average accumulated opacity is below $\tau_o$, and it has no children. The last condition restricts pruning to leaf anchors and preserves parent closure in every remaining prefix.

\renewcommand{\algorithmicrequire}{\textbf{Input : }}
\renewcommand{\algorithmicensure}{\textbf{Output : }}
\begin{algorithm}[t]
    \caption{Level-balanced strict-octree growth.}
    \label{alg:algo}
    \begin{algorithmic}[1]
    \REQUIRE Level sets $\{\mathcal{V}_l\}_{l=0}^{L-1}$, accumulated gradients $\{g_{i,j}^l\}$, thresholds $\{\tau_g^l\}$, balance parameters $l_\mathrm{bal}$, $r_\mathrm{bal}$, and $B_\mathrm{min}$.
        \STATE Compute prefix-aware scores $s_{i,j}^l$ using Eq.~\eqref{eq:growth_score}.
        \STATE Generate same-level and finer-level candidates by thresholding $s_{i,j}^l$.
        \FOR{level $l=L-1\rightarrow0$}
            \STATE Quantize candidate positions to level-$l$ voxel coordinates.
            \STATE Remove candidates whose target voxels are already occupied.
            \STATE Apply the closure threshold in Eq.~\eqref{eq:closure_threshold}.
            \STATE Queue missing ancestors required by Eq.~\eqref{eq:parent_closure}.
            \IF{$l\leq l_\mathrm{bal}$ and the candidate count exceeds $B_l$}
                \STATE Retain the top-$B_l$ branches according to $s_{i,j}^l$.
            \ENDIF
        \ENDFOR
        \STATE Materialize retained branches and their complete parent chains.
        \STATE Prune low-opacity leaf anchors while preserving internal ancestors.
    \end{algorithmic}
\end{algorithm}

\subsection{Multi-LoD Prefix Training}\label{sec:multi_lod_training}

Optimizing only the full hierarchy does not ensure that an intermediate prefix is a useful scene representation. ProGS therefore applies direct image supervision to the cumulative prefix $\mathcal{P}_l$ defined in Eq.~\eqref{eq:octree_prefix}. For a training camera $\pi$ with ground-truth image $\bm{I}$, the level-$l$ reconstruction and distortion are
\begin{equation}
    \begin{aligned}
        \hat{\bm{I}}_l&=\mathcal{R}(\mathcal{P}_l;\pi),\\
        \mathcal{D}_l&=(1-\lambda_\mathrm{s})\|\hat{\bm{I}}_l-\bm{I}\|_1  +\lambda_\mathrm{s}\left(1-\operatorname{SSIM}(\hat{\bm{I}}_l,\bm{I})\right),
    \end{aligned}
    \label{eq:prefix_distortion}
\end{equation}
where SSIM denotes the structural similarity index~\cite{wang2004image}, and $\lambda_\mathrm{s}$ balances photometric and structural similarity. During the multi-LoD stage, the full model and all lower prefixes are jointly optimized using
\begin{equation}
    \mathcal{L}_\mathrm{prefix}(t)=
    \frac{\mathcal{D}_{L-1}+\gamma_t\sum_{l=0}^{L-2}\mathcal{D}_l}
    {1+\gamma_t(L-1)},
    \label{eq:prefix_training_loss}
\end{equation}
For the default schedule,
\begin{equation}
    \gamma_t=
    \begin{cases}
        1, & 1{,}500<t\leq25{,}000,\\
        0, & \text{otherwise}.
    \end{cases}
    \label{eq:prefix_weight_schedule}
\end{equation}
Thus, training uses the full-model path alone during the initial 1{,}500-iteration warm-up and after the hierarchy-adaptation phase. The sensitivity study also evaluates schedules with $\gamma_t\in\{0.05,0.15,1\}$ for $25{,}000<t\leq40{,}000$. This schedule makes low LoDs explicit optimization targets while reserving the final stage for full-LoD convergence. Equation~\eqref{eq:prefix_training_loss} directly optimizes the rendering quality of each decoded prefix.

\subsection{Parent-Causal Entropy Coding}\label{sec:parent_causal_entropy}

The octree order provides a causal source of entropy context. Before decoding the attributes at level $l$, the decoder has already reconstructed $\hat{\mathcal{P}}_{l-1}$ and the structure of $\mathcal{V}_l$. Let $\mathcal{O}_l$ denote this decoded current-level structure. It contains the parent index and octant code of every non-root anchor; grouping these nodes by parent also determines the occupied-child vector. Thus, $\mathcal{O}_l$ reveals the current anchor positions and parent--child topology, but none of the current attributes. ProGS factorizes the attribute distribution accordingly:
\begin{equation}
    p(\hat{\bm{\mathcal{A}}}\mid\mathcal{H})=
    \prod_{l=0}^{L-1}\prod_{\bm{a}_i^l\in\mathcal{V}_l}
    p\!\left(\hat{\bm{\mathcal{A}}}_i^l\mid
    \mathcal{H},\hat{\mathcal{P}}_{l-1},\mathcal{O}_l\right),
    \label{eq:causal_factorization}
\end{equation}
where $\hat{\mathcal{P}}_{-1}=\emptyset$. Following the hash-assisted model in HAC~\cite{chen2024hachashgridassistedcontext}, a mixed 3D--2D hash-grid is queried at the current and parent positions. For a non-root anchor, the complete context is
\begin{equation}
    \bm{c}_i^l=\operatorname{concat}\!\left(
    \bm{h}(\bm{x}_i^l),\bm{h}(\bm{x}_{p(i)}^{l-1}),
    \hat{\bm{f}}_{p(i)},\bm{e}_{p(i)}^l\right),
    \label{eq:parent_causal_context}
\end{equation}
where $\hat{\bm{f}}_{p(i)}$ is the decoded parent feature. The structural descriptor $\bm{e}_{p(i)}^l$ contains the normalized parent level, summaries of its decoded scaling and offsets, and its child-occupancy vector in $\mathcal{O}_l$. All these quantities are available before decoding $\hat{\bm{\mathcal{A}}}_i^l$. For root anchors, the parent-dependent entries are set to zero. No context is obtained from levels higher than $l$.

The trainable hash-grid $\mathcal{H}$ is binarized to $\{-1,+1\}$ using straight-through estimation (STE)~\cite{shin2024binary,chen2024far}. Its bit cost is estimated by the Bernoulli cross-entropy
\begin{equation}
    \mathcal{L}_\text{hash}=M_+\!\left(-\log_2 f_+\right)+M_-\!\left(-\log_2(1-f_+)\right),
\end{equation}
where $f_+$ is the empirical frequency of ``$+1$'', and $M_+$ and $M_-$ count the two binary symbols. The grid is transmitted once in the bitstream header.

The context $\bm{c}_i^l$ predicts a Gaussian prior and adaptive quantization step for feature, scaling, and offset attributes:
\begin{equation}
    \begin{aligned}
        \bm{\mu}_i^l,\bm{\sigma}_i^l&=\operatorname{MLP}_d(\bm{c}_i^l),\\
        \bm{q}_i^l&=\bm{Q}_0\odot\left(1+\tanh(\operatorname{MLP}_q(\bm{c}_i^l))\right).
    \end{aligned}
    \label{eq:causal_prior}
\end{equation}
The context model takes decoded state as input; the current unquantized attribute $\bm{\mathcal A}_i^l$ is quantized using $\bm q_i^l$, while $(\bm\mu_i^l,\bm\sigma_i^l)$ and $\bm q_i^l$ jointly determine the probability of the resulting symbol. The position $\bm x_i^l$ is reconstructed from the octree structure and used for the hash-grid query.
For a scalar quantized symbol $\hat{a}_{i,j}^l$, its probability mass is the integral over the corresponding quantization bin:
\begin{equation}
    p_{i,j}^l=
    \Phi_{i,j}^l\!\left(\hat{a}_{i,j}^l+\tfrac{1}{2}q_{i,j}^l\right)
    -\Phi_{i,j}^l\!\left(\hat{a}_{i,j}^l-\tfrac{1}{2}q_{i,j}^l\right),
    \label{eq:quantized_probability}
\end{equation}
where $\Phi_{i,j}^l$ is the cumulative distribution function (CDF) of $\mathcal{N}(\mu_{i,j}^l,\sigma_{i,j}^l)$. Arithmetic encoding (AE)~\cite{rissanen1979arithmetic} uses these probabilities, and training minimizes
\begin{equation}
    \mathcal{L}_\mathrm{entropy}=
    \sum_{l=0}^{L-1}\sum_{\bm{a}_i^l\in\mathcal{V}_l}
    \sum_{j=1}^{D^f+D^s+3K}-\log_2 p_{i,j}^l.
    \label{eq:entropy_loss}
\end{equation}
Because Eq.~\eqref{eq:parent_causal_context} uses decoded parent quantities, the encoder and decoder construct identical contexts in the same coarse-to-fine order.

\subsection{Low-LoD Token Refinement}\label{sec:parent_anchor_refinement}

Even with direct prefix supervision, an early prefix has limited spatial capacity because future child anchors are unavailable. ProGS therefore freezes the converged Stage~1 codec and refines only the render state of anchors already decoded in $\hat{\mathcal{P}}_l$. Let $\hat{\bm{\mathcal A}}_i=\{\hat{\bm f}_i,\hat{\bm s}_i,\hat{\bm o}_i\}$ denote the frozen decoded attributes of a selected anchor and let $\bm m_i^l$ denote its LoD metadata, including its normalized anchor level, target prefix, and child occupancy visible within $\hat{\mathcal P}_l$. These quantities are deterministically available at both encoder and decoder. The only optimized per-anchor side information is a token $\bm z_i\in\mathbb R^{D^\mathrm z}$, which is transmitted once for a selected subset of low-LoD anchors. The same token is reused across the refined prefixes; the target-prefix entry in $\bm m_i^l$ allows the decoder to predict a different residual for each $l$. A lightweight refinement decoder predicts the tuple
\begin{equation}
\begin{aligned}
    \Delta\bm{\Theta}_i^l
    &=
    \{\Delta\bm f_i^l,\Delta\bm s_i^l,
    \Delta\bm o_i^l,\Delta\bm\alpha_i^l\}\\
    &=
    \bm d\odot\tanh\!\left(
    \operatorname{MLP}_r(
    \hat{\bm{\mathcal A}}_i,\bm m_i^l,\bm z_i)
    \right),
    \label{eq:parent_refinement}
\end{aligned}
\end{equation}
where $\bm{d}$ contains component-specific bounds and $\Delta\bm\alpha_i^l\in\mathbb R^K$ is a residual for the $K$ opacity values produced by the frozen opacity head. For an unselected anchor, every residual is zero. For target prefix $l$, residuals are applied only to anchors already present in $\hat{\mathcal P}_l$. The refined coded attributes are
\begin{equation}
    \widetilde{\bm{\mathcal A}}_i^l
    =\{\hat{\bm f}_i+\Delta\bm f_i^l,\,
       \hat{\bm s}_i+\Delta\bm s_i^l,\,
       \hat{\bm o}_i+\Delta\bm o_i^l\}.
    \label{eq:refined_attributes}
\end{equation}
For camera $\pi$, the frozen opacity head predicts $K$ base opacity values from the refined anchor state $\widetilde{\bm{\mathcal A}}_i^l$. The renderer uses the residual-corrected opacity
\begin{equation}
    \widetilde{\bm\alpha}_i^l(\pi)
    = \bm\alpha_i(\widetilde{\bm{\mathcal A}}_i^l;\pi)
    + \Delta\bm\alpha_i^l.
    \label{eq:refined_opacity}
\end{equation}
This rendering-time correction lies outside the arithmetic-coded attribute tuple $\bm{\mathcal A}_i$. It preserves the decoded node positions, levels, and parent--child relations and uses the attributes in the decoded prefix. Let $\widetilde{\mathcal P}_l$ denote the resulting level-$l$ representation, which consists of the decoded octree prefix and the refined render-time states. At the full LoD, the fully decoded base hierarchy is rendered directly.

\begin{figure*}[hbtp]
    \centering
    \includegraphics[width=\linewidth]{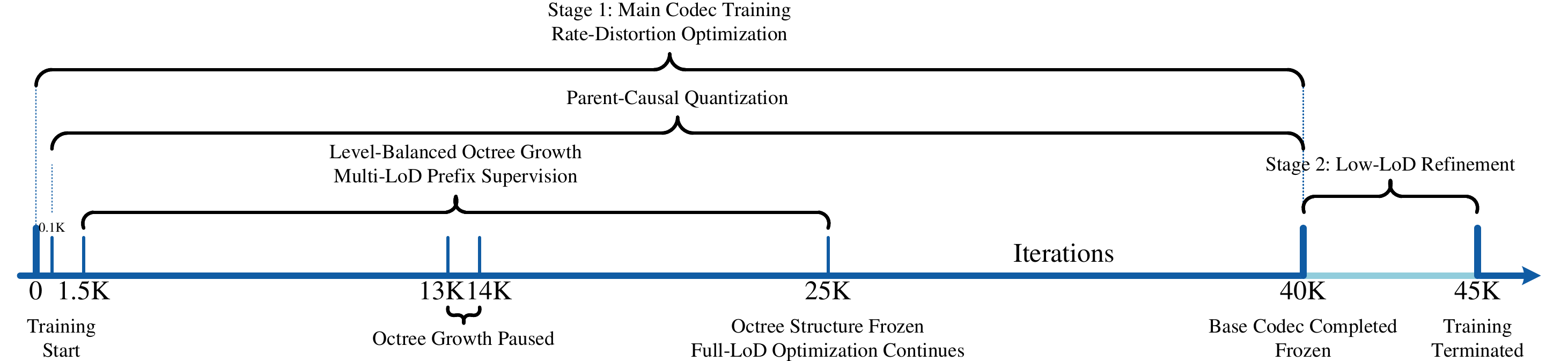}
    \caption{Training schedule of ProGS. Stage~1 optimizes the main codec for 40{,}000 iterations: the parent-causal entropy context is enabled at iteration 100, and level-balanced growth with multi-LoD supervision runs from iteration 1{,}500 to 25{,}000, with growth alone paused from 13{,}000 to 14{,}000 per the HAC-style densification schedule. From iteration 25{,}000 the octree is frozen and only the full-LoD path is optimized. Stage~2 then freezes the base codec and refines low LoDs for 5{,}000 iterations.}
    \label{fig:trainingprocess}
\end{figure*}

The complete two-stage optimization schedule is summarized in Fig.~\ref{fig:trainingprocess}.

\subsection{Training Objectives and Bitstream Syntax}\label{sec:training_and_coding}

The main codec optimizes prefix distortion and Gaussian compactness together with an average coded-attribute entropy regularizer:
\begin{equation}
    \begin{aligned}
        \mathcal{L}_\mathrm{main}
        &=\mathcal{L}_\mathrm{prefix}
        +\lambda_\mathrm{scale}\mathcal{L}_\mathrm{scale}
        +\lambda_\mathrm{e}\mathcal{L}_\mathrm{e},\\
        \mathcal{L}_\mathrm{e}
        &=\frac{\mathcal{L}_\mathrm{entropy}+\mathcal{L}_\mathrm{hash}}
        {N(D^f+D^s+3K)}.
    \end{aligned}
    \label{eq:loss}
\end{equation}
Here, $N=\sum_{l=0}^{L-1}|\mathcal V_l|$ is the total number of anchors, and $D^f+D^s+3K$ is the number of coded attribute scalars per anchor. The term $\mathcal{L}_\mathrm{scale}$ is the mean product of the generated Gaussian scales and discourages oversized primitives. The normalized term $\mathcal L_\mathrm e$ is the modeled attribute and hash-grid cost averaged over coded attribute scalars. Varying $\lambda_\mathrm{e}$ produces different rate--distortion operating points.

After the main codec is frozen, the refinement decoder and tokens are optimized on selected low LoDs. Let $\bm{I}_{L-1}$ be the frozen full-LoD rendering. The target for level $l$ mixes the ground truth and full-model teacher as $\bm{T}_l=\eta_l\bm{I}_{L-1}+(1-\eta_l)\bm{I}$. The refinement appearance loss is
\begin{equation}
    \begin{aligned}
        \widetilde{\bm{I}}_l&=\mathcal{R}(\widetilde{\mathcal{P}}_l;\pi),\\
        \mathcal{D}_\mathrm{ref}(\bm{A},\bm{B})
        &=\|\bm{A}-\bm{B}\|_1
        +\lambda_\mathrm{rs}\bigl(1-\operatorname{SSIM}(\bm{A},\bm{B})\bigr),\\
        \mathcal{L}_\mathrm{ref}
        &=\mathbb{E}_{l,\pi}\!\left[
        \mathcal{D}_\mathrm{ref}(\widetilde{\bm{I}}_l,\bm{T}_l)\right].
    \end{aligned}
    \label{eq:refinement_loss}
\end{equation}
The implementation regularizes the token magnitude during training:
\begin{equation}
    \begin{aligned}
        \mathcal{L}_\mathrm{tok}
        &=\frac{1}{|\mathcal S_\mathrm{ref}|D^\mathrm z}
        \sum_{i\in\mathcal S_\mathrm{ref}}
        \sum_{k=1}^{D^\mathrm{z}}
        |z_{i,k}|,\\
        \mathcal{L}_\mathrm{stage2}
        &=\mathcal{L}_\mathrm{ref}
        +\lambda_\mathrm{z}\mathcal{L}_\mathrm{tok},
    \end{aligned}
    \label{eq:refinement_rate_loss}
\end{equation}
where $\mathcal S_\mathrm{ref}$ is the set of anchors carrying tokens. The teacher component encourages consistency with the full model, while the ground-truth component prevents refinement from merely reproducing its errors. After training, each token dimension is uniformly quantized to $b$ bits using its transmitted minimum and maximum. Consequently, the token payload is
\begin{equation}
    R_\mathrm{tok}
    =bD^\mathrm z|\mathcal S_\mathrm{ref}|+32D^\mathrm z
    \quad\text{bits},
    \label{eq:token_payload}
\end{equation}
where the second term stores one 16-bit minimum and maximum per dimension. The token-bearing anchors are identified by one binary mask for each refined level. Let $M_l$ be the number of eligible anchors in level $l$ and $L_\mathrm{ref}=3$. The masks are concatenated and bit-packed, while one 32-bit length is stored for each level, giving
\begin{equation}
    R_\mathrm{sel}
    =8\left\lceil\frac{\sum_{l=0}^{L_\mathrm{ref}-1}M_l}{8}\right\rceil
    +32L_\mathrm{ref}
    \quad\text{bits}.
    \label{eq:selection_payload}
\end{equation}
Let $P_\mathrm{ref}$ be the number of learned parameters in the refinement decoder. These weights are stored in 32-bit floating point, so the complete refinement side stream has rate
\begin{equation}
    R_\mathrm{ref}=R_\mathrm{tok}+R_\mathrm{sel}+32P_\mathrm{ref}
    \quad\text{bits},
    \label{eq:refinement_payload}
\end{equation}
which is a fixed-rate side stream.

\begin{table*}[!t]
  \centering
  \small
  \caption{Rate organization of the compared methods. Independent rate points require separate optimization. Octree-GS-EC is the entropy-coded adaptation of Octree-GS used in this comparison. Here, a coded prefix denotes a serialized cumulative bitstream prefix; N/A for official Octree-GS indicates that it reports renderable LoDs rather than coded prefixes. N/R: property not uniformly reported within a group.}
  \label{tab:method_properties}
  \setlength\tabcolsep{0.18cm}
  \begin{tabular}{@{}llcccl@{}}
    \toprule
    Method / category & Rate organization & \shortstack{Cumulative\\chunks} & \shortstack{Coded-prefix\\renderable} & \shortstack{Parent-closed\\octree} & \shortstack{Decoding\\dependency} \\
    \midrule
    Reference representations & Single endpoint & No & No & No & -- \\
    General single-rate compression & Independent models & No & No & N/R & Method-specific \\
    Entropy-coded single-rate compression & Independent models & No & No & No & Non-progressive \\
    Octree-GS (official)~\cite{ren2024octreegsconsistentrealtimerendering} & Adaptive LoDs & No & N/A & No & View-dependent LoD \\
    Octree-GS-EC & Cumulative LoD groups & Yes & Yes & No & Decoded previous LoD \\
    PCGS~\cite{chen2026pcgs} & Cumulative levels & Yes & Yes & No & Previous layers \\
    ProGS & Cumulative octree levels & Yes & Yes & Yes & Decoded parents \\
    \bottomrule
  \end{tabular}
\end{table*}

\begin{table*}[!t]
  \centering
  \scriptsize
  \caption{Main quantitative comparison. Top three entries per column are \textbf{bold}, \underline{underlined}, and \textit{italic}; ties share a rank. Octree-GS-EC denotes the entropy-coded adaptation of Octree-GS used in this comparison. For Octree-GS-EC and ProGS, LR/HR use $\lambda=0.004/0.0005$; full denotes the highest available LoD. Official Octree-GS and the reference representations report checkpoint storage; PCGS, Octree-GS-EC, and ProGS report cumulative coded payloads; the remaining compression methods report complete-model storage.}
  \label{tab:main_results}
  \setlength\tabcolsep{0.11cm}
  \resizebox{\textwidth}{!}{%
  \begin{tabular}{@{}lcccccccccccc@{}}
    \toprule
    & \multicolumn{4}{c}{Mip-NeRF360~\cite{barron2022mip}} & \multicolumn{4}{c}{BungeeNeRF~\cite{xiangli2022bungeenerf}} & \multicolumn{4}{c}{Tanks and Temples~\cite{knapitsch2017tanks}} \\
    Method / point & PSNR$\uparrow$ & SSIM$\uparrow$ & LPIPS$\downarrow$ & MB$\downarrow$ & PSNR$\uparrow$ & SSIM$\uparrow$ & LPIPS$\downarrow$ & MB$\downarrow$ & PSNR$\uparrow$ & SSIM$\uparrow$ & LPIPS$\downarrow$ & MB$\downarrow$ \\
    \midrule
    \multicolumn{13}{@{}l}{\textit{Reference representations}} \\
    3DGS~\cite{kerbl3Dgaussians} & 27.46 & \methodThirdRank{.812} & \methodThirdRank{.222} & 750.90 & 24.87 & .841 & .205 & 1616.00 & 23.69 & .844 & .178 & 431.00 \\
    Scaffold-GS~\cite{lu2023scaffoldgsstructured3dgaussians} & 27.50 & .806 & .252 & 253.90 & 26.62 & .865 & .241 & 183.00 & 23.96 & .853 & \methodThirdRank{.177} & 86.50 \\
    \midrule
    \multicolumn{13}{@{}l}{\textit{General single-rate compression}} \\
    SOG~\cite{morgenstern2023compact} & 26.56 & .791 & .241 & 16.70 & 22.43 & .708 & .339 & 48.25 & 23.15 & .828 & .198 & 9.30 \\
    RDO-Gaussian~\cite{wang2024rdogaussian} & 27.05 & .802 & .239 & 23.46 & 23.37 & .762 & .286 & 39.06 & 23.34 & .835 & .195 & 12.03 \\
    EAGLES~\cite{girish2024eaglesefficientaccelerated3d} & 27.14 & .809 & .231 & 58.91 & 25.89 & .865 & .197 & 115.20 & 23.28 & .835 & .203 & 28.99 \\
    CompGS (high)~\cite{liu2024compgsefficient3dscene} & 27.26 & .803 & .239 & 16.50 & -- & -- & -- & -- & 23.70 & .837 & .208 & 9.60 \\
    R-3DGS~\cite{papantonakis2024reducing} & 27.19 & .807 & .230 & 29.54 & 24.57 & .812 & .228 & 65.39 & 23.57 & .840 & .188 & 14.00 \\
    LightGaussian~\cite{fan2024lightgaussianunbounded3dgaussian} & 27.00 & .799 & .249 & 44.54 & 24.52 & .825 & .255 & 87.28 & 22.83 & .822 & .242 & 22.43 \\
    Compact3D~\cite{navaneet2024compact3dsmallerfastergaussian} & 27.12 & .806 & .240 & 19.33 & 24.70 & .815 & .266 & 33.39 & 23.44 & .838 & .198 & 12.50 \\
    Compressed3D~\cite{niedermayr2024compressed} & 26.98 & .801 & .238 & 28.80 & 24.13 & .802 & .245 & 55.79 & 23.32 & .832 & .194 & 17.28 \\
    \midrule
    \multicolumn{13}{@{}l}{\textit{Learned entropy-coded single-rate compression}} \\
    HAC (low)~\cite{chen2024hachashgridassistedcontext} & 27.53 & .807 & .238 & 15.26 & 26.48 & .845 & .250 & 18.49 & 24.04 & .846 & .187 & 8.10 \\
    HAC (high)~\cite{chen2024hachashgridassistedcontext} & 27.77 & .811 & .230 & 21.87 & 27.08 & .872 & .209 & 29.72 & 24.40 & .853 & \methodThirdRank{.177} & 11.24 \\
    HAC++ (low)~\cite{chen2025hacplus} & 27.60 & .803 & .253 & \methodThirdRank{8.34} & 26.78 & .858 & .235 & \methodSecondRank{11.75} & 24.22 & .849 & .190 & 5.18 \\
    HAC++ (high)~\cite{chen2025hacplus} & \methodSecondRank{27.82} & .811 & .231 & 18.48 & 27.17 & .879 & .196 & 20.82 & 24.32 & .854 & .178 & 8.63 \\
    CAT-3DGS (low)~\cite{zhan2025cat3dgs} & 25.82 & .730 & .362 & \methodFirstRank{1.72} & 25.19 & .808 & .279 & \methodFirstRank{10.14} & 22.97 & .786 & .293 & \methodFirstRank{1.42} \\
    CAT-3DGS (high)~\cite{zhan2025cat3dgs} & 27.77 & .809 & .241 & 12.35 & 27.35 & \methodThirdRank{.886} & .183 & 26.59 & 24.41 & .853 & .189 & 6.93 \\
    \midrule
    \multicolumn{13}{@{}l}{\textit{LoD representation}} \\
    Octree-GS (official)~\cite{ren2024octreegsconsistentrealtimerendering} & 27.45 & .808 & .226 & 120.28 & \methodFirstRank{28.10} & \methodFirstRank{.919} & \methodSecondRank{.093} & 317.75 & 22.59 & .787 & .233 & 77.94 \\
    \midrule
    \multicolumn{13}{@{}l}{\textit{Adapted entropy-coded LoD baseline}} \\
    Octree-GS-EC-LR (full) & 27.28 & .808 & .234 & 11.41 & 26.03 & .880 & .152 & 14.90 & 23.47 & .843 & .184 & 12.13 \\
    Octree-GS-EC-HR (full) & 27.21 & .811 & .228 & 18.61 & 20.19 & .736 & .232 & 38.14 & 22.57 & .836 & .190 & 17.75 \\
    \midrule
    \multicolumn{13}{@{}l}{\textit{Progressive compression}} \\
    PCGS (first)~\cite{chen2026pcgs} & \methodThirdRank{27.81} & .808 & .240 & 12.19 & 26.89 & .864 & .223 & \methodThirdRank{12.97} & \methodSecondRank{24.50} & .852 & .186 & 5.67 \\
    PCGS (full)~\cite{chen2026pcgs} & \methodFirstRank{27.96} & .811 & .236 & 25.34 & 27.15 & .872 & .213 & 22.82 & \methodFirstRank{24.59} & \methodSecondRank{.856} & .181 & 9.87 \\
    \midrule
    \multicolumn{13}{@{}l}{\textit{ProGS}} \\
    ProGS-LR (lv0) & 24.43 & .676 & .385 & \methodSecondRank{7.54} & 25.43 & .855 & .164 & 18.14 & 19.05 & .625 & .467 & \methodSecondRank{3.00} \\
    ProGS-LR (full) & 27.29 & \methodSecondRank{.813} & \methodSecondRank{.219} & 20.84 & \methodThirdRank{27.54} & \methodSecondRank{.913} & \methodThirdRank{.102} & 27.97 & 24.24 & \methodThirdRank{.855} & \methodSecondRank{.167} & 10.56 \\
    ProGS-HR (lv0) & 24.14 & .668 & .392 & 10.36 & 24.91 & .842 & .173 & 25.88 & 18.44 & .605 & .483 & \methodThirdRank{4.27} \\
    ProGS-HR (full) & 27.57 & \methodFirstRank{.818} & \methodFirstRank{.210} & 32.73 & \methodSecondRank{27.91} & \methodFirstRank{.919} & \methodFirstRank{.091} & 43.45 & \methodThirdRank{24.49} & \methodFirstRank{.866} & \methodFirstRank{.150} & 19.56 \\
    \bottomrule
  \end{tabular}}
\end{table*}

The encoded representation consists of a shared header $\mathcal{B}_H$ followed by level chunks $\mathcal{B}_0,\ldots,\mathcal{B}_{L-1}$. The header contains the global octree parameters, binary hash-grid, and Stage~1 decoder parameters. Chunk $\mathcal{B}_l$ contains the level-$l$ structure and the arithmetic-coded feature, scaling, and offset streams. In the coded-payload accounting, each root coordinate is assigned three 16-bit values. At subsequent levels, a node location is assigned a 20-bit index of its parent in the previous level and the 3-bit octant code defined after Eq.~\eqref{eq:octree_parent_coordinate}. The base-only profile through level $l$ is

\begin{equation}
    \mathcal{B}_{\leq l}^{\mathrm{base}}
    =\mathcal{B}_H\,\|\,\mathcal{B}_0\,\|\cdots\|\,\mathcal{B}_l.
    \label{eq:progressive_bitstream}
\end{equation}
The refined progressive profile inserts one side stream $\mathcal B_\mathrm{ref}$, containing the 32-bit refinement-decoder weights and the uniformly quantized tokens, after $\mathcal B_0$:
\begin{equation}
    \mathcal{B}_{\leq l}^{\mathrm{ref}}
    =\mathcal{B}_H\,\|\,\mathcal{B}_0\,\|\,\mathcal{B}_\mathrm{ref}
    \,\|\,\mathcal{B}_1\,\|\cdots\|\,\mathcal{B}_l,
    \qquad l\geq0,
    \label{eq:refined_progressive_bitstream}
\end{equation}
where the tail beginning with $\mathcal B_1$ is empty for $l=0$. The same $\mathcal B_\mathrm{ref}$ refines lv0--lv2 and remains part of the cumulative transfer cost at later prefixes. The complete-model renderer uses the base hierarchy. Clients may select the base-only profile in Eq.~\eqref{eq:progressive_bitstream} when early-prefix refinement is unnecessary.

Base decoding proceeds from root to leaves. At each level, ProGS first reconstructs the structure, then builds Eq.~\eqref{eq:parent_causal_context} from the previously decoded prefix, and finally entropy-decodes the current attributes. For a refined low LoD, the decoder additionally reads the corresponding level's selection-mask segment, binds the dequantized tokens to its set bits, loads the refinement weights, evaluates Eq.~\eqref{eq:parent_refinement} from the frozen decoded attributes and LoD metadata, and assembles Eq.~\eqref{eq:refined_attributes} without changing the octree. Each completed level is therefore incrementally decodable and directly renderable.

\definecolor{goldPro}{RGB}{255,215,0}
\definecolor{silverPro}{RGB}{191,193,194}
\definecolor{colorProGS}{RGB}{57,81,162}
\newcommand{\firstRank}[1]{\textbf{#1}}
\newcommand{\secondRank}[1]{\underline{#1}}
\newcommand{\thirdRank}[1]{\textit{#1}}
\newcommand{\qualimage}[2][0.09\textwidth]{%
  \raisebox{-0.5\height}{\includegraphics[width=#1]{#2}}}
\newcommand{\qualdetail}[3][0.115\textwidth]{%
  \begin{tikzpicture}[baseline=(img.center)]
    \node[draw=red,line width=0.6pt,inner sep=0.7pt,outer sep=0pt] (img)
      {\includegraphics[width=#1,trim=#2,clip]{#3}};
  \end{tikzpicture}}
\newcommand{\qualcrop}[3][0.09\textwidth]{%
  \raisebox{-0.5\height}{%
    \includegraphics[width=#1,trim=#2,clip]{#3}}}
\newcommand{\qualmetriccrop}[5][0.115\textwidth]{%
  \begin{tikzpicture}[baseline=(img.center)]
    \node[inner sep=0pt,outer sep=0pt] (img)
      {\includegraphics[width=#1,trim=#2,clip]{#3}};
    \node[anchor=south west,fill=black!65,text=white,font=\tiny,
      inner xsep=1.5pt,inner ysep=0.6pt]
      at (img.south west) {\shortstack{#4\,dB\\#5\,MB}};
  \end{tikzpicture}}
\newcommand{\qualcontext}[6][0.115\textwidth]{%
  \begin{tikzpicture}[baseline=(img.center)]
    \node[inner sep=0pt,outer sep=0pt] (img) {\includegraphics[width=#1]{#2}};
    \coordinate (sw) at (img.south west);
    \coordinate (se) at (img.south east);
    \coordinate (nw) at (img.north west);
    \coordinate (ne) at (img.north east);
    \coordinate (ybL) at ($(sw)!#5!(nw)$);
    \coordinate (ybR) at ($(se)!#5!(ne)$);
    \coordinate (ytL) at ($(sw)!#6!(nw)$);
    \coordinate (ytR) at ($(se)!#6!(ne)$);
    \coordinate (cropSW) at ($(ybL)!#3!(ybR)$);
    \coordinate (cropSE) at ($(ybL)!#4!(ybR)$);
    \coordinate (cropNW) at ($(ytL)!#3!(ytR)$);
    \coordinate (cropNE) at ($(ytL)!#4!(ytR)$);
    \draw[red,line width=0.6pt]
      (cropSW) -- (cropSE) -- (cropNE) -- (cropNW) -- cycle;
  \end{tikzpicture}}
\newcommand{\qualrowlabel}[2]{%
  \rotatebox[origin=c]{90}{%
    \shortstack{\textsc{#1}\\[-0.2ex]\textit{#2}}}}
\newcommand{\qualscenelabel}[1]{%
  \begin{tikzpicture}[baseline=(label.center)]
    \node[inner sep=0pt,rotate=90,font=\itshape] (label) {#1};
  \end{tikzpicture}}
\newcommand{\qualhead}[1]{%
  \begin{tikzpicture}[baseline=(head.center)]
    \node[inner sep=0pt,minimum height=2.4\baselineskip,
      align=center,font=\bfseries] (head) {#1};
  \end{tikzpicture}}
\section{Experiments}\label{sec:experiments}

\begin{figure*}[!t]
    \centering
    \includegraphics[width=\textwidth]{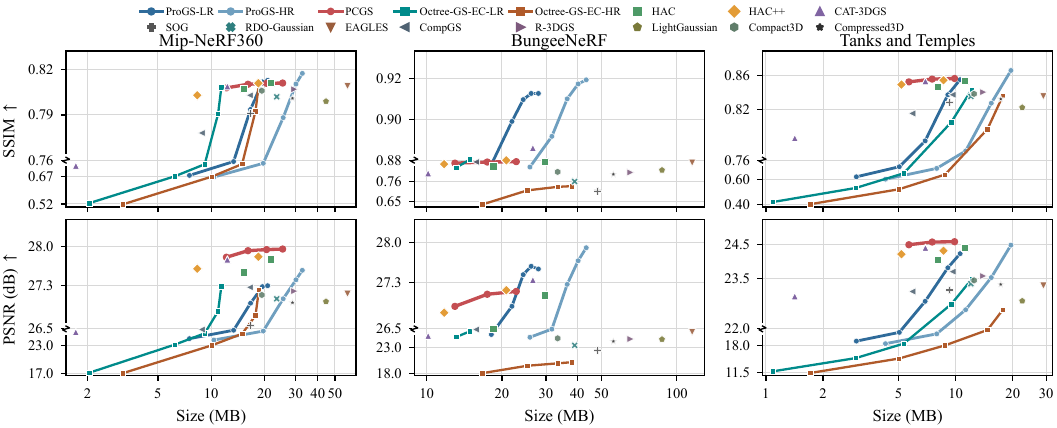}
    \caption{Dataset-average SSIM--rate (top) and PSNR--rate (bottom). The two blue ProGS curves are the lv0--lv4 prefixes of fixed-$\lambda$ LR and HR models. Solid lines connect only successive points of one cumulative progressive stream (ProGS, PCGS, and Octree-GS-EC); isolated markers are separately optimized single-rate models. Scenes with fewer occupied Octree-GS-EC levels retain their final LoD. Piecewise-linear Y-axes compress the sparse low-quality range; a kink on each left spine marks the scale transition, and ticks remain in the original units.}
    \label{fig:rd}
\end{figure*}

\begin{figure*}[t]
    \centering
    \includegraphics[width=\textwidth]{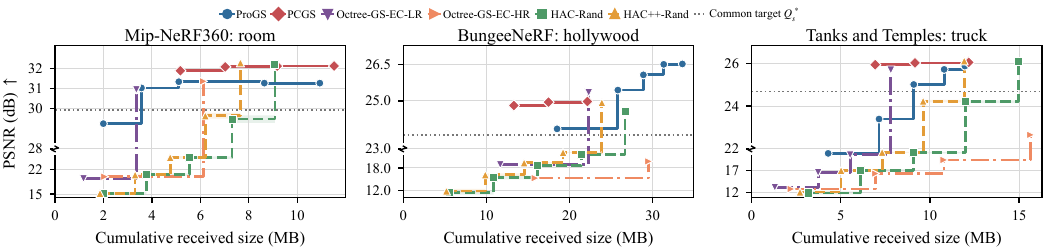}
    \caption{Progressive PSNR versus cumulative received size on matched scenes. ProGS and PCGS use native prefixes; Octree-GS-EC-LR uses decoded cumulative LoD groups. HAC/HAC++ random packetizations report ten-seed mean and standard deviation. The common target $Q_s^*$ is defined over these five methods in Eq.~\eqref{eq:streaming_quality_target}. Piecewise-linear Y-axes compress the sparse low-quality range; a kink on each left spine marks the scale transition, and ticks remain in dB.}
    \label{fig:packetized_streaming}
\end{figure*}

\subsection{Experimental Setup}\label{sec:experimental_setup}

\noindent\textbf{Implementation.}
ProGS is implemented in PyTorch~\cite{paszke2019pytorch} on top of the anchor representation and entropy-coding pipeline of HAC~\cite{chen2024hachashgridassistedcontext}, and all experiments run on a single NVIDIA RTX~4090 GPU. Each anchor consists of a 32-dimensional feature vector, six scaling parameters, and $K=10$ offset vectors, with $L=5$ LoD levels from lv0 to lv4. The main codec is optimized for 40{,}000 iterations with $\lambda_\mathrm{s}=0.2$ and $\lambda_\mathrm{scale}=0.01$, with multi-LoD prefix supervision and anchor growth active from iteration 1{,}500 to 25{,}000 and the octree frozen thereafter (Fig.~\ref{fig:trainingprocess}). Growth updates every $T=100$ iterations use $\tau_g=5\times10^{-5}$ and $\beta=0.01$, and leaf pruning uses $\tau_o=0.005$. The low-LoD range is lv0--lv1 ($l_\mathrm{low}=1$), level balancing is applied to lv0 with $r_\mathrm{bal}=0.02$ and $B_\mathrm{min}=4096$, and the growth-score weights are $w_\mathrm{full}=2.0$, $w_\mathrm{low}=1.0$, $w_\mathrm{rate}=0.5$, and $w_\mathrm{cl}=0.75$.

\noindent\textbf{Coding and refinement.}
The implementation uses the mixed 3D--2D binary hash-grid of HAC~\cite{chen2024hachashgridassistedcontext} with four-dimensional entries, base quantization steps of 1, 0.001, and 0.2 for feature, scaling, and offset attributes, and five rate points $\lambda_\mathrm{e}\in\{0.004,0.003,0.002,0.001,0.0005\}$.

The frozen low-prefix refinement stage runs 5{,}000 additional iterations on lv0--lv2 with Adam and a two-hidden-layer decoder of width 64. The 25\% of lv0--lv2 anchors with the largest maximum, over the three target prefixes, of the mean absolute decoder-context value carry two-dimensional tokens, uniformly quantized to 8 bits after training. The refinement stage uses $\lambda_\mathrm{rs}=0.05$ and teacher weights $(\eta_0,\eta_1,\eta_2)=(0.5,0.6,0.9)$; the residual bounds for feature, offset, scaling, and opacity are 0.20, 0.02, 0.015, and 0.08. The side stream contains refinement-decoder weights, the level-partitioned selection masks, and the quantized tokens; it is transmitted once after lv0 and reused by lv0--lv2.

\noindent\textbf{Datasets and protocols.}
The evaluation is conducted on all nine scenes of Mip-NeRF360~\cite{barron2022mip}, the six BungeeNeRF~\cite{xiangli2022bungeenerf} scenes shared by Scaffold-GS and HAC (\textit{amsterdam}, \textit{bilbao}, \textit{hollywood}, \textit{pompidou}, \textit{quebec}, and \textit{rome}), and the \textit{train} and \textit{truck} scenes of Tanks and Temples~\cite{knapitsch2017tanks}. ProGS uses five rate settings per scene. Octree-GS-EC follows the official implementation's public 40{,}000-iteration structural, LoD, visibility, and distance settings and adds learned entropy models, offset masking, hash-grid and inter-level context, quantization, rate estimation, and per-LoD coding for decoded-rate evaluation. After iteration 20{,}000, its objective adds $\lambda$-weighted estimated rate and mask regularization; EC-LR/HR set $\lambda=0.004/0.0005$. Decoded lv0--lv4 quality and cumulative bitstream size are reported.

\noindent\textbf{Metrics and rate accounting.}
Evaluation metrics include peak signal-to-noise ratio (PSNR), structural similarity index (SSIM), learned perceptual image patch similarity (LPIPS)~\cite{zhang2018unreasonable}, encoded size, training and coding time, and rendering frames per second (FPS). Prefix size is cumulative: it includes the shared header, all level chunks received up to the evaluated LoD, and the refinement side stream transmitted after lv0. The main rate--distortion curves and endpoint comparisons include this side stream at lv3 and lv4; tables labeled \textit{base full size} report the Stage~1 base-only profile.

\begin{figure*}[htbp]
    \centering
    \footnotesize
    \setlength{\tabcolsep}{1.5pt}
    \renewcommand{\arraystretch}{1.0}
    \begin{tabular*}{\textwidth}{@{\extracolsep{\fill}}cccccccc@{}}
        \toprule
        \qualhead{Scene} & \qualhead{GT Context} & \qualhead{GT Detail} &
        \qualhead{lv0} & \qualhead{lv1} & \qualhead{lv2} & \qualhead{lv3} & \qualhead{lv4} \\
        \midrule
        \qualscenelabel{bicycle} &
        \qualcontext{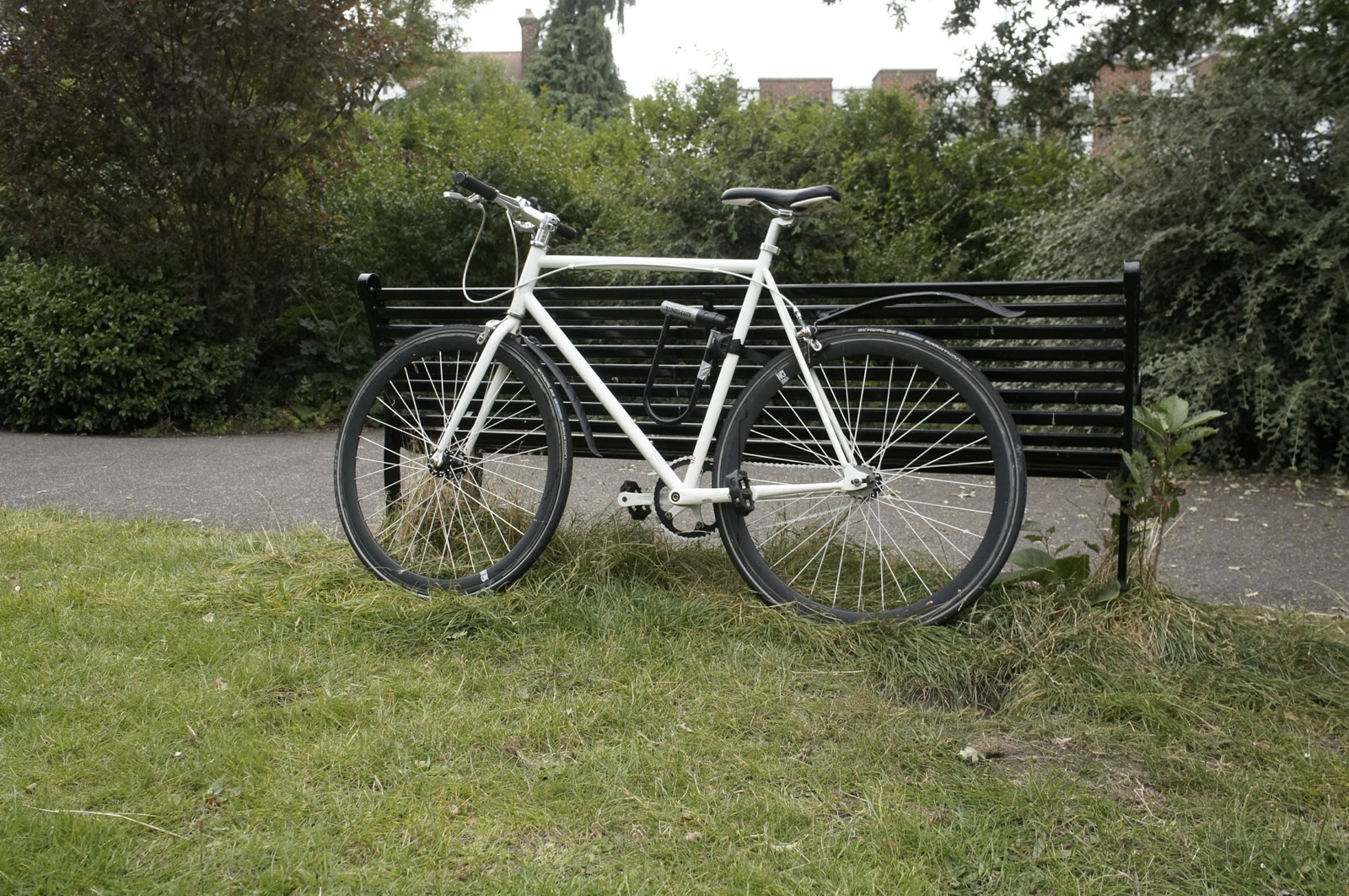}{0.367}{0.696}{0.389}{0.718} &
        \qualdetail{587 413 487 300}{visualization/mipnerf360/bicycle/GT.jpg} &
        \qualmetriccrop{587 413 487 300}{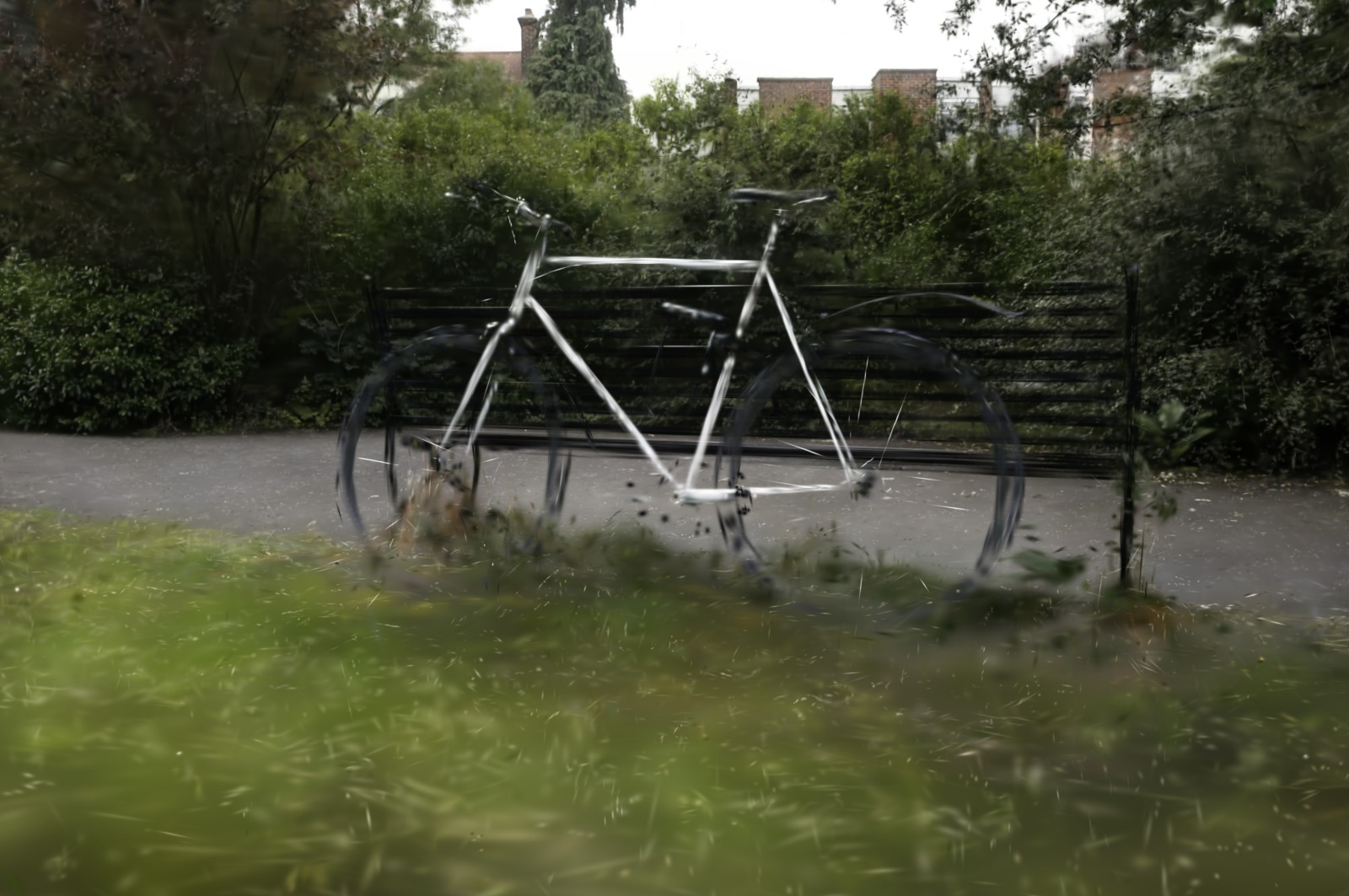}{16.72}{11.01} &
        \qualmetriccrop{587 413 487 300}{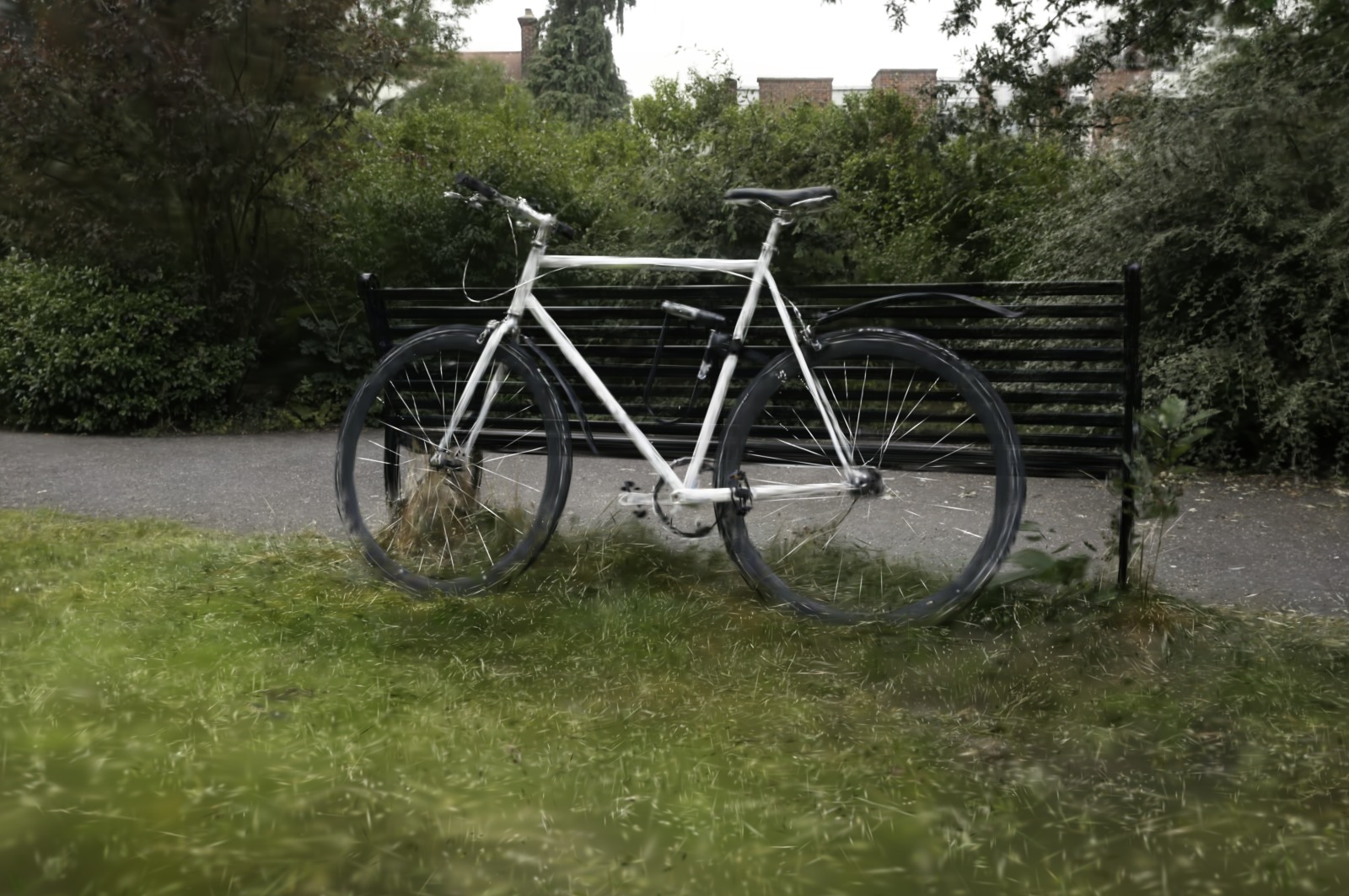}{18.38}{20.24} &
        \qualmetriccrop{587 413 487 300}{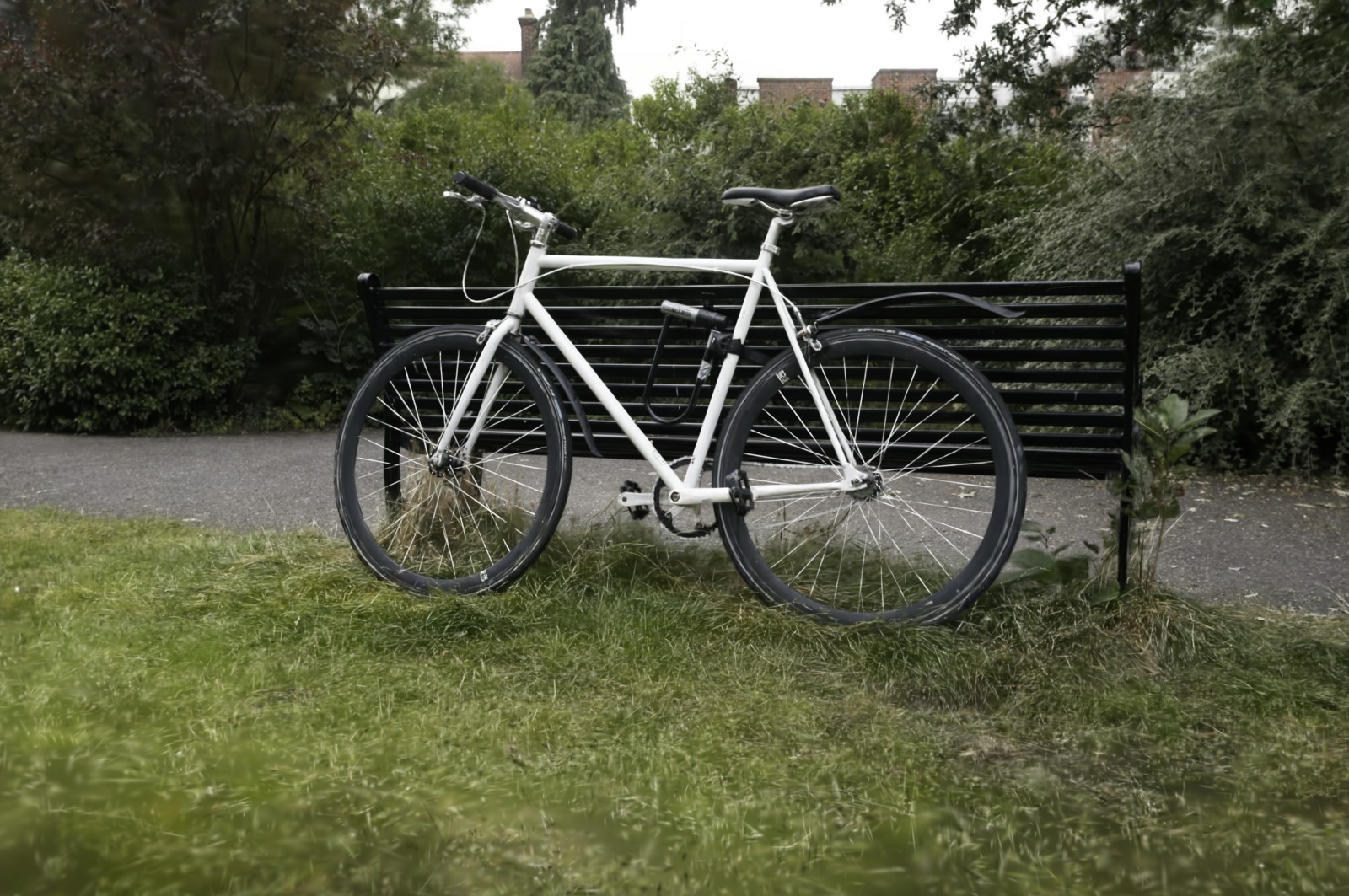}{20.20}{25.26} &
        \qualmetriccrop{587 413 487 300}{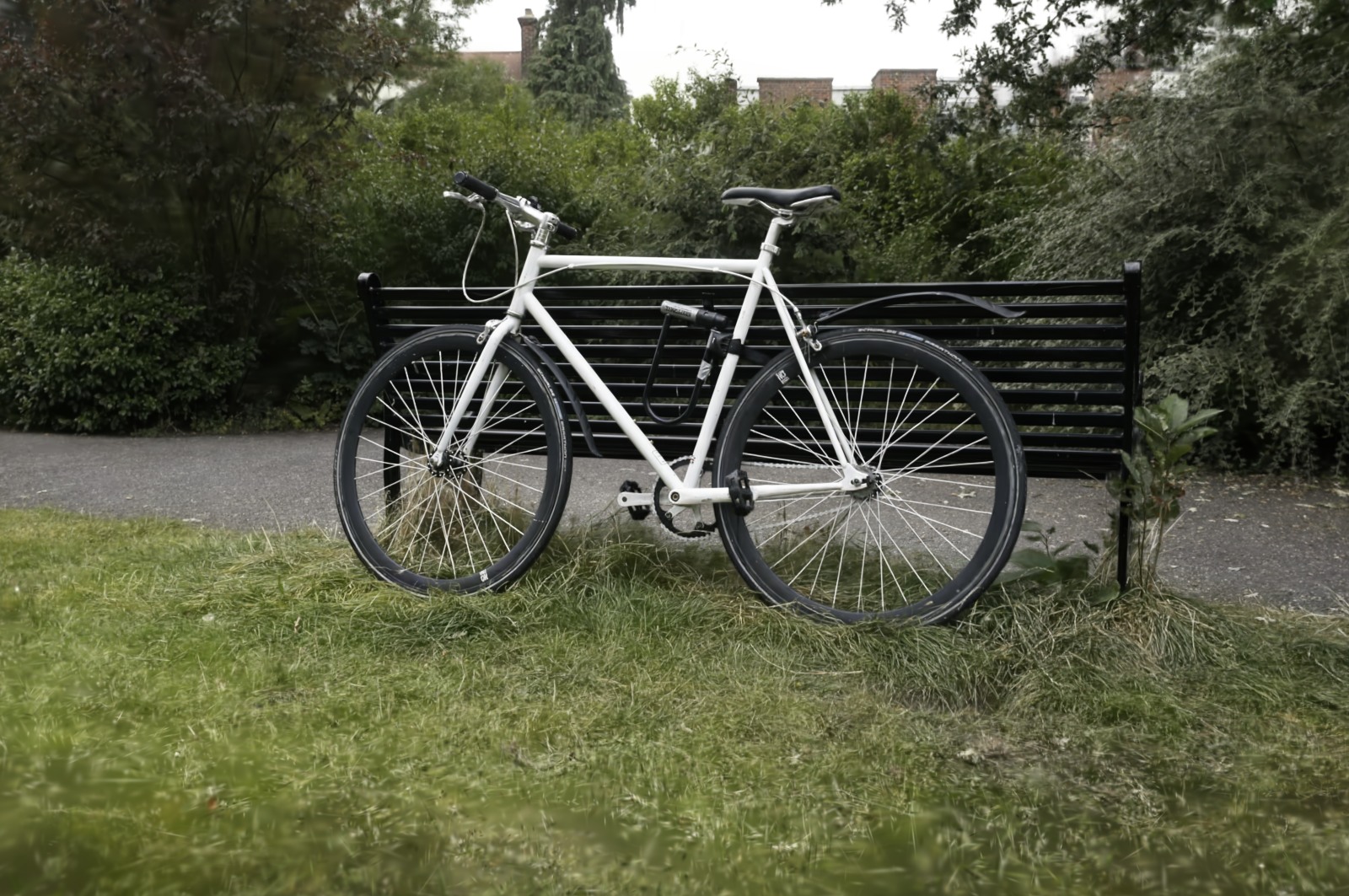}{21.16}{27.98} &
        \qualmetriccrop{587 413 487 300}{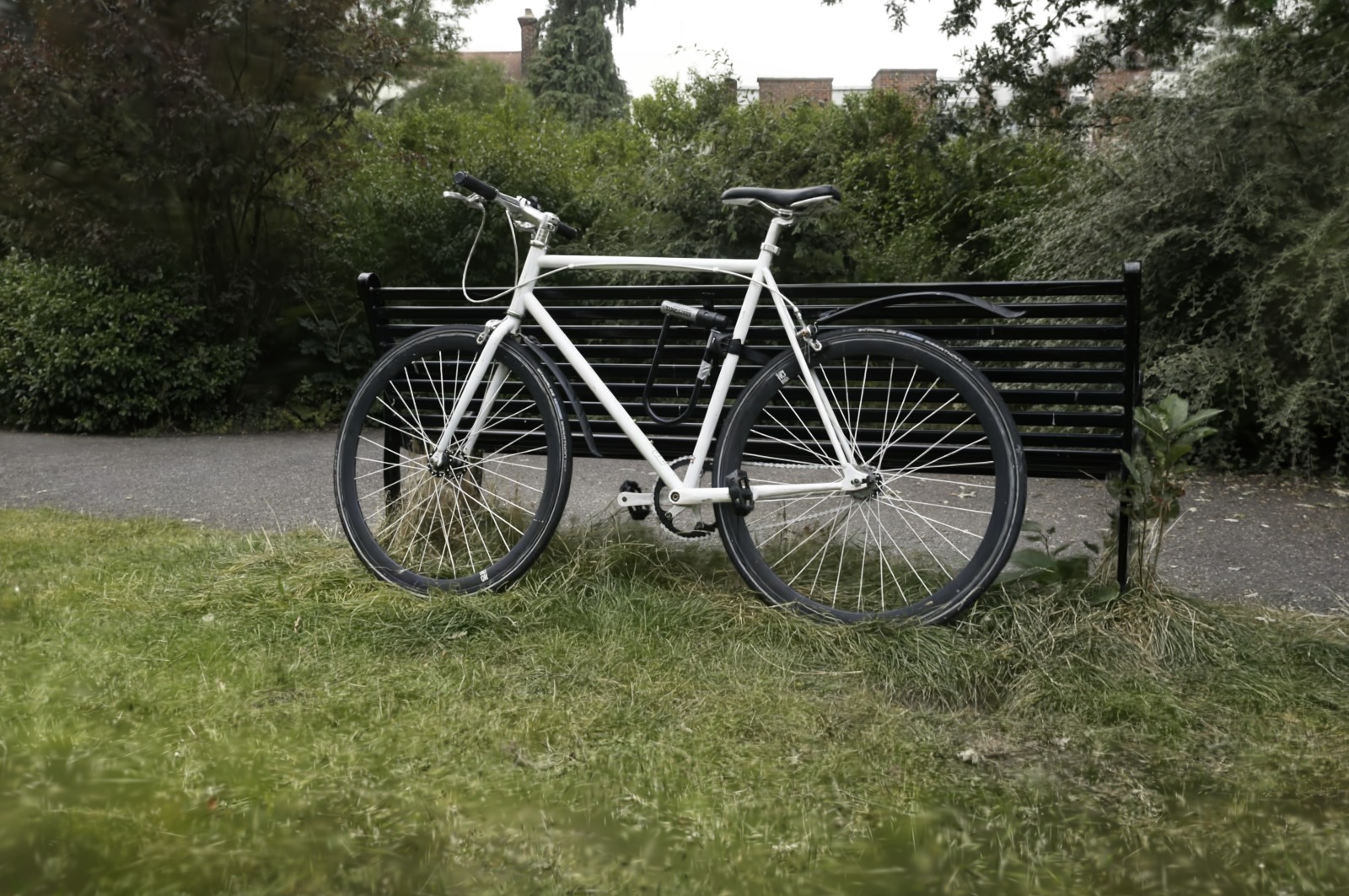}{21.20}{28.82} \\
        \qualscenelabel{room} &
        \qualcontext{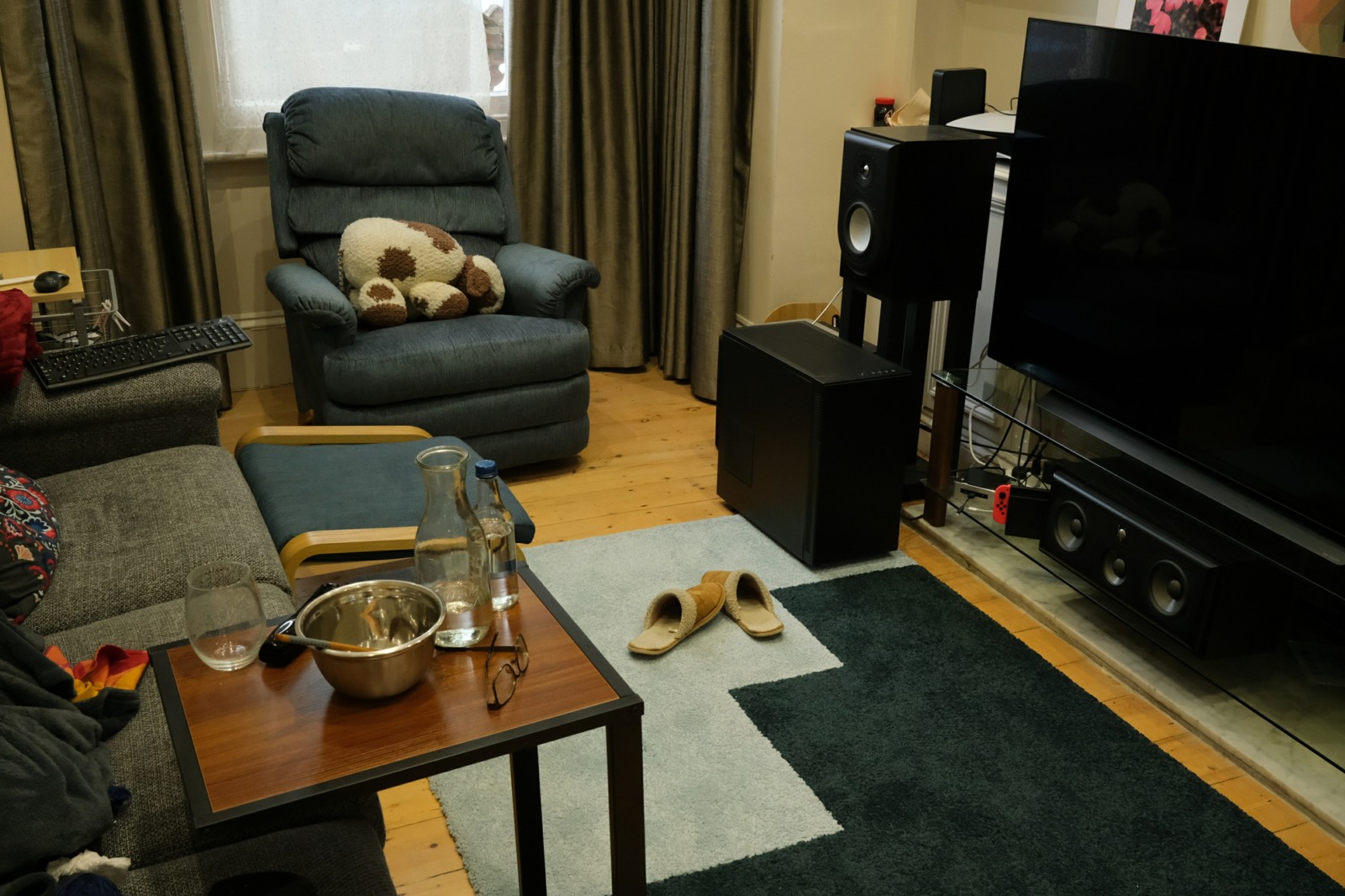}{0.192}{0.464}{0.559}{0.831} &
        \qualdetail{307 596 857 180}{visualization/mipnerf360/room/GT.jpg} &
        \qualmetriccrop{307 596 857 180}{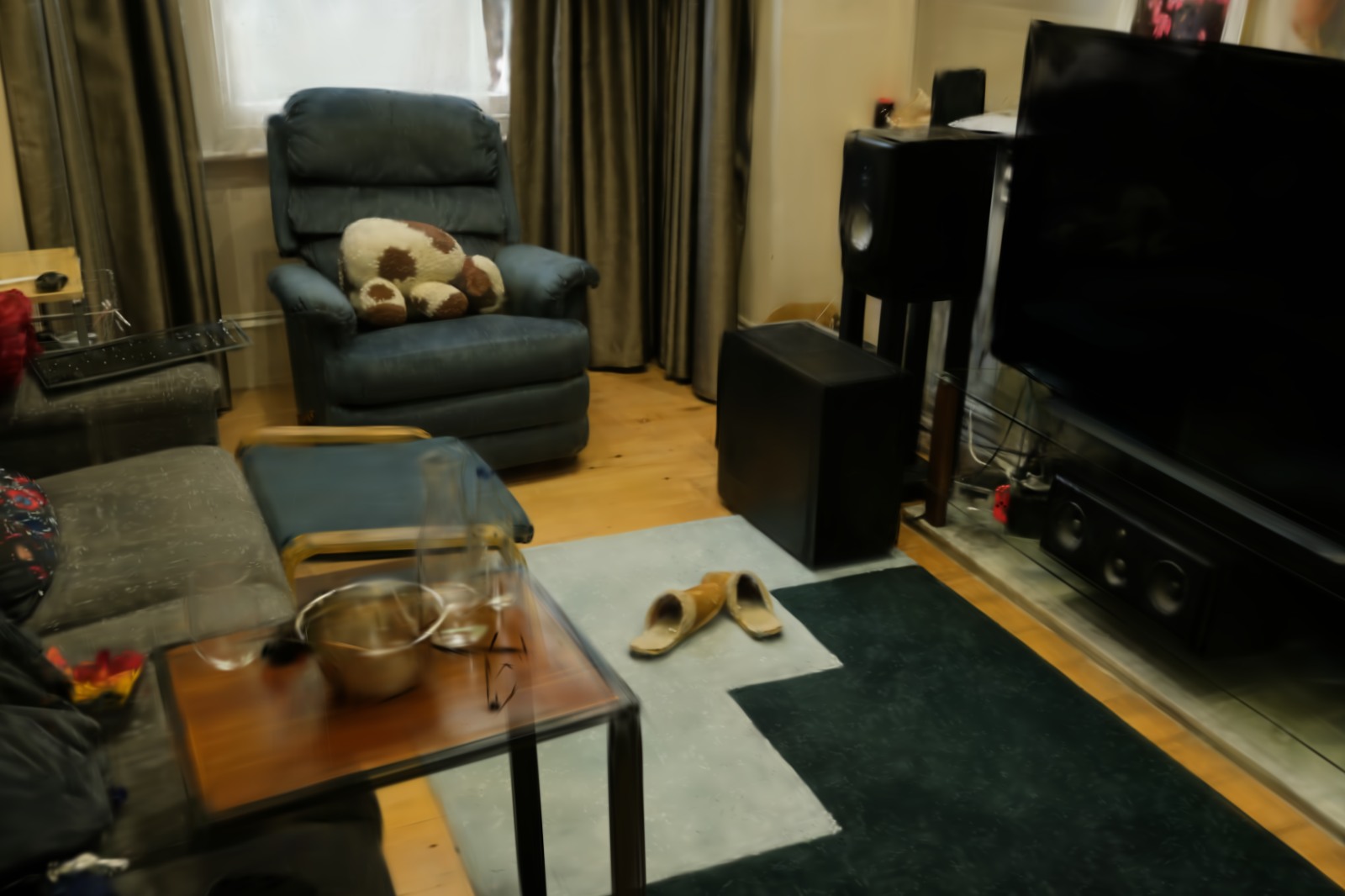}{25.91}{1.99} &
        \qualmetriccrop{307 596 857 180}{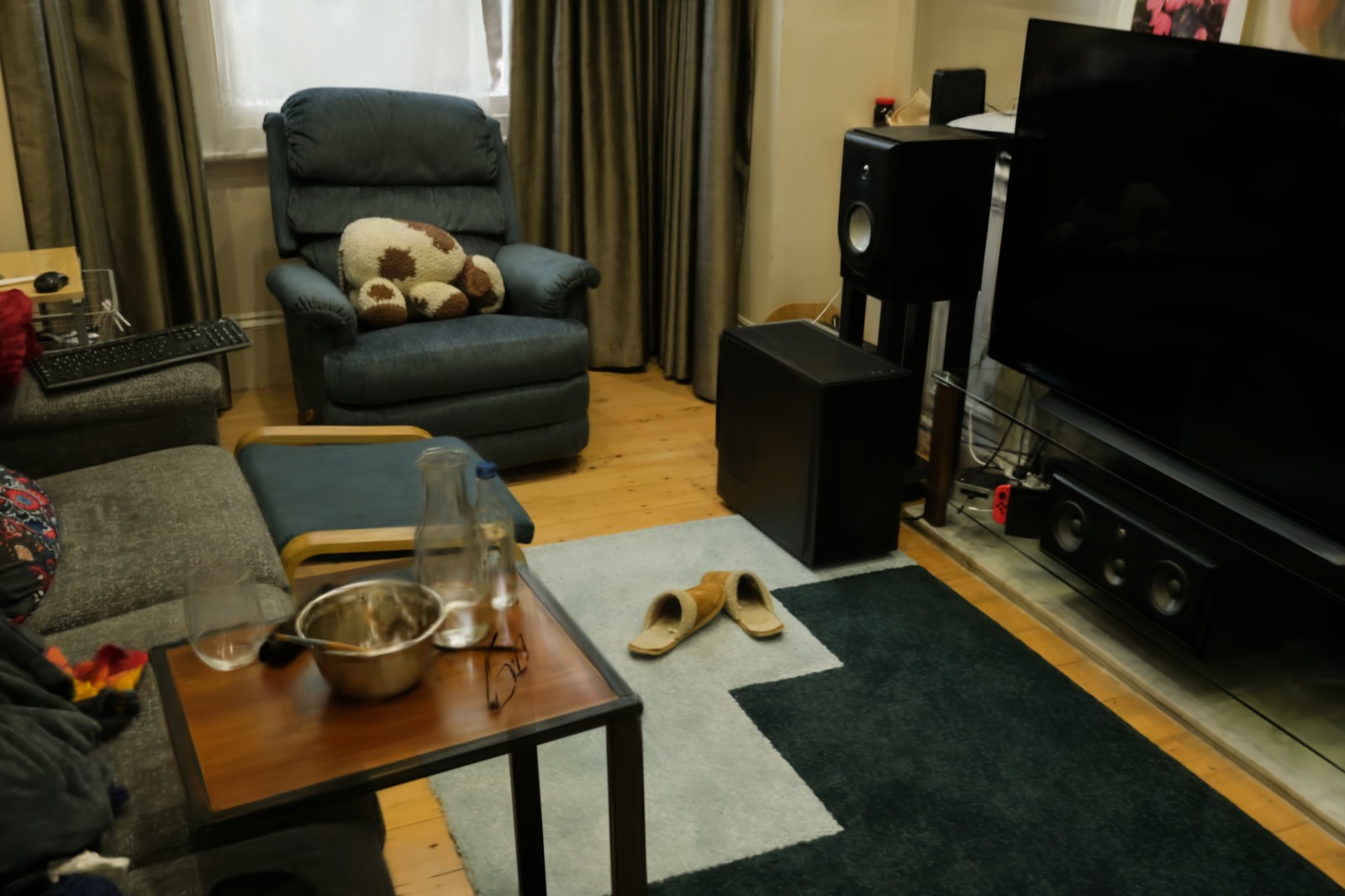}{29.08}{3.57} &
        \qualmetriccrop{307 596 857 180}{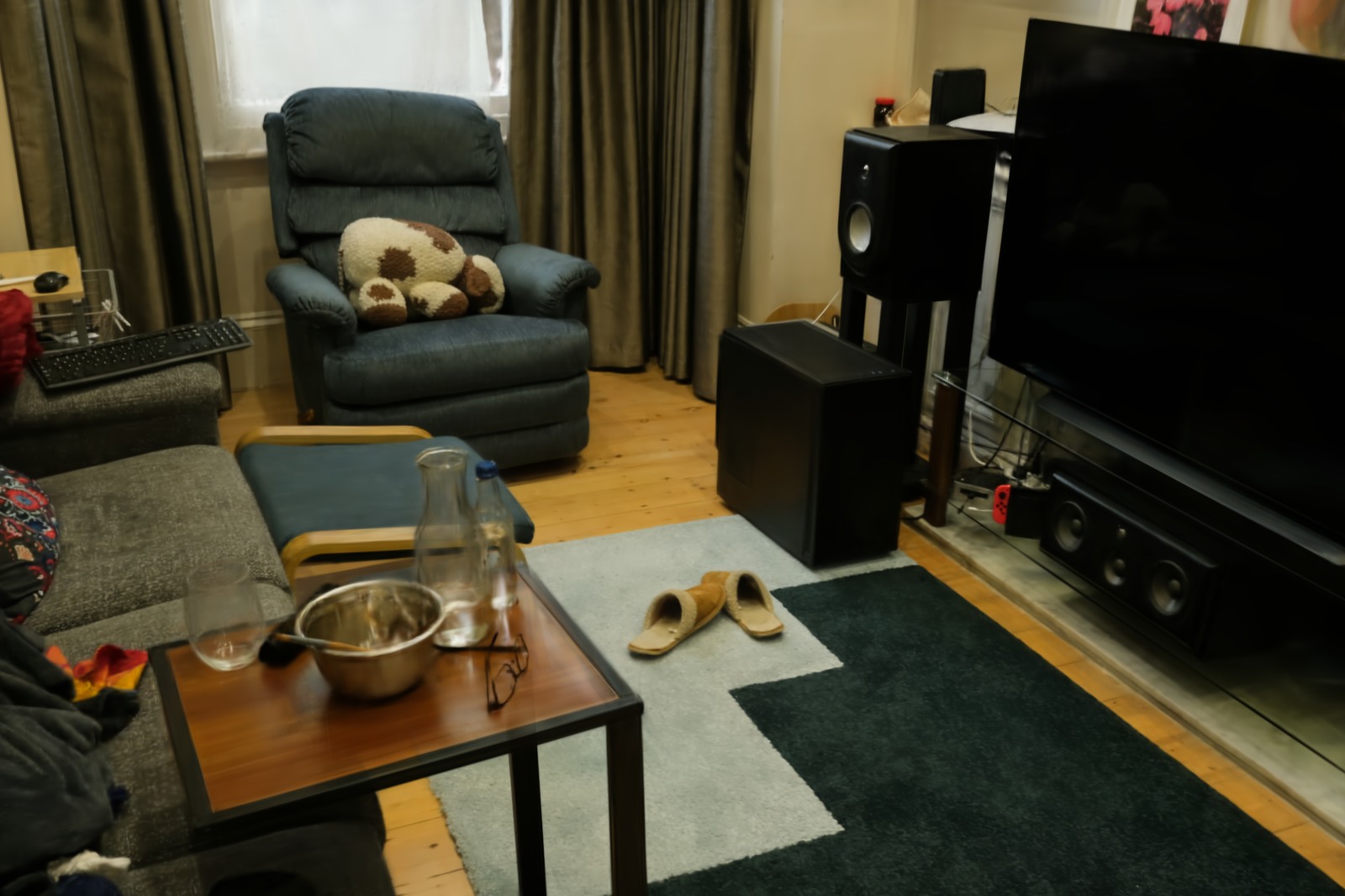}{29.47}{5.11} &
        \qualmetriccrop{307 596 857 180}{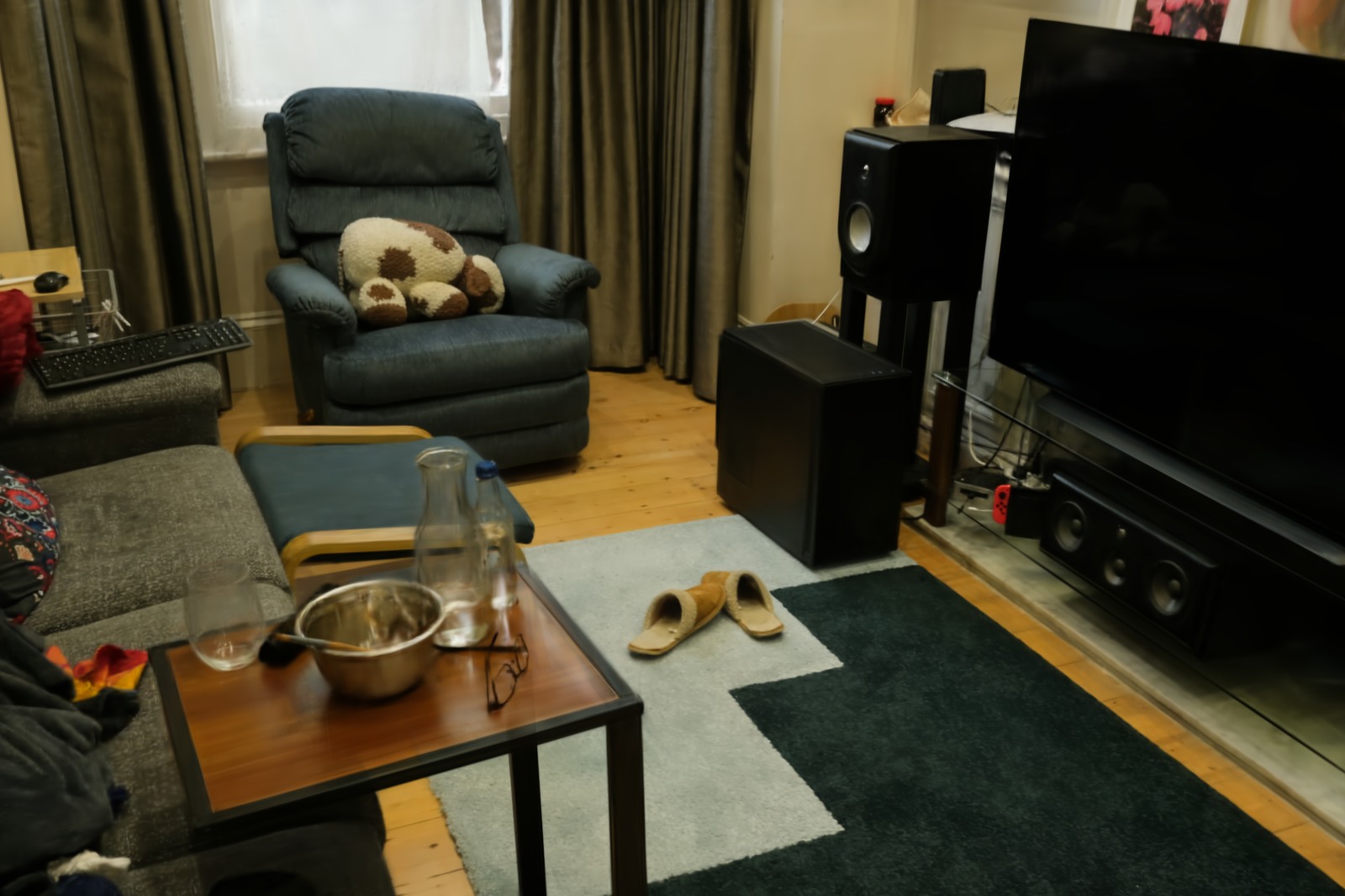}{29.47}{8.64} &
        \qualmetriccrop{307 596 857 180}{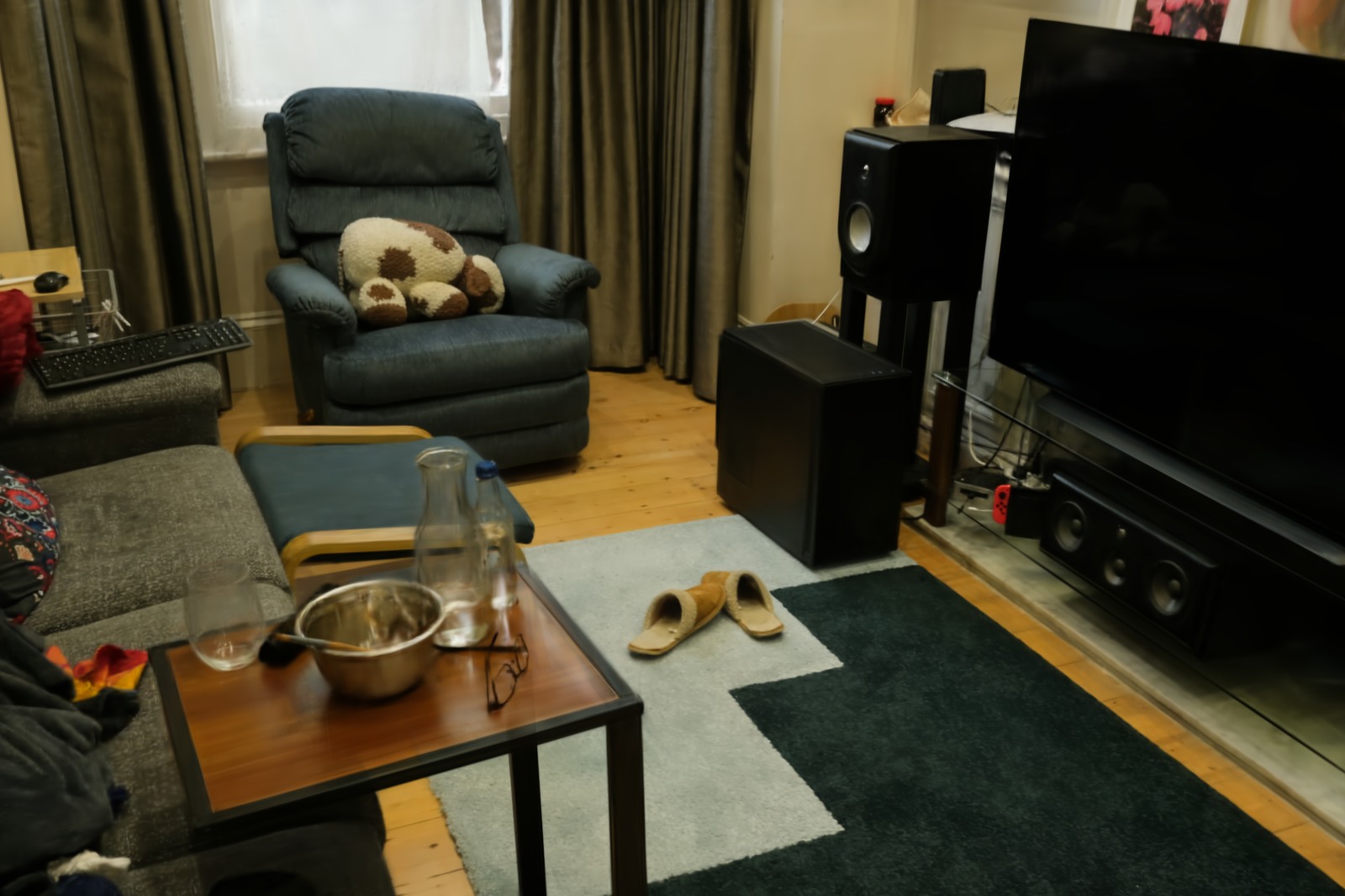}{29.47}{10.93} \\
        \qualscenelabel{truck} &
        \qualcontext{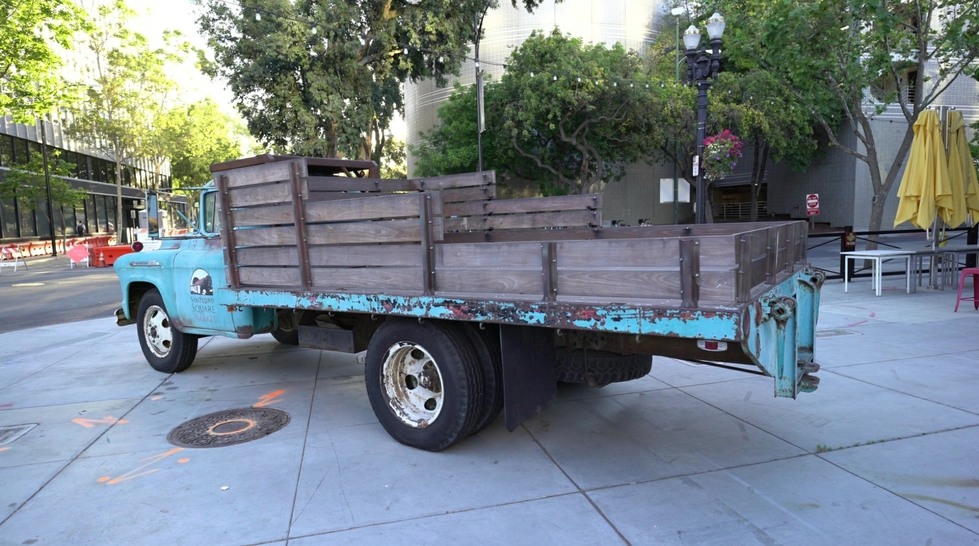}{0.405}{0.679}{0.231}{0.505} &
        \qualdetail{396 126 314 270}{visualization/tanksandtemples/truck/GT.jpg} &
        \qualmetriccrop{396 126 314 270}{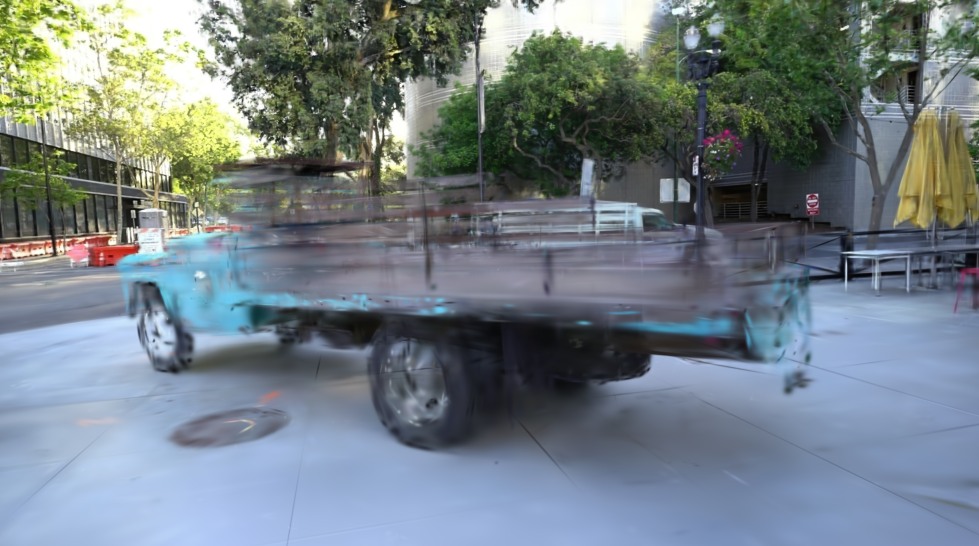}{20.38}{4.29} &
        \qualmetriccrop{396 126 314 270}{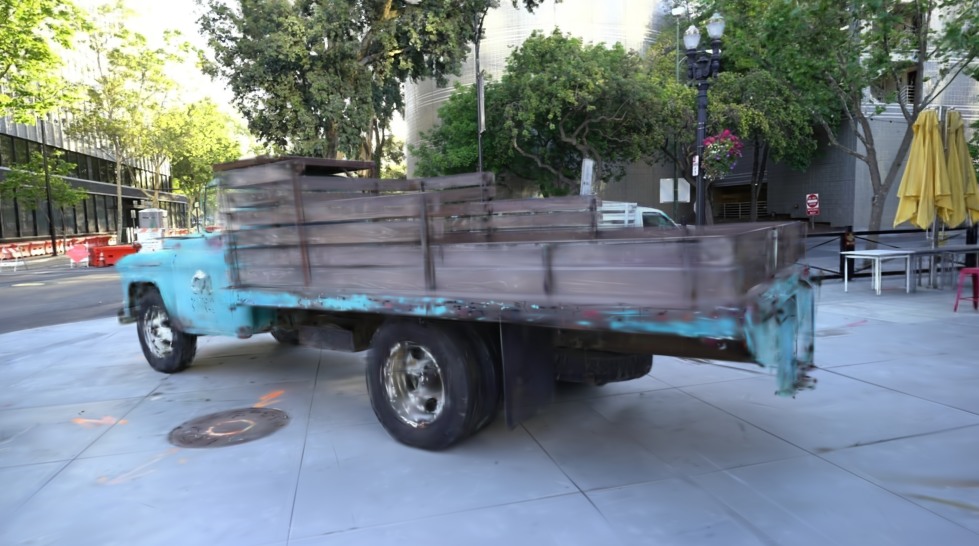}{23.18}{7.15} &
        \qualmetriccrop{396 126 314 270}{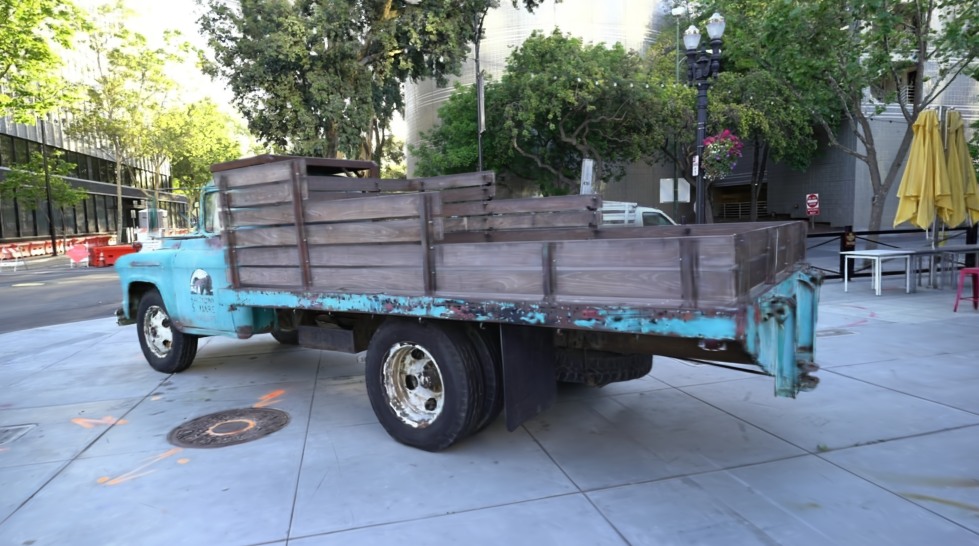}{25.73}{9.09} &
        \qualmetriccrop{396 126 314 270}{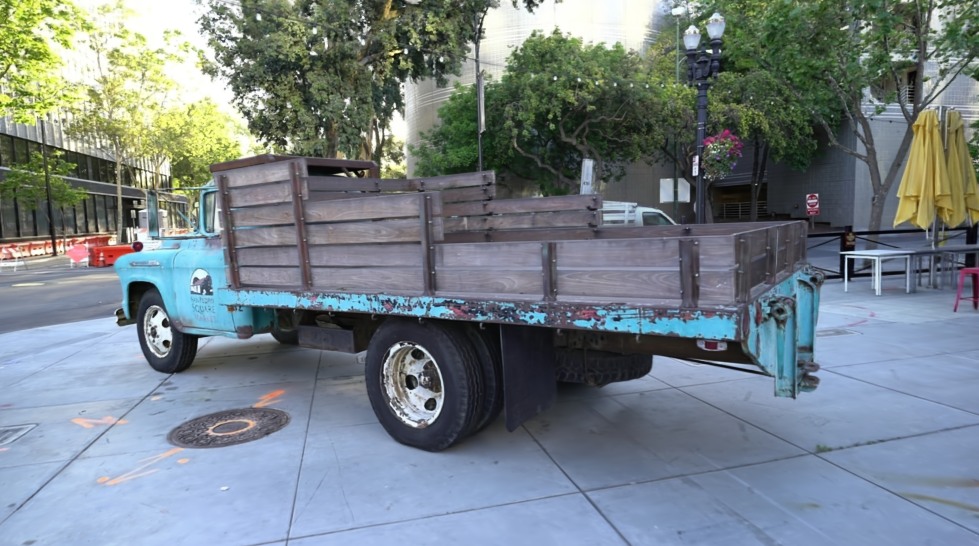}{26.75}{10.80} &
        \qualmetriccrop{396 126 314 270}{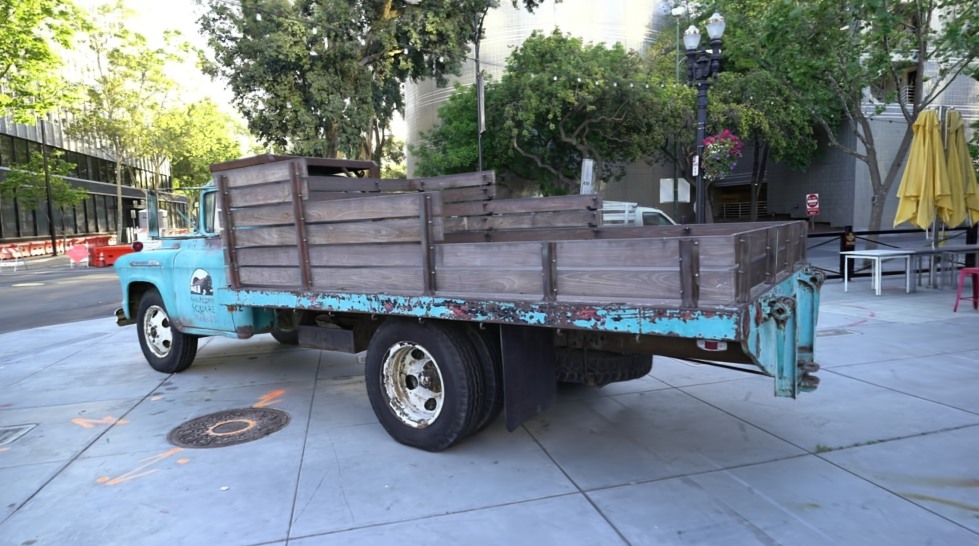}{26.76}{11.96} \\
        \bottomrule
    \end{tabular*}
    \caption{Progressive decoding on representative test views; each row shows the scene context, ground-truth detail, and the crop from lv0 to lv4. Red boxes mark the detail region; gray tags report full-view PSNR (dB) and cumulative size (MB).}
    \label{fig:LoD}
\end{figure*}

\begin{figure*}[htbp]
    \centering
    \footnotesize
    \setlength{\tabcolsep}{1.5pt}
    \renewcommand{\arraystretch}{1.0}
    \begin{tabular*}{\textwidth}{@{\extracolsep{\fill}}cccccccc@{}}
        \toprule
        \qualhead{Scene} & \qualhead{GT Context} & \qualhead{GT Detail} &
        \qualhead{ProGS\\lv0} &
        \qualhead{ProGS\\lv1} &
        \qualhead{PCGS\\prefix} &
        \qualhead{Octree-GS-EC-LR\\LoD} &
        \qualhead{HAC++\\single rate} \\
        \midrule
        \qualscenelabel{garden} &
        \qualcontext[0.115\textwidth]{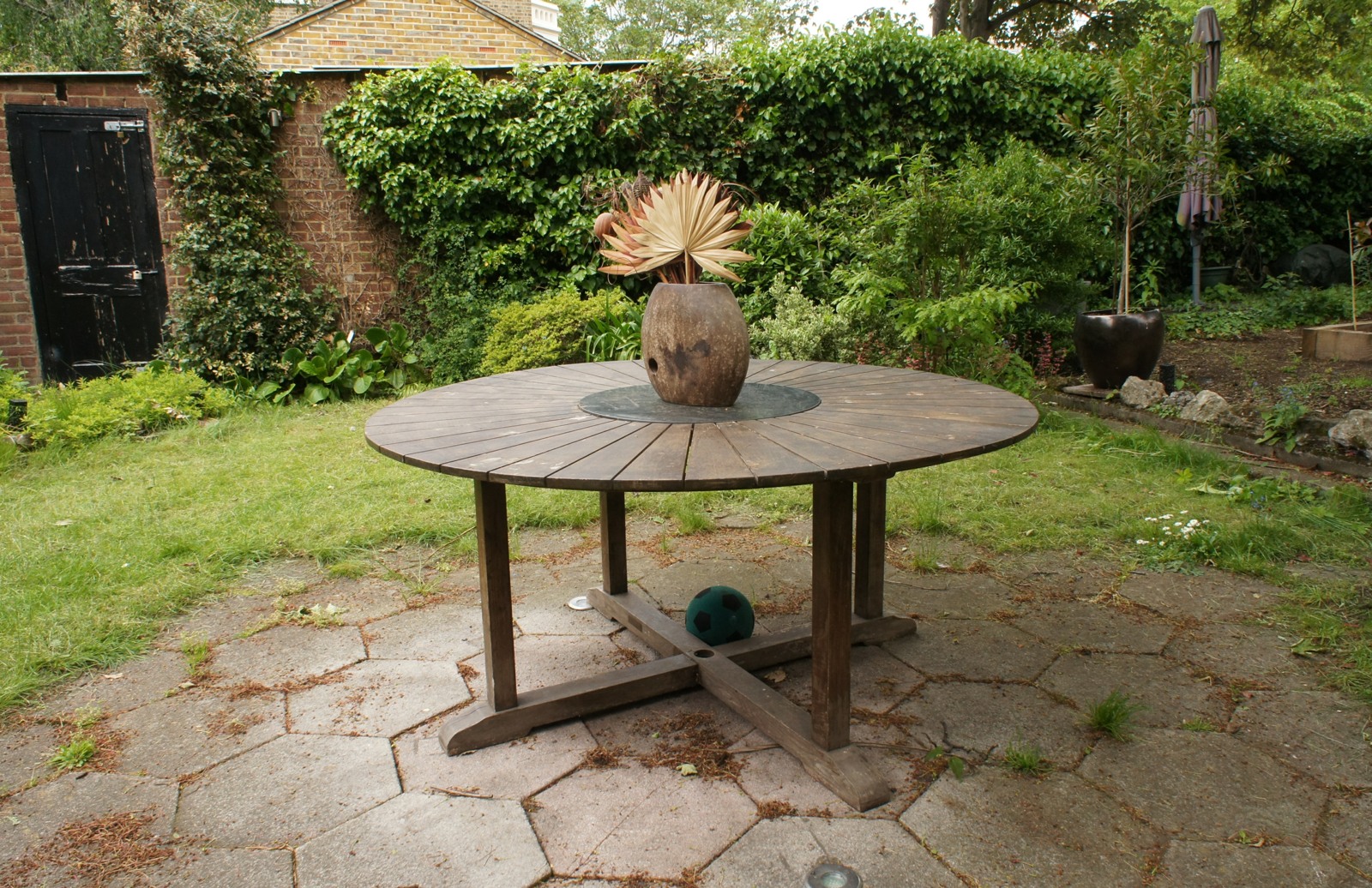}{0.300}{0.700}{0.479}{0.826} &
        \qualdetail[0.115\textwidth]{480 496 480 180}{visualization/mipnerf360/garden/GT.jpg} &
        \qualmetriccrop[0.115\textwidth]{480 496 480 180}{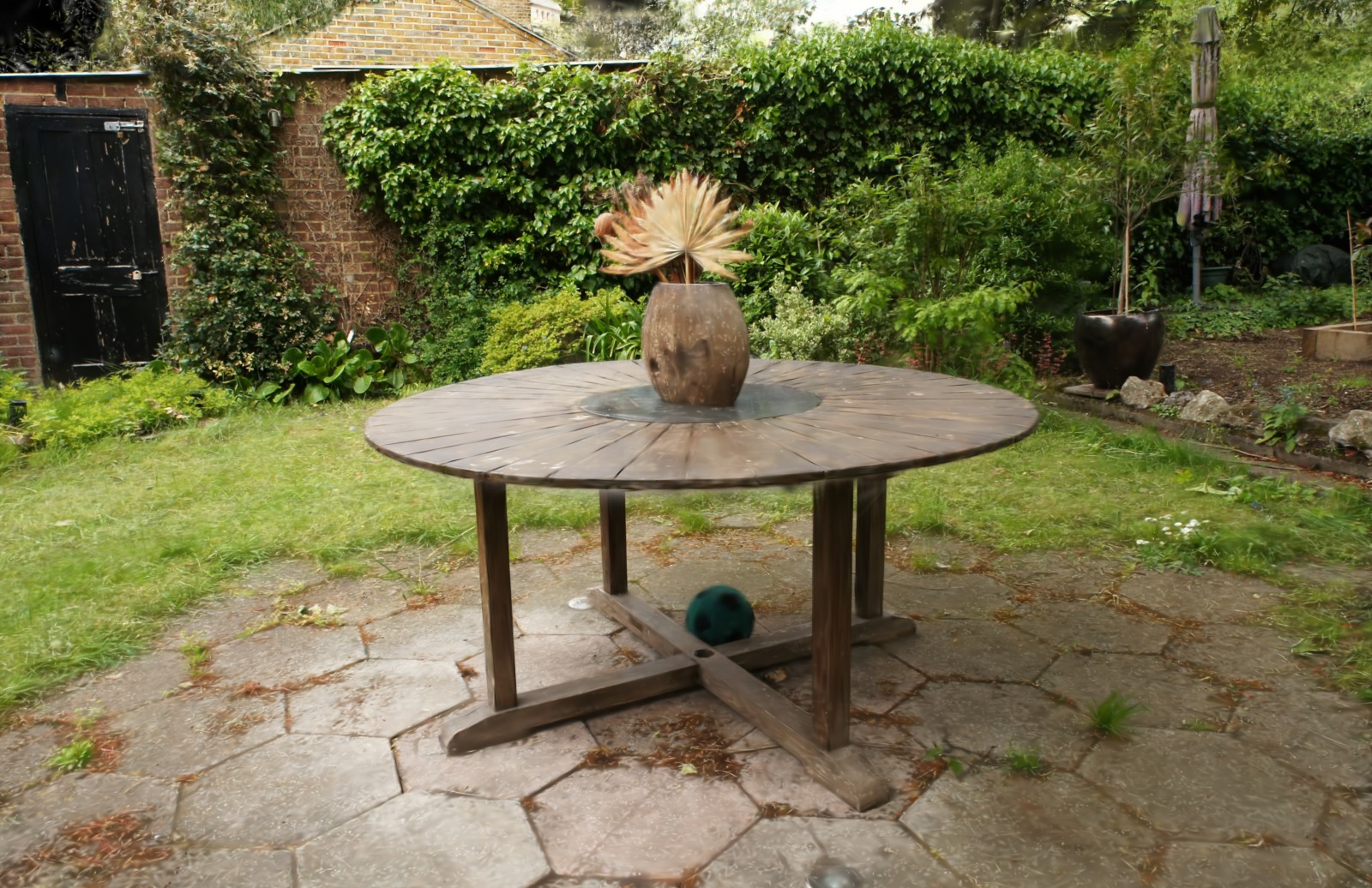}{24.95}{13.45} &
        \qualmetriccrop[0.115\textwidth]{480 496 480 180}{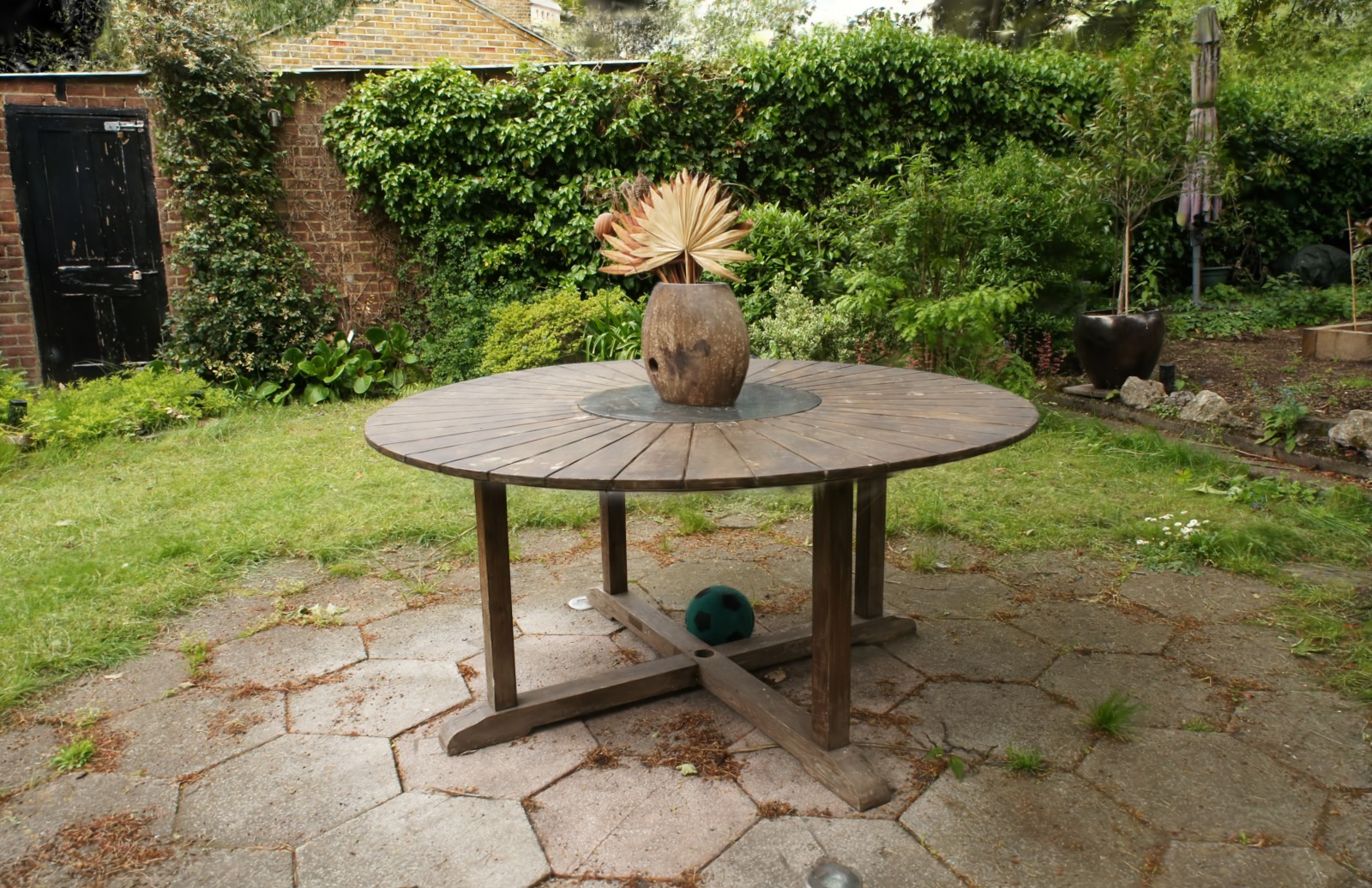}{26.95}{24.27} &
        \qualmetriccrop[0.115\textwidth]{480 496 480 180}{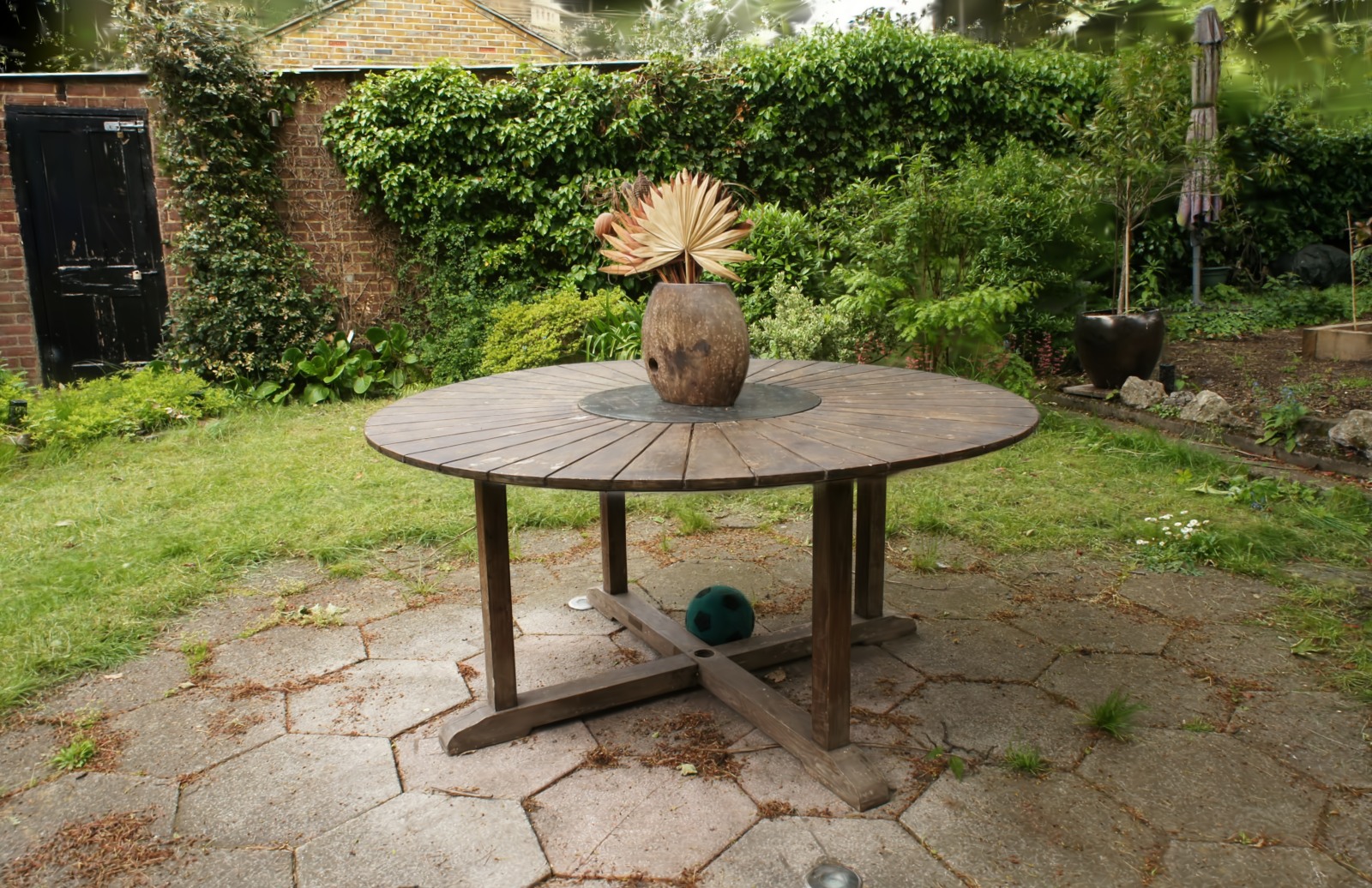}{27.47}{24.93} &
        \qualmetriccrop[0.115\textwidth]{480 496 480 180}{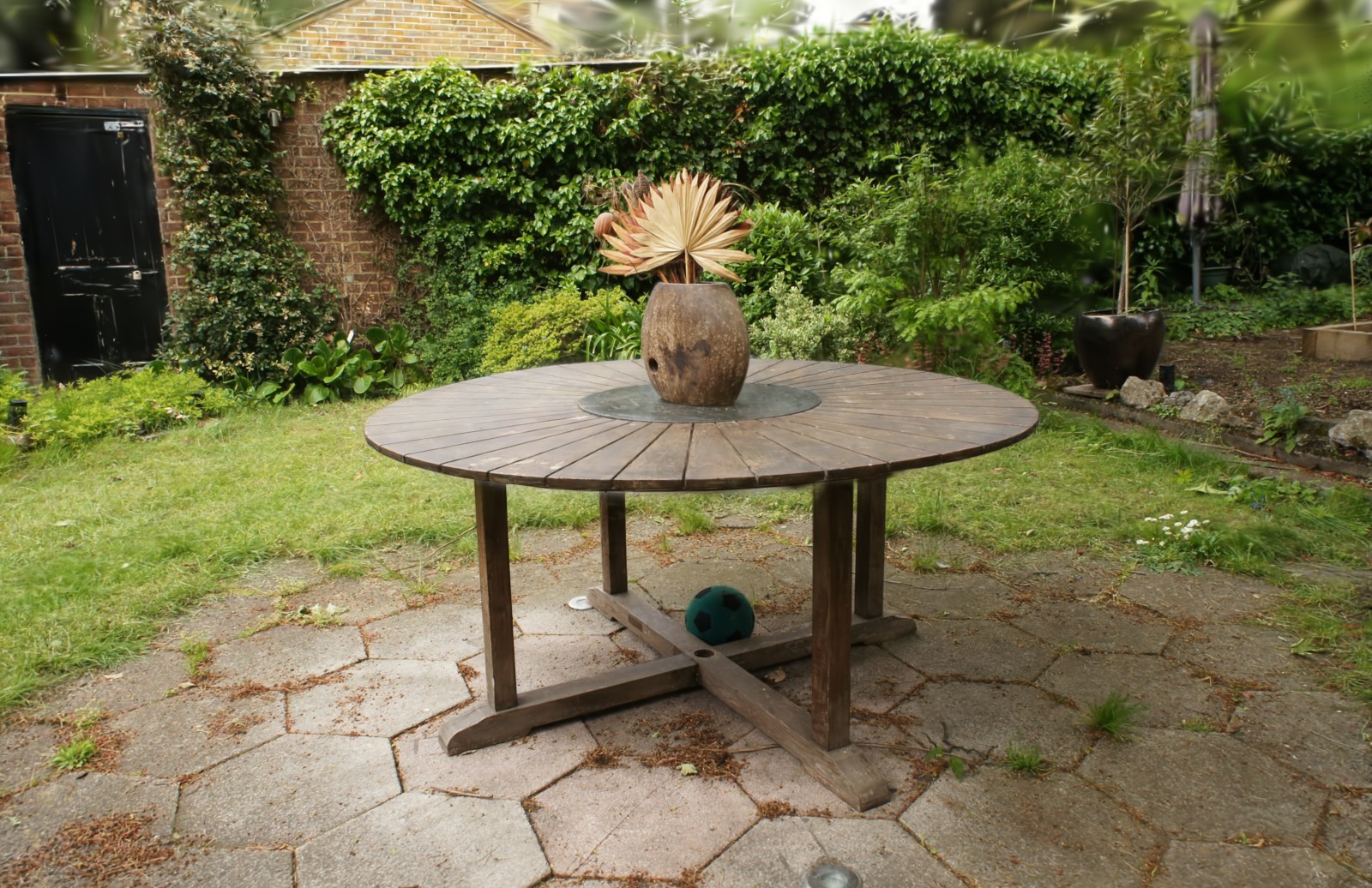}{27.13}{18.65} &
        \qualmetriccrop[0.115\textwidth]{480 496 480 180}{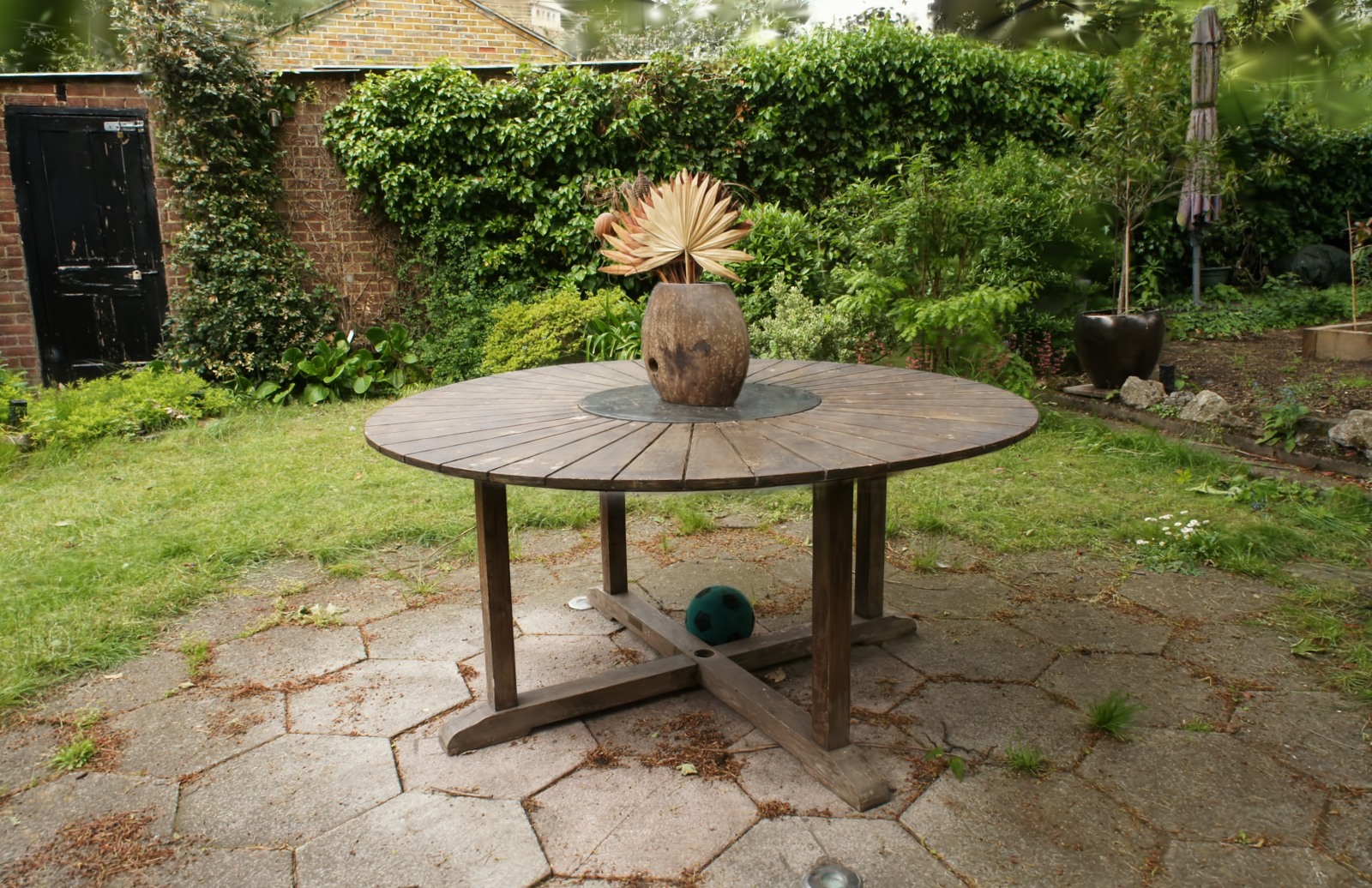}{27.50}{27.49} \\
        \qualscenelabel{pompidou} &
        \qualcontext[0.115\textwidth]{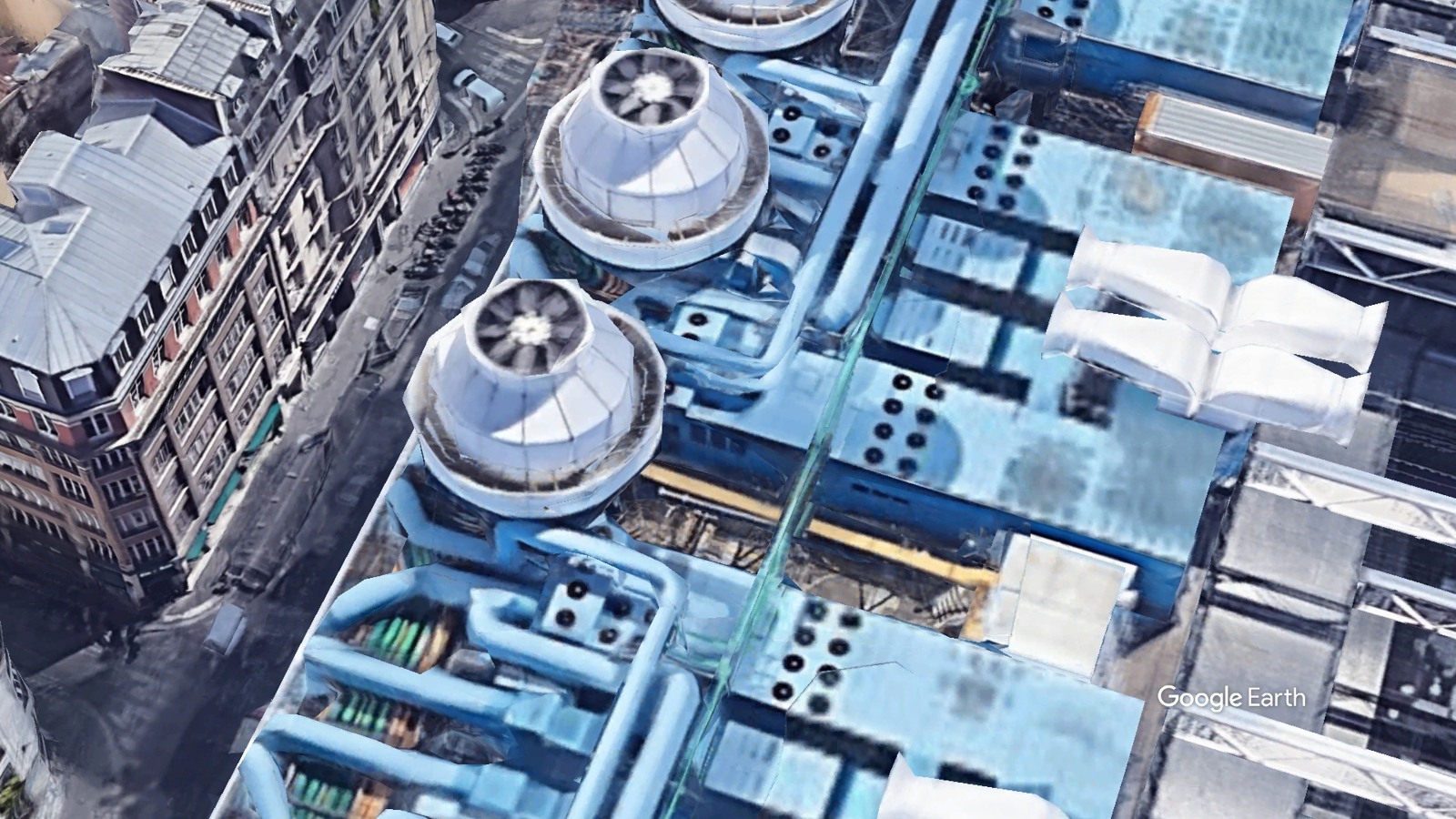}{0.050}{0.375}{0.508}{0.833} &
        \qualdetail[0.115\textwidth]{80 457 1000 150}{visualization/bungeenerf/pompidou/GT.jpg} &
        \qualmetriccrop[0.115\textwidth]{80 457 1000 150}{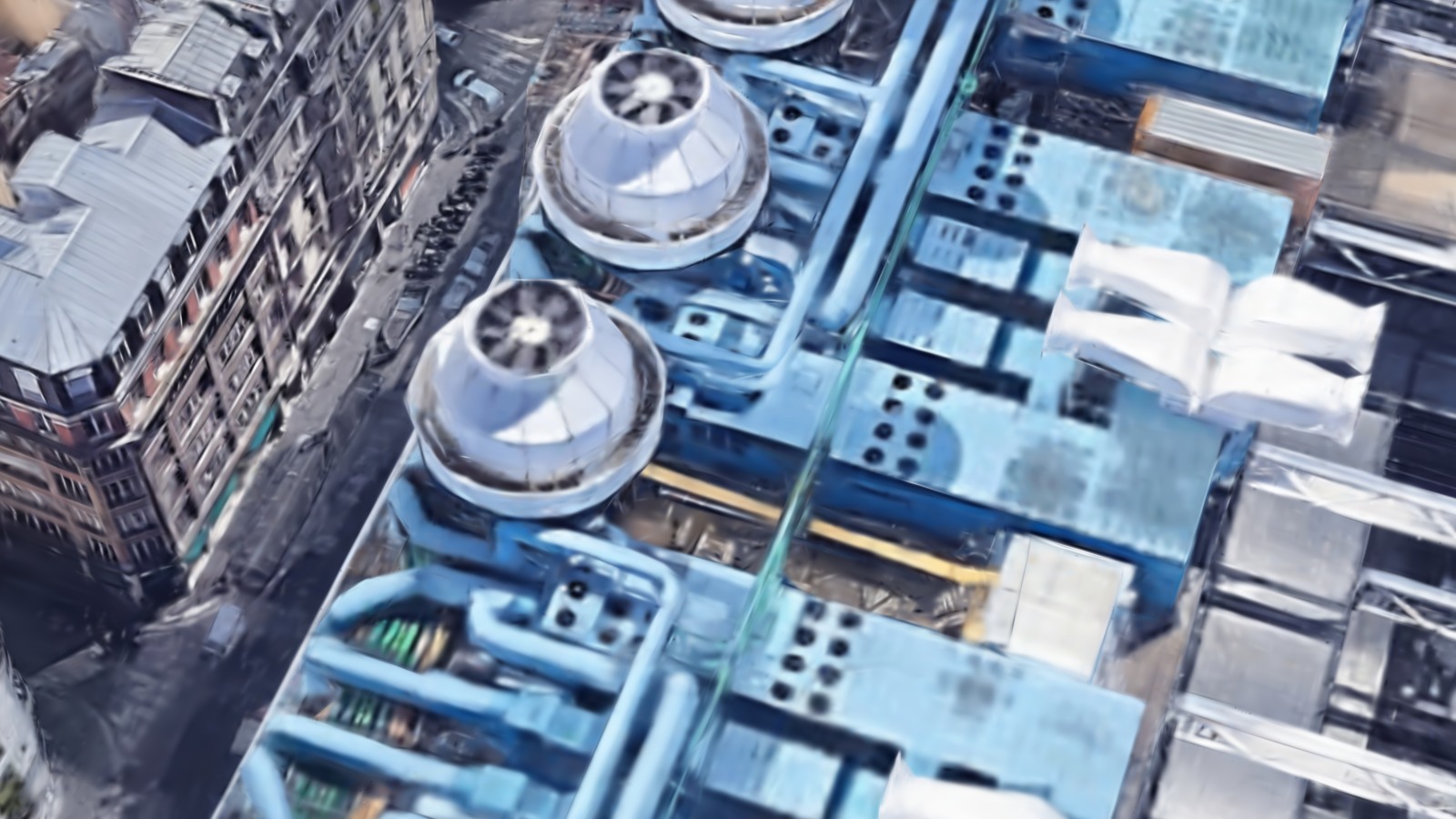}{24.80}{21.62} &
        \qualmetriccrop[0.115\textwidth]{80 457 1000 150}{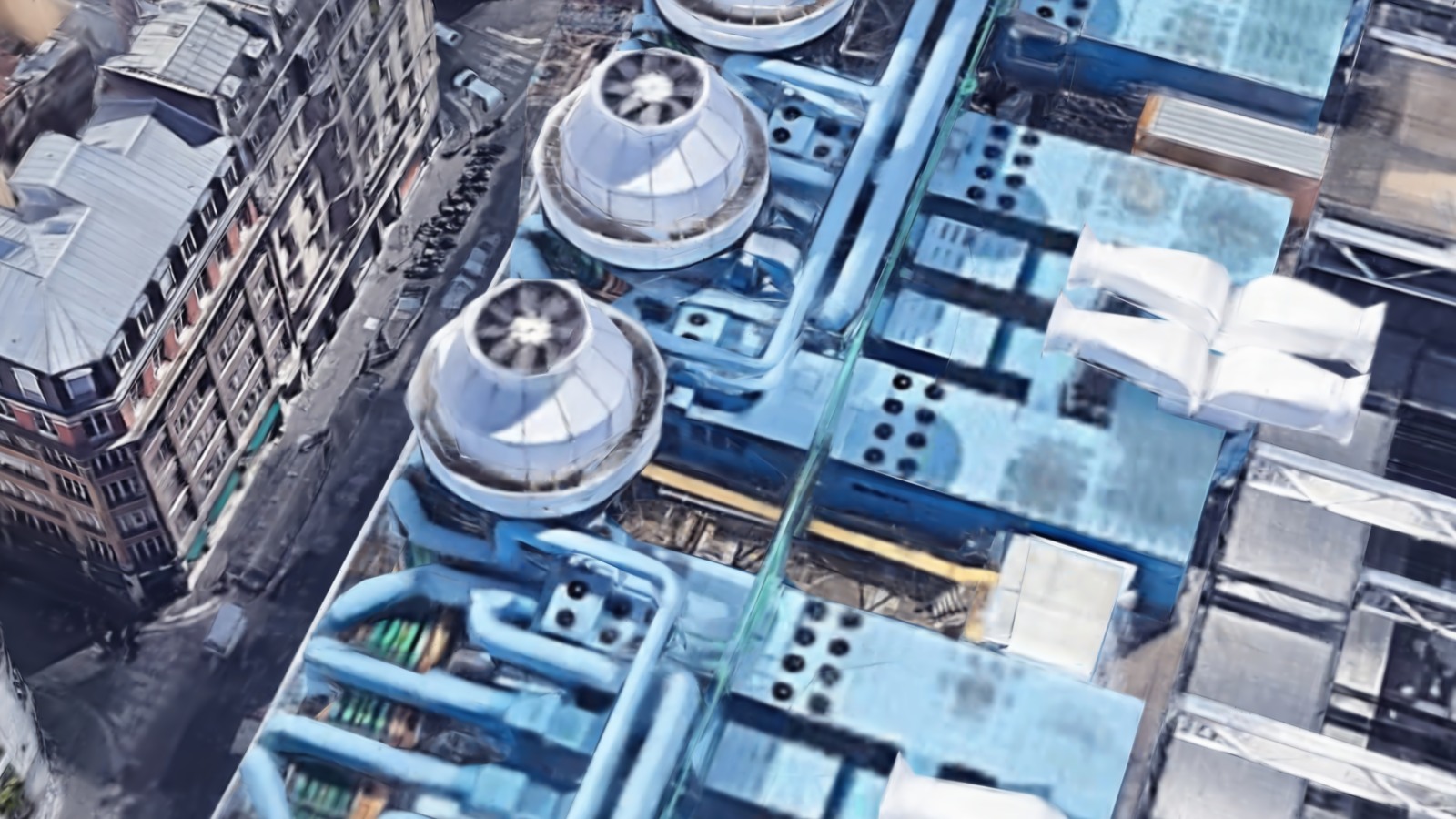}{26.21}{25.21} &
        \qualmetriccrop[0.115\textwidth]{80 457 1000 150}{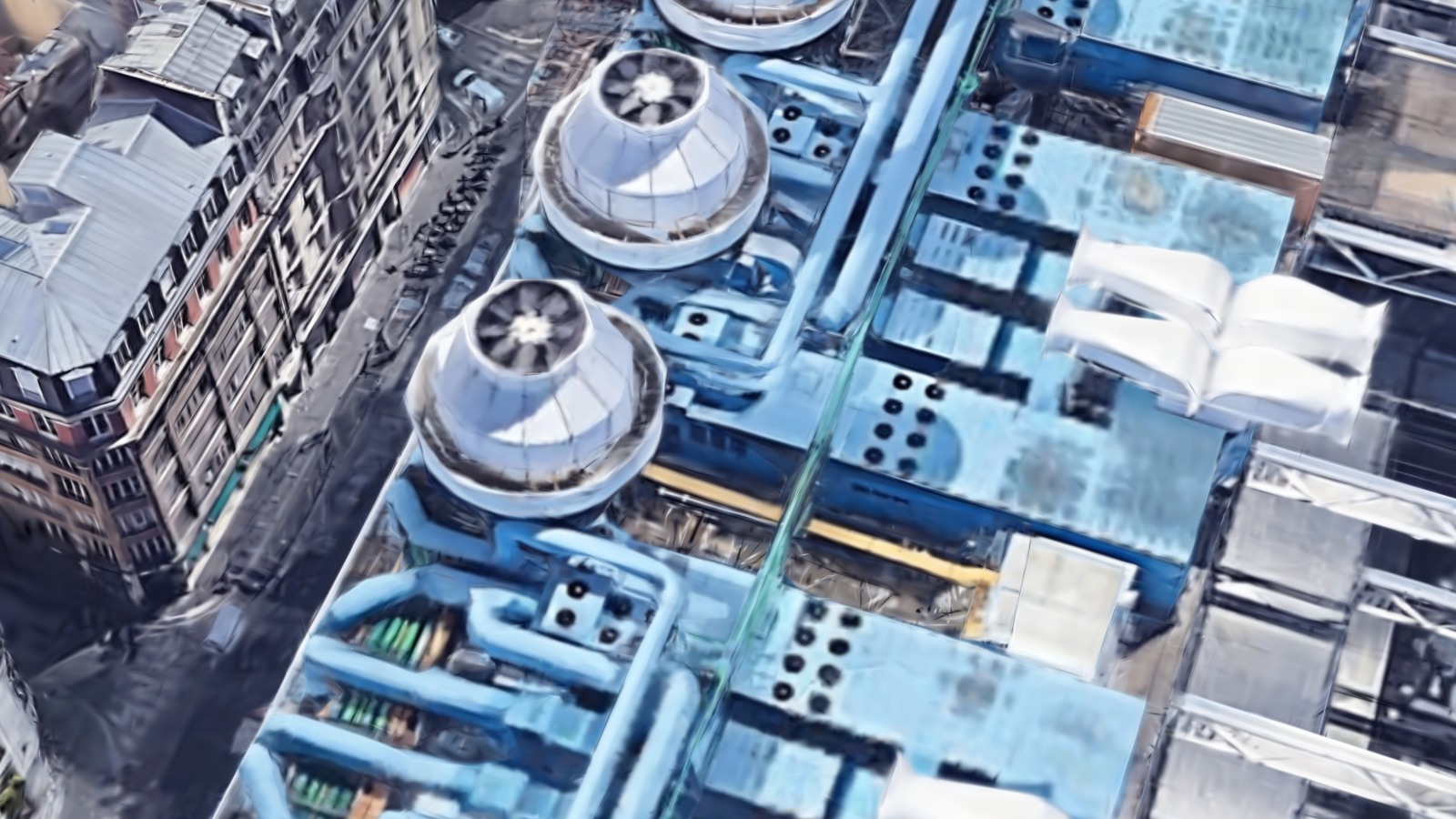}{25.83}{26.51} &
        \qualmetriccrop[0.115\textwidth]{80 457 1000 150}{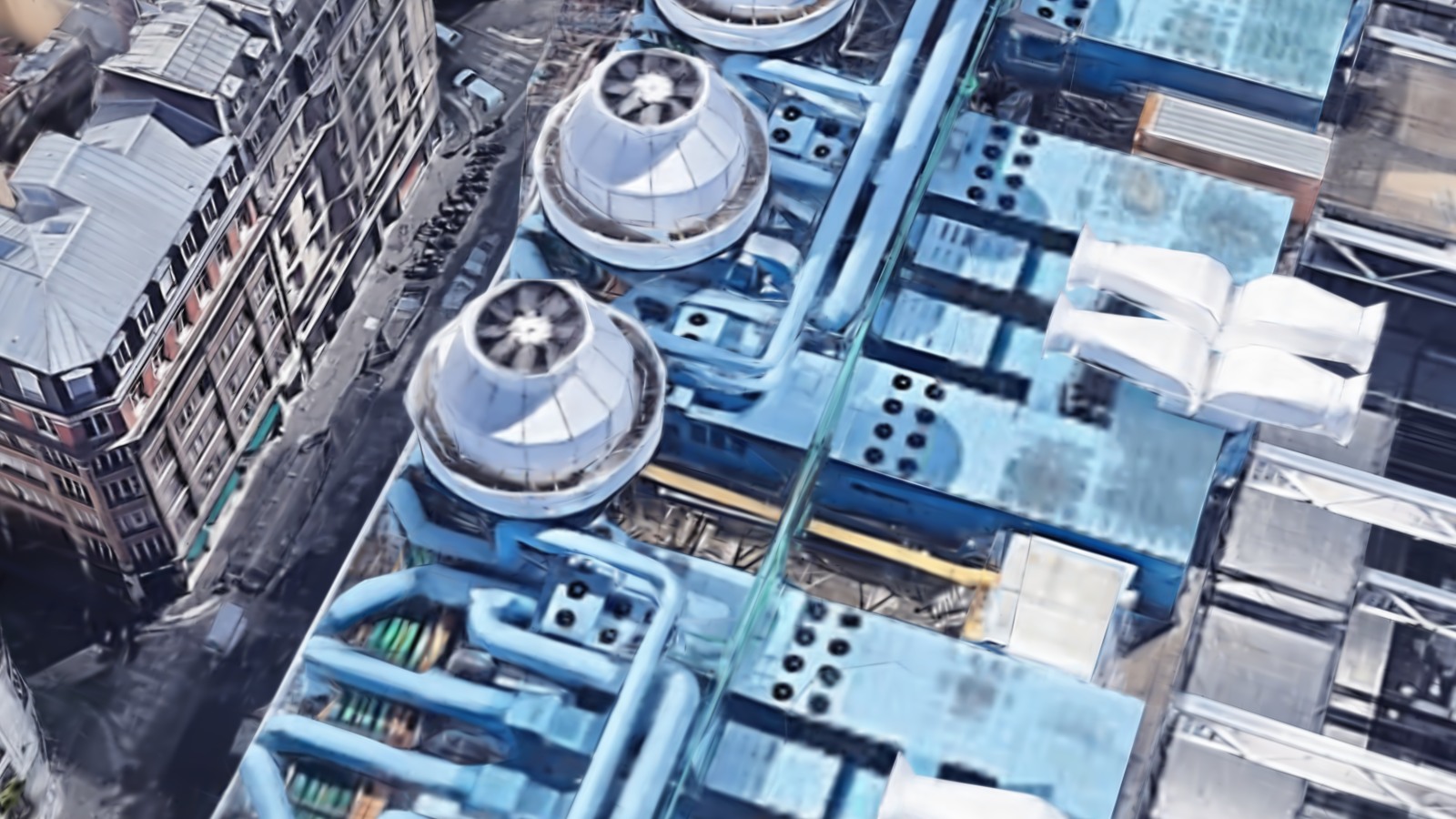}{25.77}{14.22} &
        \qualmetriccrop[0.115\textwidth]{80 457 1000 150}{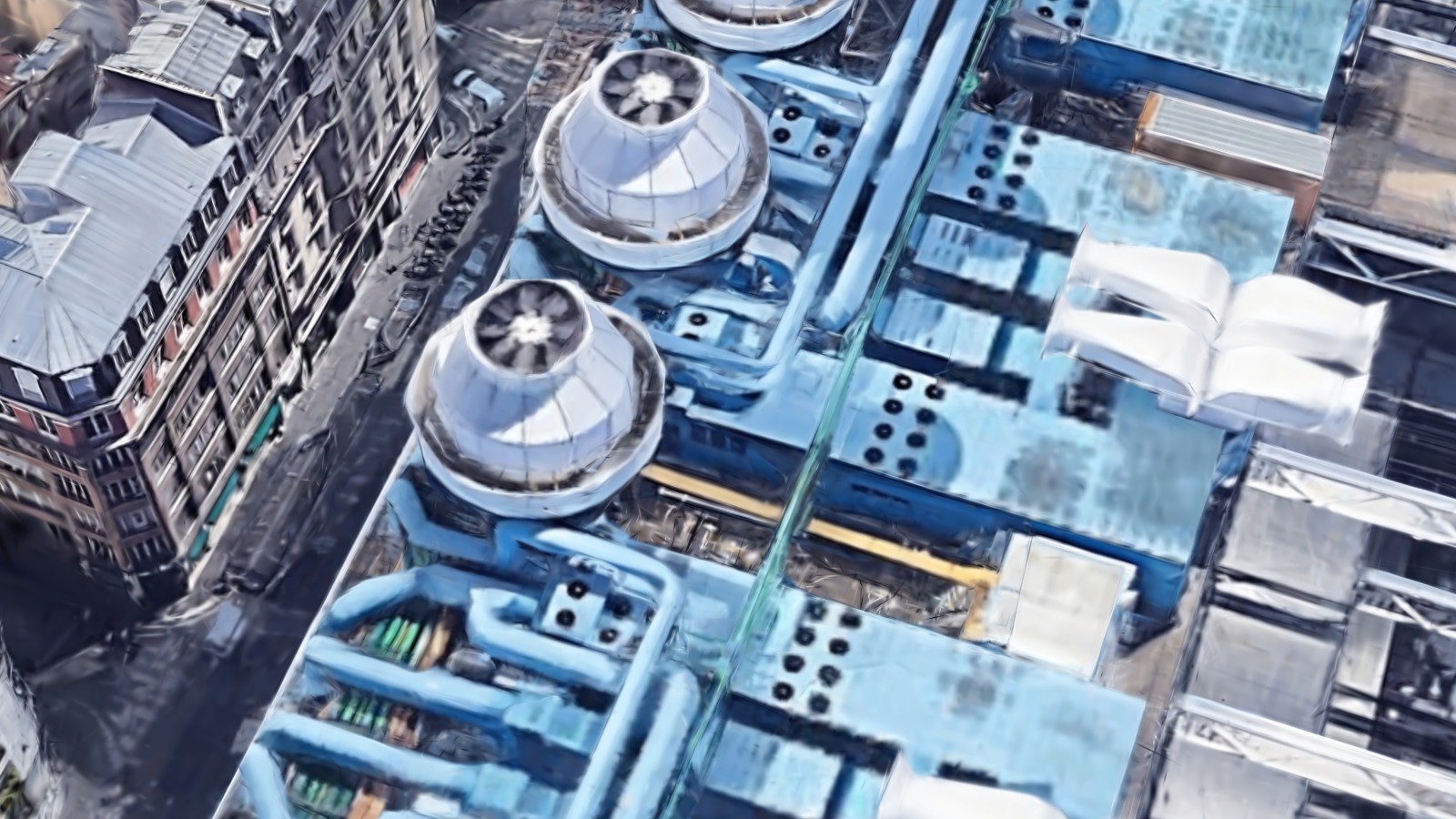}{25.34}{28.20} \\
        \qualscenelabel{rome} &
        \qualcontext[0.115\textwidth]{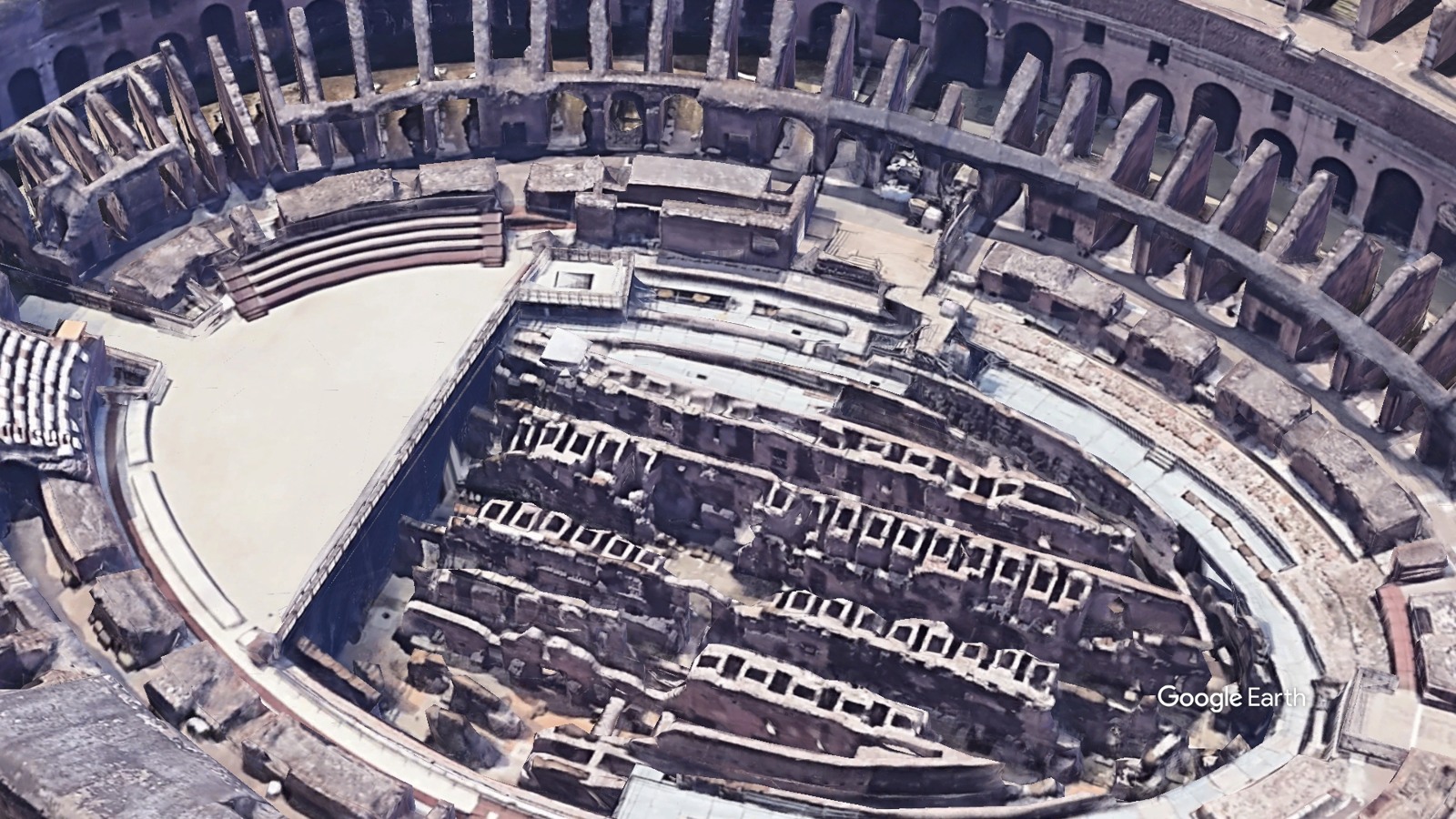}{0.381}{0.681}{0.278}{0.578} &
        \qualdetail[0.115\textwidth]{610 250 510 380}{visualization/bungeenerf/rome/GT.jpg} &
        \qualmetriccrop[0.115\textwidth]{610 250 510 380}{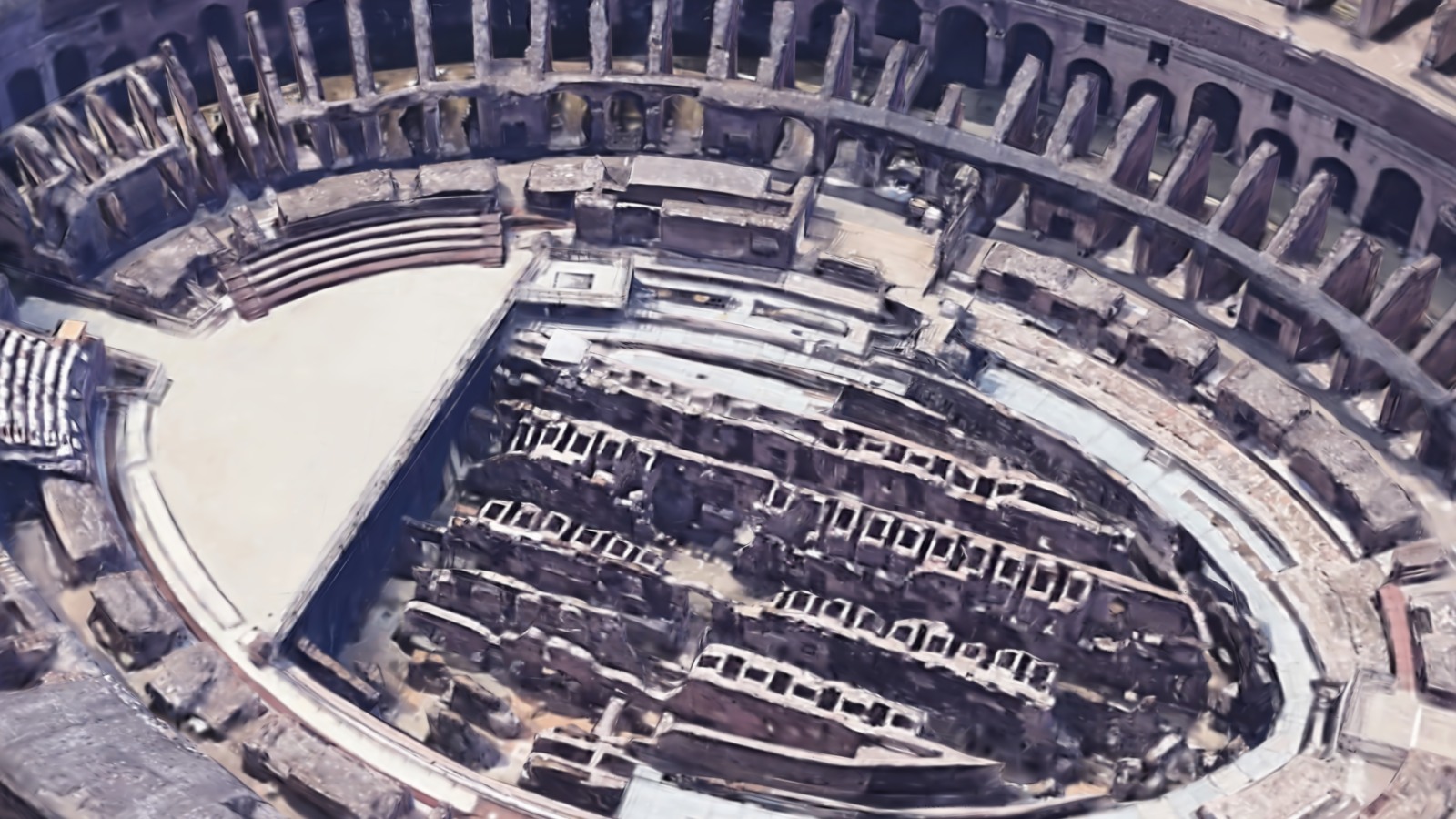}{24.81}{18.66} &
        \qualmetriccrop[0.115\textwidth]{610 250 510 380}{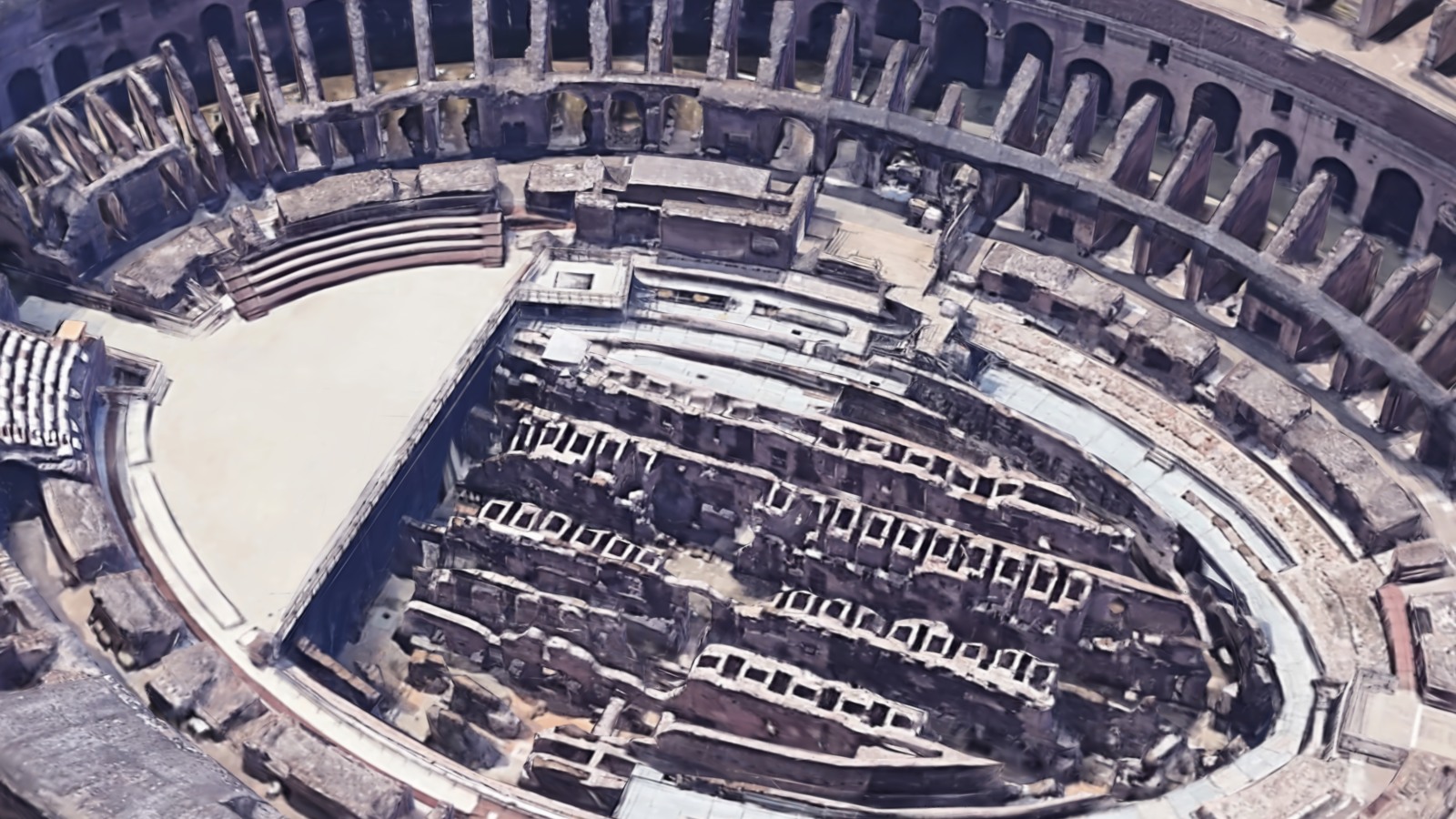}{26.47}{22.25} &
        \qualmetriccrop[0.115\textwidth]{610 250 510 380}{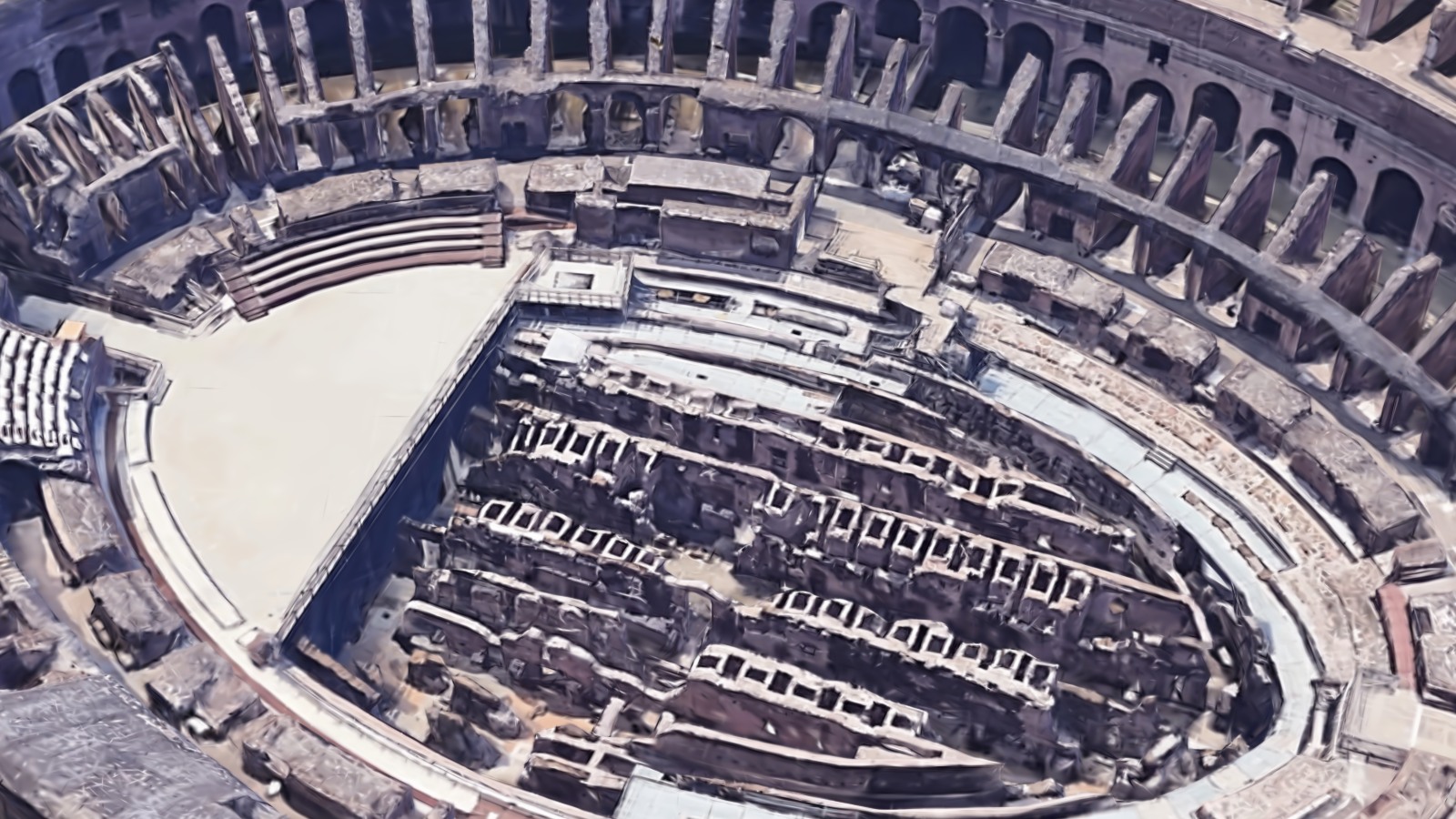}{26.80}{24.16} &
        \qualmetriccrop[0.115\textwidth]{610 250 510 380}{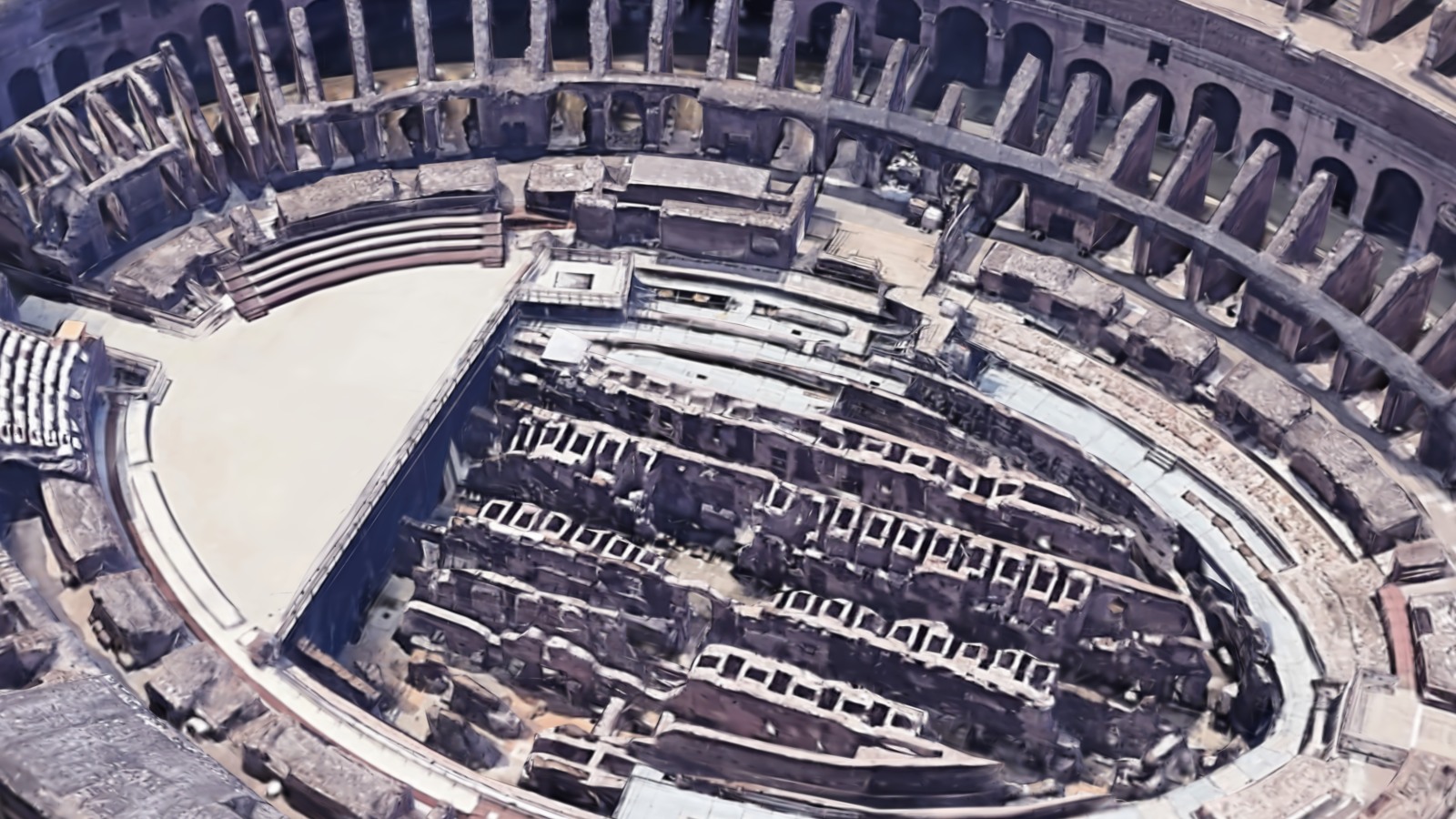}{25.56}{15.81} &
        \qualmetriccrop[0.115\textwidth]{610 250 510 380}{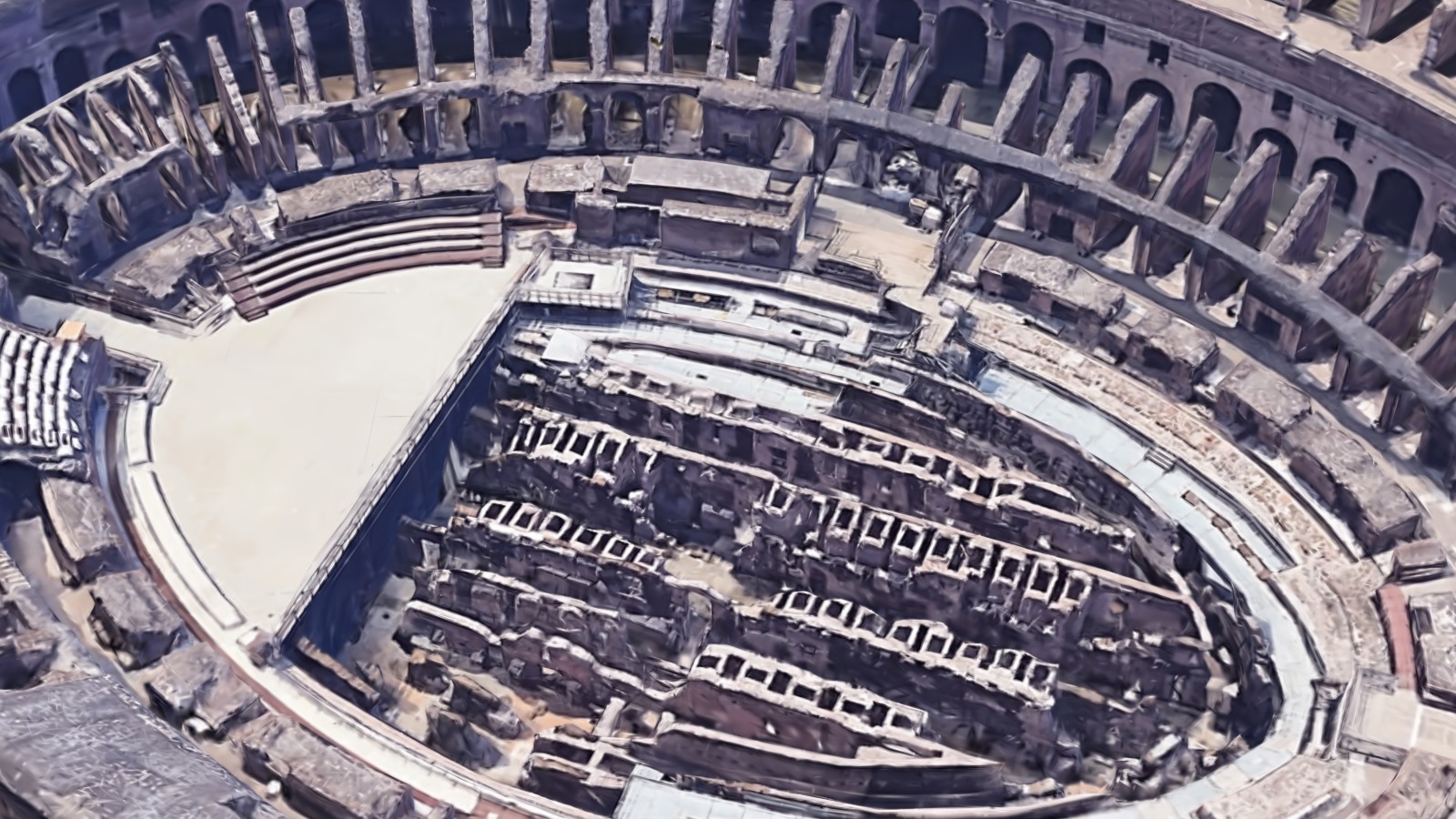}{26.77}{25.42} \\
        \bottomrule
    \end{tabular*}
    \caption{Early-prefix comparison on representative test views. Each row shows the scene context and a shared detail crop. Red boxes mark the detail region; gray tags report full-view PSNR (dB) and cumulative size (MB). The selected PCGS prefixes/decoded Octree-GS-EC-LR LoDs are lv1/LoD2, lv2/LoD0, and lv2/LoD0, respectively.}
    \label{fig:qualitative_methods}
\end{figure*}

\begin{table}[t]
  \centering
  \footnotesize
    \caption{Configurations of the ablation variants. Bounded prefix supervision runs from iteration 1{,}500 to 25{,}000; \texttt{None} uses only the full-model loss.}
  \label{tab:ablation_settings}
  \setlength\tabcolsep{0.04cm}
  \begin{tabular*}{\columnwidth}{@{\extracolsep{\fill}}lcccccc@{}}
    \toprule
    Setting & Closure & \shortstack{Par. feat.\\context} & \shortstack{Entropy\\context} & \shortstack{Level\\balance} & \shortstack{Prefix\\superv.} & Refine. \\
    \midrule
    Loose hierarchy & Loose & Off & Off & Off & Bounded & Off \\
    Strict hierarchy & Strict & Off & Off & Off & Bounded & Off \\
    w/o entropy context & Strict & On & Off & On & Bounded & On \\
    w/o level balance & Strict & On & On & Off & Bounded & On \\
    w/o refinement & Strict & On & On & On & Bounded & Off \\
    w/o prefix supervision & Strict & On & On & On & None & On \\
    \textbf{Full ProGS} & \textbf{Strict} & \textbf{On} & \textbf{On} & \textbf{On} & \textbf{Bounded} & \textbf{On} \\
    \bottomrule
  \end{tabular*}

\end{table}

\begin{table*}[t]
  \centering
  \footnotesize
  \setlength\tabcolsep{0.10cm}
    \caption{Absolute ablation results. Prefix size is cumulative through lv0 and includes the refinement side stream when enabled; base full size excludes it to isolate the Stage~1 codec. Each delta is the variant value minus Full ProGS.}
  \label{tab:component_ablation}
  \begin{tabular*}{\textwidth}{@{\extracolsep{\fill}}lcccccccc@{}}
    \toprule
    Setting & \shortstack{Base full\\size (MB)} & \shortstack{$\Delta$ full\\size (MB)} & \shortstack{lv0 prefix\\size (MB)} & \shortstack{$\Delta$ lv0\\size (MB)} & \shortstack{Full\\PSNR (dB)} & \shortstack{$\Delta$ full\\(dB)} & \shortstack{lv0\\PSNR (dB)} & \shortstack{$\Delta$ lv0\\(dB)} \\
    \midrule
    Loose hierarchy & 16.718 & $-5.047$ & 4.199 & $-6.636$ & 26.696 & $-0.322$ & 21.877 & $-2.277$ \\
    Strict hierarchy & 34.724 & $+12.958$ & 19.482 & $+8.647$ & 27.300 & $+0.282$ & 22.696 & $-1.458$ \\
    \textbf{Full ProGS} & \textbf{21.766} & -- & \textbf{10.747} & -- & \textbf{27.018} & -- & \textbf{24.154} & -- \\
    \midrule
    w/o entropy context & 22.412 & $+0.647$ & 10.778 & $+0.032$ & 27.054 & $+0.036$ & 24.135 & $-0.019$ \\
    w/o level balance & 24.376 & $+2.610$ & 13.470 & $+2.723$ & 27.156 & $+0.138$ & 24.321 & $+0.167$ \\
    w/o refinement & 21.766 & $0.000$ & 10.365 & $-0.382$ & 27.018 & $0.000$ & 22.602 & $-1.552$ \\
    w/o prefix supervision & 74.291 & $+52.525$ & 37.062 & $+26.316$ & 26.367 & $-0.651$ & 19.263 & $-4.891$ \\
    \textbf{Full ProGS} & \textbf{21.766} & -- & \textbf{10.747} & -- & \textbf{27.018} & -- & \textbf{24.154} & -- \\
    \bottomrule
  \end{tabular*}

\end{table*}

\subsection{Comparative Experiments}

\noindent\textbf{Compared methods.}
Tab.~\ref{tab:main_results} includes reference representations, general and learned single-rate compressors, the official Octree-GS reproduction as an LoD representation~\cite{ren2024octreegsconsistentrealtimerendering}, the native progressive codec PCGS~\cite{chen2026pcgs}, the adapted coded LoD baseline Octree-GS-EC, and ProGS. Tab.~\ref{tab:method_properties} distinguishes serialized checkpoints, independently optimized complete models, view-dependent LoDs, and cumulative coded prefixes.

\noindent\textbf{Rate--distortion comparison.}
Tab.~\ref{tab:main_results} and Fig.~\ref{fig:rd} compare cumulative progressive streams and single-rate endpoints. At the reported complete endpoints, ProGS-HR has higher SSIM and lower LPIPS than PCGS on all three dataset averages. On BungeeNeRF, it reaches 27.91~dB/.919/.091 at 43.45~MB, compared with 27.15~dB/.872/.213 at 22.82~MB for PCGS. The continued improvement at high rates suggests greater refinement headroom for city-scale unbounded scenes, although it requires a larger payload. The key distinction is topology-rate scalability: PCGS activates and refines a complete anchor layout, whereas ProGS reveals a parent-closed topology so each prefix owns its renderable structure and causal decoding state. Tab.~\ref{tab:component_ablation} quantifies the rate cost of parent closure. Over the common PSNR--rate intervals, ProGS is 1.00--1.67~dB below PCGS. Official Octree-GS reaches 28.10~dB on BungeeNeRF; EC-LR dominates EC-HR in dataset-average PSNR and is used below.

\noindent\textbf{Progressive transmission under byte budgets.}
The byte-budget evaluation compares native ProGS and PCGS prefixes, decoded Octree-GS-EC-LR LoDs, and two post-hoc baselines on \textit{room}, \textit{hollywood}, and \textit{truck}. HAC-Rand and HAC++-Rand split a trained single-rate model into five byte-balanced anchor packets and report the ten-seed mean. For scene $s$, let $\mathcal M_s$ contain ProGS, PCGS, Octree-GS-EC-LR, HAC-Rand, and HAC++-Rand.
Octree-GS-EC-HR is excluded because its dataset-average endpoints are PSNR-dominated by EC-LR. The common reachable target is defined as
\begin{equation}
    Q_s^*=\min_{m\in\mathcal M_s}Q_{m,s}^{\mathrm{final}}-1\ \mathrm{dB},
    \label{eq:streaming_quality_target}
\end{equation}
where $Q_{m,s}^{\mathrm{final}}$ is the final-point PSNR. The received size is the first cumulative point reaching $Q_s^*$, using mean PSNR for the randomized baselines. At this target, ProGS/PCGS/Octree-GS-EC-LR require 3.57/5.17/3.36 MB on \textit{room}, 18.50/13.31/22.31 MB on \textit{hollywood}, and 9.09/6.93/7.81 MB on \textit{truck}.

Against the post-hoc HAC-Rand and HAC++-Rand baselines, ProGS reduces received size by 30.7--60.7\% and 22.6--53.4\%, respectively. This demonstrates the benefit of optimizing renderable prefixes instead of post-hoc partitioning. Relative to the native progressive or coded-LoD alternatives, the result is scene-dependent: ProGS uses 31.0\% fewer bytes than PCGS on \textit{room} and 17.1\% fewer than EC-LR on \textit{hollywood}, while requiring 6.3--39.1\% more bytes in the other four comparisons.

\subsection{Qualitative Results}

Fig.~\ref{fig:LoD} shows a fixed detail crop under cumulative ProGS decoding (layout as in the caption). The first prefix preserves the dominant structure but remains blurred in high-frequency regions, while later prefixes recover texture and boundary detail in a predictable coarse-to-fine order.

Fig.~\ref{fig:qualitative_methods} extends the early-prefix comparison across \textit{garden}, \textit{pompidou}, and \textit{rome}. On \textit{garden}, ProGS lv1 gives 26.95~dB/0.150 LPIPS at 24.27~MB, versus 27.47/0.148 at 24.93~MB for PCGS lv1. On \textit{pompidou}, ProGS lv1 gives 26.21/0.108 at 25.21~MB, versus 25.83/0.221 at 26.51~MB for PCGS lv2 and 25.34/0.221 at 28.20~MB for HAC++. Selected Octree-GS-EC-LR points are lower-rate references where the hierarchy contains no nearby occupied LoD; HAC++ is complete and single-rate.

\begin{figure*}[t]
    \centering
    \includegraphics[width=\textwidth]{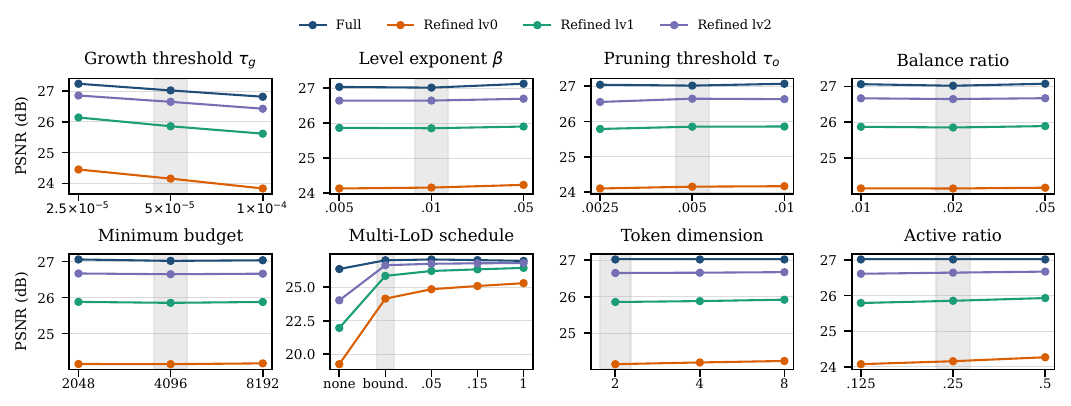}
    \caption{Average parameter sensitivity. From left to right and top to bottom, the panels scan the growth threshold, level exponent, pruning threshold, balance ratio, minimum budget, multi-LoD schedule, token dimension, and active ratio; the two-setting quantization comparison is reported in Tab.~\ref{tab:parameter_sensitivity}. The gray band marks the default setting in each panel; full denotes the base complete path and refined lv0--lv2 the frozen-base refined prefixes.}
    \label{fig:parameter_sensitivity}
\end{figure*}

\begin{table*}[t]
  \centering
  \scriptsize
  \caption{Parameter sensitivity. Bold marks the main-method settings. Panel (a) independently retrains each anchor-growing or leaf-pruning setting, panel (b) retrains the main model, and panel (c) varies only the refinement stage. Base excludes the refinement side stream; Ref. includes the decoder weights, selection masks, and quantized tokens used by refined-prefix quality.}
  \label{tab:parameter_sensitivity}
  \begin{minipage}[t]{0.33\textwidth}
    \centering
    \textbf{(a) Anchor-growth and leaf-pruning parameters}\\[0.3em]
    \setlength\tabcolsep{0.015cm}
    \begin{tabular*}{\linewidth}{@{\extracolsep{\fill}}llcccc@{}}
      \toprule
      Param. & Set. & \shortstack{Base\\(MB)} & \shortstack{Full\\(dB)} & \shortstack{lv0\\(dB)} & \shortstack{Anch.\\(k)} \\
      \midrule
      $\tau_g$ & $2.5\!\times\!10^{-5}$ & 37.382 & 27.234 & 24.450 & 1278.9 \\
                 & $\boldsymbol{5\!\times\!10^{-5}}$ & 21.766 & 27.018 & 24.154 & 775.6 \\
                 & $1\!\times\!10^{-4}$ & 13.879 & 26.813 & 23.834 & 529.5 \\
      \midrule
      $\beta$ & 0.005 & 21.283 & 27.039 & 24.128 & 761.2 \\
               & \textbf{0.01} & 21.766 & 27.018 & 24.154 & 775.6 \\
               & 0.05 & 23.318 & 27.129 & 24.232 & 819.9 \\
      \midrule
      $\tau_o$ & 0.0025 & 22.032 & 27.043 & 24.103 & 787.8 \\
                & \textbf{0.005} & 21.766 & 27.018 & 24.154 & 775.6 \\
                & 0.01 & 21.438 & 27.070 & 24.167 & 761.1 \\
      \bottomrule
    \end{tabular*}
  \end{minipage}\hfill
  \begin{minipage}[t]{0.35\textwidth}
    \centering
    \textbf{(b) Main-model and supervision parameters}\\[0.3em]
    \setlength\tabcolsep{0.015cm}
    \begin{tabular*}{\linewidth}{@{\extracolsep{\fill}}lcccc@{}}
      \toprule
      Param. & Set. & \shortstack{Base\\(MB)} & \shortstack{Full\\(dB)} & \shortstack{lv0\\(dB)} \\
      \midrule
      Balance ratio & 0.01 & 23.119 & 27.062 & 24.157 \\
       & \textbf{0.02} & 21.766 & 27.018 & 24.154 \\
       & 0.05 & 23.415 & 27.073 & 24.175 \\
      \midrule
      Min. budget & 2048 & 22.928 & 27.057 & 24.159 \\
       & \textbf{4096} & 21.766 & 27.018 & 24.154 \\
       & 8192 & 22.841 & 27.035 & 24.176 \\
      \midrule
      LoD schedule & none & 75.068 & 26.367 & 19.263 \\
       & \textbf{bounded} & 21.766 & 27.018 & 24.154 \\
       & tail 0.05 & 21.903 & 27.080 & 24.858 \\
       & tail 0.15 & 21.910 & 27.030 & 25.098 \\
       & tail 1.0 & 21.664 & 26.975 & 25.308 \\
      \bottomrule
    \end{tabular*}
  \end{minipage}\hfill
  \begin{minipage}[t]{0.31\textwidth}
    \centering
    \textbf{(c) Frozen-base refinement parameters}\\[0.3em]
    \setlength\tabcolsep{0.015cm}
    \begin{tabular*}{\linewidth}{@{\extracolsep{\fill}}lccccc@{}}
      \toprule
      Param. & Set. & \shortstack{Ref.\\(MB)} & \shortstack{lv0\\(dB)} & \shortstack{lv1\\(dB)} & \shortstack{lv2\\(dB)} \\
      \midrule
      Token dim. & \textbf{2} & 0.382 & 24.154 & 25.854 & 26.647 \\
       & 4 & 0.650 & 24.201 & 25.879 & 26.656 \\
       & 8 & 1.186 & 24.242 & 25.918 & 26.671 \\
      \midrule
      Active ratio & 0.125 & 0.248 & 24.071 & 25.792 & 26.615 \\
       & \textbf{0.25} & 0.382 & 24.154 & 25.854 & 26.647 \\
       & 0.5 & 0.649 & 24.265 & 25.932 & 26.676 \\
      \midrule
      Quantization & 6 bit & 0.315 & 24.158 & 25.854 & 26.647 \\
       & \textbf{8 bit} & 0.382 & 24.154 & 25.854 & 26.647 \\
      \bottomrule
    \end{tabular*}
  \end{minipage}
\end{table*}

\begin{figure*}[t]
    \centering
    \includegraphics[width=\textwidth]{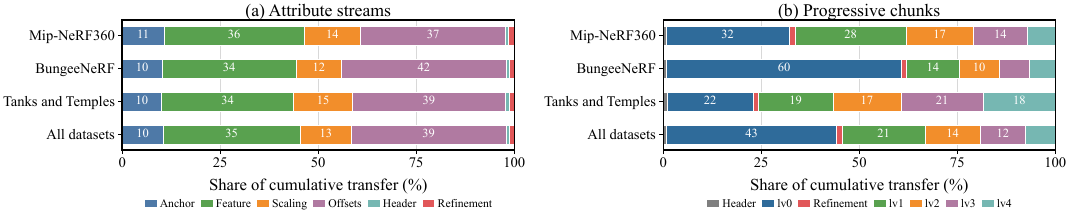}
    \caption{Average bitstream allocation. \textbf{Left:} shares by parameter type. \textbf{Right:} shares by progressive chunk. The refinement side stream follows lv0, is shown separately, and remains in subsequent cumulative payloads.}
    \label{fig:bitstream_analysis}
\end{figure*}

\subsection{Ablation Studies}

Tabs.~\ref{tab:ablation_settings} and~\ref{tab:component_ablation} report the variant switches and absolute three-dataset averages; the upper block probes hierarchy closure with the other components disabled, whereas the lower block isolates each component of Full ProGS. Strict closure is the mechanism that guarantees decoder-valid prefixes: it eliminates missing-parent violations in all 17 scenes (16 loose runs violate), so decoding never depends on fetching or inferring unavailable ancestors. This guarantee adds 18.005 MB relative to the loose hierarchy while improving full/lv0 PSNR by 0.604/0.819 dB. In the component probes, entropy context saves 0.647 MB with negligible PSNR change, and level balancing saves 2.610 MB at a 0.138/0.167 dB full/lv0 cost. The 0.382-MB refinement stream recovers 1.552 dB at lv0. Most critically, removing bounded prefix supervision adds 52.525 MB and reduces full/lv0 PSNR by 0.651/4.891 dB.

\subsection{Parameter Sensitivity}

Fig.~\ref{fig:parameter_sensitivity} and Tab.~\ref{tab:parameter_sensitivity} summarize the three-dataset study at $\lambda_\mathrm{e}=0.004$. The growth threshold $\tau_g$ is the main structural rate control: halving/doubling it changes the base stream by $+15.616/-7.887$ MB and full/lv0 PSNR by $+0.216/+0.296$ and $-0.205/-0.320$ dB, while the other growth parameters have smaller effects. This asymmetric response places the default near a knee of the structural rate--quality curve. Nonzero supervision tails keep rate nearly unchanged and improve lv0 by 0.704--1.155 dB. Increasing token dimension or active ratio costs up to 0.804/0.267 MB for at most 0.088/0.111 dB lv0 gain, and 6- and 8-bit quantization are nearly identical, supporting the default two-dimensional, 25\%-active, 8-bit design. The defaults are reference trade-offs rather than independently optimized extrema, with nonzero tails available when lv0 quality is the deployment priority.

\begin{table}[t]
  \centering
  \footnotesize
    \caption{Mean ProGS prefix efficiency on one RTX~4090 (17 scenes; $\lambda_\mathrm{e}=0.004$). lv0--lv4 are the first through fifth renderable levels. $\mathrm{TTFF}_{50}$ represents the time-to-first-frame from the start of transmission over a 50-Mbps link.}
  \label{tab:efficiency}
  \setlength\tabcolsep{0.05cm}
  \begin{tabular*}{\columnwidth}{@{\extracolsep{\fill}}lccccc@{}}
    \toprule
    Metric & lv0 & lv1 & lv2 & lv3 & lv4 \\
    Size (MB) & 10.75 & 15.42 & 18.23 & 20.62 & 22.15 \\
    Cum. Dec. (s) & 12.84 & 19.62 & 23.72 & 26.68 & 28.49 \\
    FPS & 120.2 & 114.1 & 105.0 & 143.6 & 149.8 \\
    $\mathrm{TTFF}_{50}$ (s) & 14.57 & 22.10 & 26.65 & 29.99 & 32.04 \\
    \midrule
    \multicolumn{3}{l}{Train: 3.35 h} & \multicolumn{3}{r}{Ref.: 0.382 MB; anchors: 555k} \\
    \bottomrule
  \end{tabular*}

\end{table}

\subsection{Bitstream Analysis and System Costs}

Fig.~\ref{fig:bitstream_analysis} shows that feature and offset attributes consume 74.3\% of the bytes, while the lv0 and lv0--lv1 level chunks account for 43.5\% and 64.7\%; a useful prefix thus has a substantial initial cost. At nearby native endpoints (22.15 versus 23.69~MB), a same-RTX~4090 check gives ProGS/PCGS encoding times of 12.54/11.58~s and complete-stream decoding times of 28.49/21.34~s, respectively. Tab.~\ref{tab:efficiency} reports cumulative size, decoding, FPS, and time-to-first-frame (TTFF) for all five ProGS prefixes. TTFF is the elapsed time from the start of transmission to the first rendered frame. For a 50-Mbps link, it is defined that
\begin{equation}
    \mathrm{TTFF}_{50}=\frac{8S}{50}+D+\frac{1}{\mathrm{FPS}},
    \label{eq:ttff50}
\end{equation}
where $S$ is the cumulative received size in MB and $D$ is the cumulative decoding time in seconds. At 50~Mbps, decoding accounts for 88.1\% of lv0 TTFF and 88.9\% of lv4 TTFF; transmission contributes 11.8\% and 11.1\%, with rendering below 0.1\% at both endpoints. FPS rises after lv2 because refinement is bypassed. Thus, latency reduction on this hardware depends mainly on accelerating decoding or overlapping it with transmission. The 43.5\% lv0 share makes base-prefix savings cumulative across later operating points, while the 74.3\% feature-and-offset share identifies the main target for improved attribute coding.

\section{Conclusion}\label{sec:conclusion}

ProGS structures anchor-based 3DGS into parent-closed octree prefixes in which every completed level is causally decodable by the proposed parent-conditioned entropy model and directly renderable. Parent-causal entropy coding, level-balanced growth, bounded multi-LoD training, and parent-anchor refinement jointly produce compact early prefixes while preserving the complete-model path. One fixed-$\lambda$ model therefore provides five deployment rates from a single progressive stream.

Across completed prefixes, ProGS provides topology-rate scalability and real-time rendering from one stream. ProGS-HR has higher endpoint SSIM and lower LPIPS than PCGS on all three dataset averages. At common targets on one scene per dataset, ProGS uses 30.7--60.7\% fewer bytes than HAC-Rand and 22.6--53.4\% fewer than HAC++-Rand. These gains are operating-point dependent: ProGS is 1.00--1.67~dB below PCGS over their common PSNR--rate intervals, while cumulative entropy decoding and training remain the main efficiency bottlenecks. Future work will investigate incremental or parallel decoding, tighter joint allocation of structure and attribute rates, and online prefix scheduling under measured network and device constraints.

{
\normalem
\bibliographystyle{IEEEtran}
\bibliography{main}

}
\newpage

\vfill

\end{document}